\documentclass[twoside,11pt]{article}

\usepackage{jmlr2e}
\usepackage{multirow}
\usepackage{rotating}
\usepackage{wrapfig}
\usepackage{lineno}
\usepackage{epsfig}
\usepackage{amsmath,amssymb}
\usepackage{color}
\usepackage{booktabs}       
\usepackage{amsfonts,amsmath}       
\usepackage{nicefrac}       
\usepackage{microtype} 
\usepackage{float}
\usepackage{amsmath}
 
\usepackage{enumerate}
\usepackage{enumitem}   

\usepackage{epstopdf}

\jmlrheading{1}{2022}{1-48}{4/00}{10/00}{Xiaofeng Liu, et al. Contact to: xliu61@mgh.harvard.edu}

\ShortHeadings{Deep Unsupervised Domain Adaptation}{X. Liu, C. Yoo, F. Xing, H. Oh, G. El Fakhri, J. W. Kang, and J. Woo}
\firstpageno{1}

\begin{document}

\title{Deep Unsupervised Domain Adaptation: A Review of Recent Advances and Perspectives}

\author{\name Xiaofeng Liu$^a$, Chaehwa Yoo$^b$, Fangxu Xing$^a$, Hyejin Oh$^b$, Georges El Fakhri$^a$, Je-Won Kang$^b$, Jonghye Woo$^a$ \\
       \addr $^a$ Gordon Center for Medical Imaging, Massachusetts
General Hospital and Harvard Medical School, Boston, MA, USA\\
       \addr $^b$ Dept. of Electronic and Electrical Engineering and Graduate Program in Smart Factory, Ewha Womans University, Seoul, South Korea}

\editor{C.C. Jay Kuo}

\maketitle

\begin{abstract}

Deep learning has become the method of choice to tackle real-world problems in different domains, partly because of its ability to learn from data and achieve impressive performance on a wide range of applications. However, its success usually relies on two assumptions: (i) vast troves of labeled datasets are required for accurate model fitting, and (ii) training and testing data are independent and identically distributed.~Its performance on unseen target domains, thus, is not guaranteed, especially when encountering out-of-distribution data at the adaptation stage.~The performance drop on data in a target domain is a critical problem in deploying deep neural networks that are successfully trained on data in a source domain.~Unsupervised domain adaptation (UDA) is proposed to counter this, by leveraging both labeled source domain data and unlabeled target domain data to carry out various tasks in the target domain.~UDA has yielded promising results on natural image processing, video analysis, natural language processing, time-series data analysis, medical image analysis, etc. In this review, as a rapidly evolving topic, we provide a systematic comparison of its methods and applications. In addition, the connection of UDA with its closely related tasks, e.g., domain generalization and out-of-distribution detection, has also been discussed. Furthermore, deficiencies in current methods and possible promising directions are highlighted.
\end{abstract}

\begin{keywords}
  Deep Learning, Unsupervised Domain Adaptation, Transfer Learning, Adversarial Training, Self Training.
\end{keywords}

\section{Introduction}\label{sec1}

Deep learning is a subfield of machine learning, which aims at discovering multiple levels of distributed representations of input data via hierarchical
architectures \citep{goodfellow2016deep}. For the past several years, there has been an explosion of deep learning-based approaches, where deep learning has substantially improved state-of-the-art approaches to diverse machine learning problems and applications \citep{lecun2015deep}. In particular, deep learning has transformed conventional signal processing approaches into simultaneously learning both features and a prediction model in an end-to-end fashion \citep{bengio2013representation}. Although supervised deep learning is the most prevalent and successful approach for a variety of tasks, its success hinges on (i) vast troves of labeled training data and (ii) the assumption of independent and identically distributed ($i.i.d.$) training and testing datasets \citep{huo2022domain}. Because reliable labeling of massive datasets for various application domains is often expensive and prohibitive, for a task without sufficient labeled datasets in a target domain, there is strong demand to apply trained models, by leveraging rich labeled data from a source domain \citep{xu2022cycle}.~This learning strategy, however, suffers from shifts in data distributions, i.e., domain shift, between source and target domains \citep{zhang2022latent}. As a result, the performance of a trained model can be severely degraded, when encountering out-of-distribution (OOD) data, i.e., a source distribution differs from a target distribution \citep{che2021deep}. For example, the performance of a disease diagnostic system, applied to a population in a target domain that is different from a population in a source domain, cannot be guaranteed. 

\begin{figure}[t]
\begin{center}
\includegraphics[width=1\linewidth]{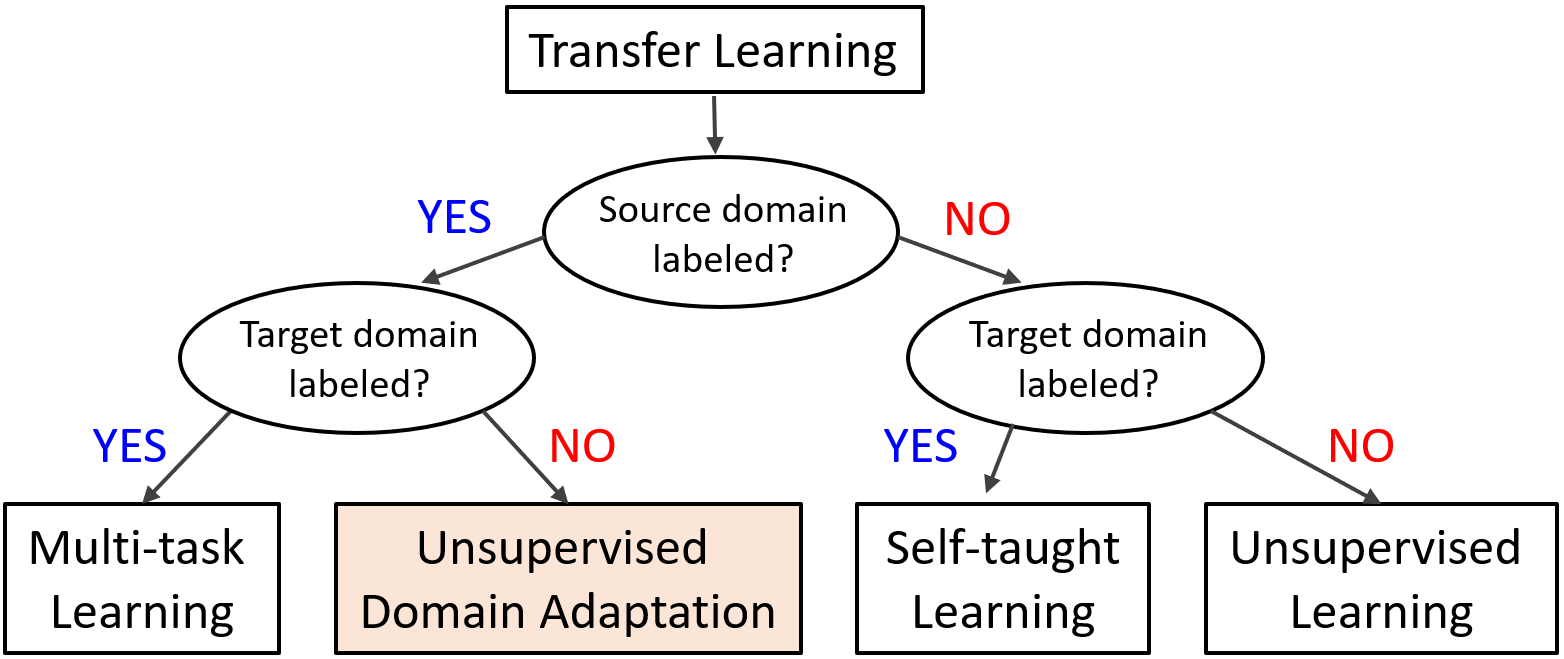}
\end{center} 
\caption{A taxonomy of transfer learning approaches based on the availability of labeled data in a source or target domain.}
\label{fig:1} 
\end{figure}

\begin{figure}[t]
\begin{center}
\includegraphics[width=1\linewidth]{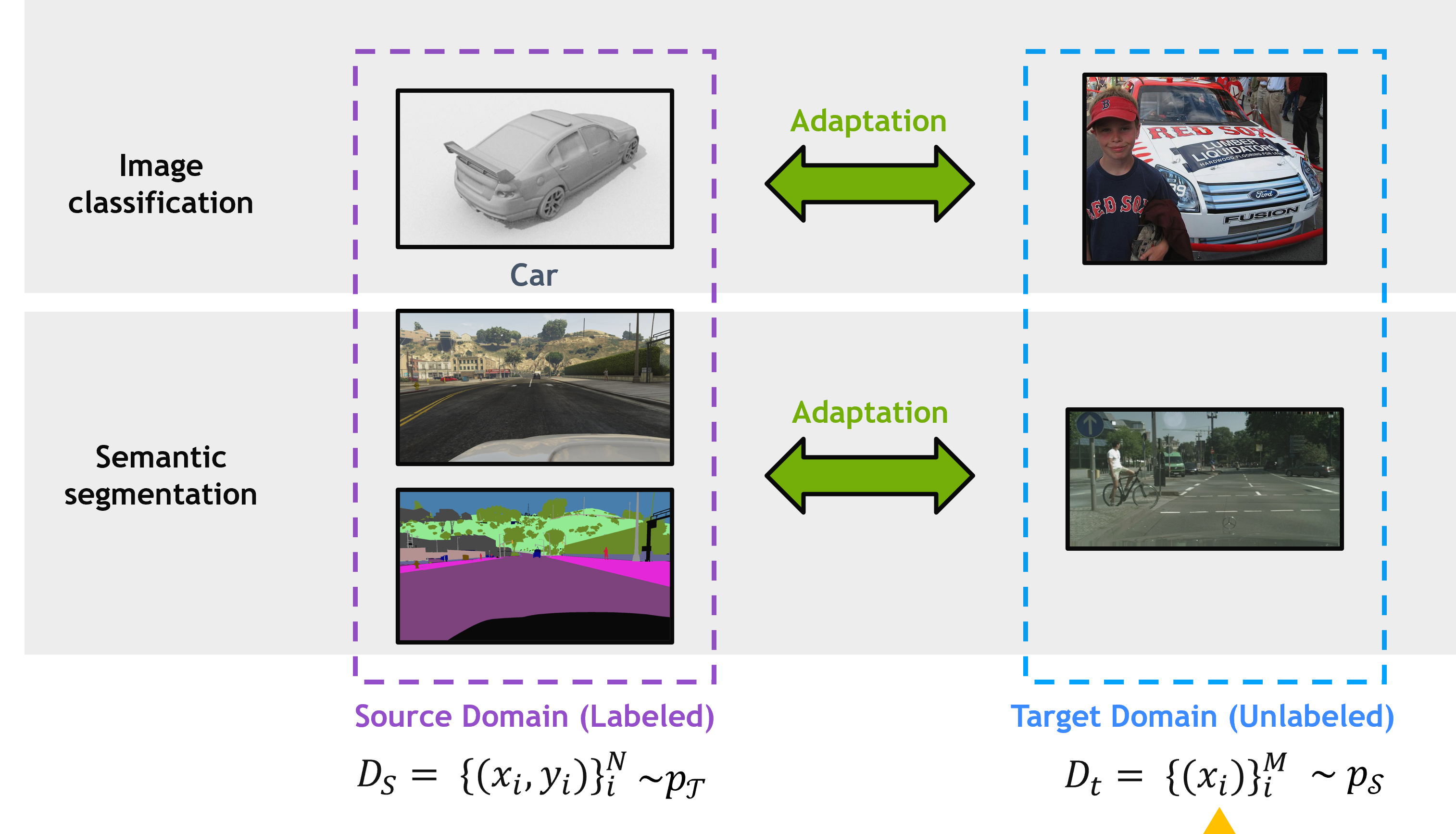}
\end{center} 
\caption{Illustration of the UDA classification and segmentation with the examples on the VisDA17 challenge database. The target domain data are unlabeled, as indicated by the orange triangle.}
\label{fig:UDAconcept} 
\end{figure}

To counter this, unsupervised domain adaptation (UDA) is proposed as a viable solution to migrate knowledge learned from a labeled source domain to unseen, heterogeneous, and unlabeled target domains \citep{liu2021subtype,liu2022subtype}, as shown in Fig. \ref{fig:UDAconcept}. UDA is aimed at mitigating domain shifts between source and target domains \citep{kouw2018introduction}. The solution to UDA is primarily classified into statistic moment matching (e.g., maximum mean discrepancy (MMD) \citep{long2018conditional}), domain style transfer \citep{sankaranarayanan2018generate}, self-training \citep{zou2019confidence,liu2021energy,liu2021generative}, and feature-level adversarial learning \citep{ganin2016domain,he2020classification,he2020image2audio,liu2018data}.

There are several previous review papers focusing on domain adaptation \citep{beijbom2012domain,bungum2011survey,betlehem2015personal,sun2015survey,wang2018deep,csurka2017domain,zhao2018unsupervised,kouw2018introduction,kouw2019review,zhao2020multi,wilson2020survey,ramponi2020neural}, and the broader problem of transfer learning \citep{pan2009survey,cook2013transfer,lazaric2012transfer,shao2014transfer,tan2018survey,zhang2019recent}. As shown in Fig. \ref{fig:1}, domain adaptation can be seen as a special case of transfer learning, with the assumption that labeled data are available only in a source domain \citep{pan2009survey}. In this review paper, we aim to provide a wide coverage of models and algorithms for UDA from a theoretical and practical point of view. This review also touches on emerging approaches, especially those developed recently, providing a thorough comparison of different techniques as well as a discussion of the connection of unique components and methods with unsupervised deep domain adaptation. The coverage of UDA, especially deep learning-based UDA, has been limited in the general transfer learning reviews. Many prior reviews of domain adaption do not incorporate deep learning approaches; however, deep learning-based approaches have been the mainstream of UDA. In addition, some reviews do not touch deeply on domain mapping \citep{csurka2017domain,kouw2018introduction,kouw2019review}, normalization statistic-based \citep{csurka2017domain,kouw2019review,zhao2018unsupervised,zhao2020multi}, ensemble-based \citep{csurka2017domain,wang2018deep,kouw2019review,zhao2018unsupervised}, or self-training-based methods \citep{wilson2020survey}. Moreover, some of them only focus on limited application areas, such as visual data analytics \citep{wang2018deep,csurka2017domain,oza2021unsupervised} or natural language processing (NLP) \citep{ramponi2020neural}. In this review, we provide a holistic view of this promising technique for a wide range of application areas, including natural image processing, video analysis, NLP, time-series data analysis, medical image analysis, and climate and geosciences. The topics with which other review papers dealt are summarized in Table \ref{tabel:1}.

\begin{table}[t]
\centering
\caption{Comparison with the previous UDA survey papers. } 
\resizebox{1\linewidth}{!}{
\begin{tabular}{l||c||c|c|c|c|c|c}
\hline

Surveys & Deep UDA & Generative Mapping & Normalization & Ensemble & Self-training & Self-supervision & Low density\\\hline\hline

\cite{beijbom2012domain} & $\times$ & $\times$& $\times$& $\times$& $\times$& $\times$& $\times$\\

\cite{sun2015survey} & $\times$ & $\times$& $\times$& $\times$& $\times$& $\times$& $\times$\\

\cite{csurka2017domain} & $\surd$& $\times$& $\times$& $\times$& $\times$& $\times$& $\times$\\

\cite{wang2018deep}& $\surd$ & $\surd$ &$\times$& $\times$& $\times$& $\times$& $\times$\\

\cite{zhao2018unsupervised} & $\surd$ & $\surd$& $\times$& $\times$& $\times$& $\times$& $\times$\\

\cite{kouw2019review} & $\surd$ & $\surd$ & $\times$& $\times$& $\surd$* (no deep) & $\times$& $\times$\\

\cite{wilson2020survey}&$\surd$ & $\surd$ & $\surd$& $\surd$& $\times$ & $\times$& $\surd$ \\

\cite{ramponi2020neural}&$\surd$ & $\surd$ & $\times$& $\times$&  $\surd$ & $\times$& $\times$ \\

\cite{zhang2021survey}&$\surd$ & $\surd$ & $\surd$& $\times$& $\surd$ & $\times$& $\surd$ \\

\cite{guan2021domain} & $\surd$ & $\surd$& $\times$& $\times$& $\times$& $\times$& $\times$\\\hline

Ours & $\surd$ & $\surd$& $\surd$& $\surd$& $\surd$& $\surd$& $\surd$\\\hline

\end{tabular}
}
\label{tabel:1} 
\end{table}

The rest of the paper is organized as follows.~We first analyze possible domain shifts in UDA in Sec.~2. Then, various recent UDA methods are discussed and compared to each other in Sec. 3. Next, we show how UDA is applied to multiple application areas in Sec. 4. In Sec. 5, we highlight promising future directions. Finally, we conclude this paper in Sec. 6.

\section{Overview} 

In this section, without loss of generality, we first introduce terms and notations as well as a formal definition of UDA. In UDA, there are an underlying source domain distribution $p_s(x,y)\in p_\mathcal{S}$ and a different target domain distribution $p_t(x,y)\in p_\mathcal{T}$. Then, a labeled dataset $\mathcal{D_{S}}$ is selected $i.i.d.$ from $p_s(x,y)$, and an unlabeled dataset $\mathcal{D_{T}}$ is selected $i.i.d.$ from the marginal distribution $p_t(x)$.~The goal of UDA is to improve a generalization ability of a trained model in a target domain, by learning on both $\mathcal{D_{S}}$ and $\mathcal{D_{T}}$. We note that $\mathcal{Y}=\left\{1,2,\dots,c\right\}$ is the set of the class labels for discriminative tasks, e.g., classification and segmentation. In contrast, $\mathcal{Y}$ can be continuous values, sentences, images, or languages in generative tasks \citep{ge2021baco}. UDA \citep{ganin2016domain,tzeng2017adversarial} is motivated by the following theorem \citep{kouw2018introduction}: 

\vspace{+5pt}
\noindent\textbf{Theorem 1}
For a hypothesis $h$
{\begin{align}
&\mathcal{L}_t(h)\leq \mathcal{L}_s(h)+d[p_\mathcal{S},p_\mathcal{T}]+  \label{e:1}\\
&{\rm{min}}[\mathbb{E}_{x\sim p_s}|p_s(y|x)-p_t(y|x)|,\mathbb{E}_{x\sim p_t}|p_s(y|x)-p_t(y|x)|].\nonumber
\end{align}}Here, $\mathcal{L}_s(h)$ and $\mathcal{L}_t(h)$ are predefined losses with a hypothesis $h$ in source and target domains, respectively. $d[\cdot]$ represents a divergence measure, e.g., the Jensen–Shannon (JS) divergence in the case of conventional adversarial UDA \citep{salimans2016improved}. Of note, the third term on the right hand, ${\rm{min}}[\mathbb{E}_{x\sim p_s}|p_s(y|x)-p_t(y|x)|,\mathbb{E}_{x\sim p_t}|p_s(y|x)-p_t(y|x)|]$, is a negligible value, for which the error in a source domain $\mathcal{L}_s(h)$ and the divergence between two domains is considered an upper bound of the error in a target domain $\mathcal{L}_t(h)$. $\mathcal{L}_s(h)$ can be minimized, using recent advances in supervised learning, e.g., advanced deep feature extractor networks. Overall, UDA methods aim at minimizing the divergence between two domains to lower the upper bound of the generalization error in the target domain $\mathcal{L}_t(h)$.

\begin{figure}[t]
\begin{center}
\includegraphics[width=0.6\linewidth]{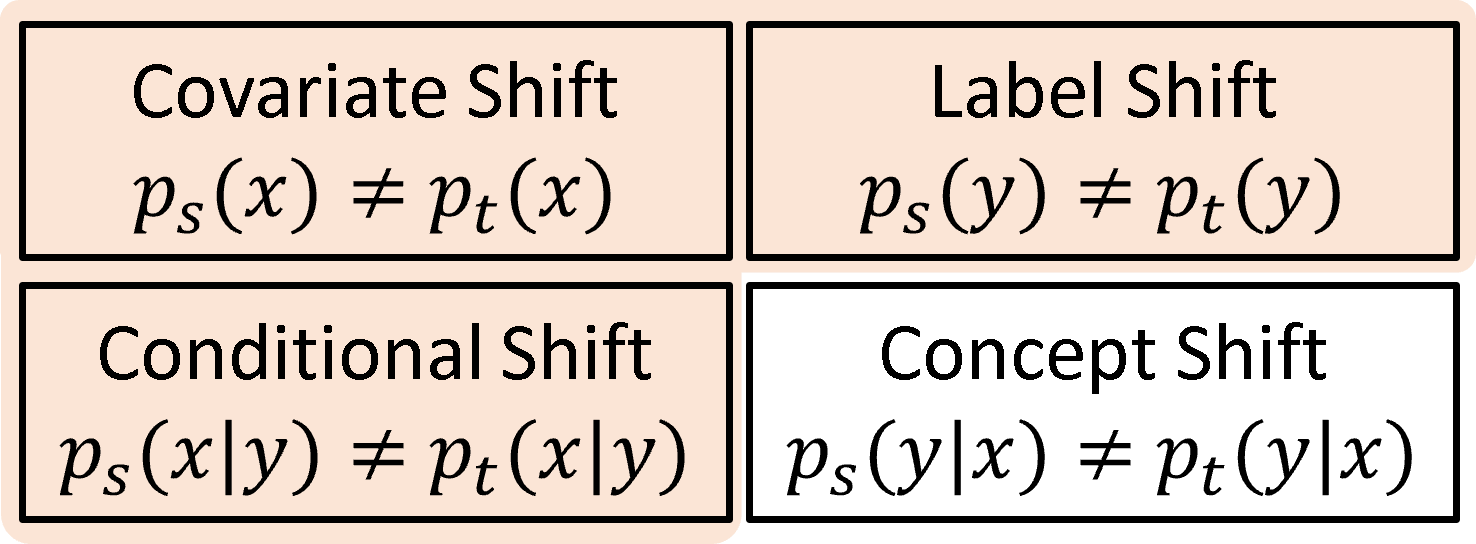}
\end{center} 
\caption{A summary of four possible domain shifts. The red mask indicates the most common domain shift scenarios in UDA \citep{liu2021adversarial}. Note that $p(x)$ can be aligned, if $p(x|y)$ is aligned with the law of total probability \citep{zhang2013domain}.}
\label{fig:d} 
\end{figure}

Domain shifts can be categorized into four types~\citep{kouw2018introduction}, as shown in Fig.~\ref{fig:d}.~Existing work primarily focuses on a single shift only, by assuming that other shifts remain invariant across domains.~The covariate shift w.r.t.~$p(x)$ is to align the marginal distribution for all of the data samples. At a more fine-grained level, the conditional shift is used to align the shift of $p(x|y)$, which is a more realistic setting than the covariate shift only setting, since different classes could have their own shift protocols. For instance, some street lamps glitter, while other lamps are dim at night \citep{liu2021adversarial}. However, estimating $p_t(x|y)$ without $p_t(y)$ is ill-posed \citep{zhang2013domain}. Moreover, the label shift \citep{chan2005word}, a.k.a. target shift, indicates the sample proportion of involved classes is different between two domains.~Furthermore, the concept shift \citep{kouw2018introduction} can arise, when classifying, for example, tomato as a vegetable or fruit in different countries; it is, however, usually not a common problem in popular object classification or semantic segmentation tasks. As such, this review mainly focuses on the covariate shift alignment in UDA, as is most commonly studied. The challenges of aligning the other shifts and their combinations are also discussed as directions for future research. 
 
\section{Methodology}

\begin{figure}[t]
\begin{center}
\includegraphics[width=1\linewidth]{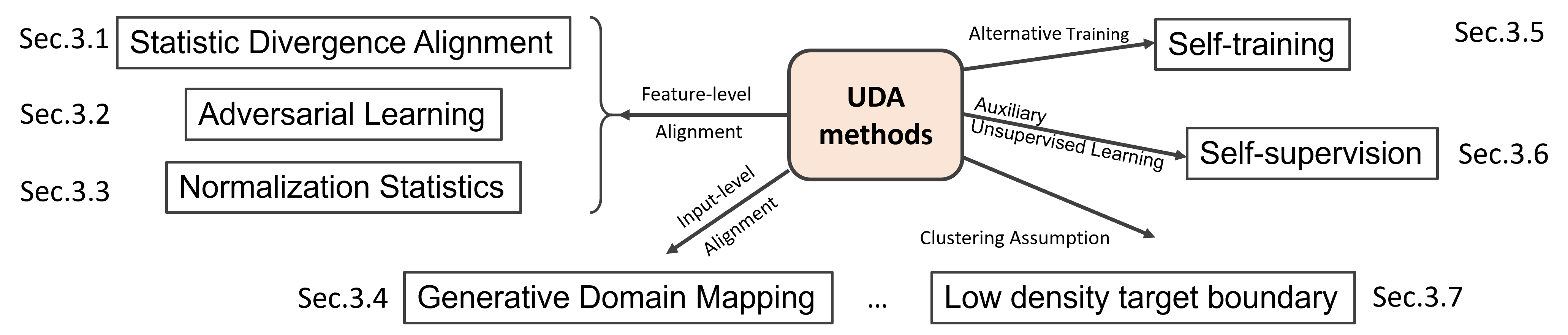}
\end{center} 
\caption{A summary of the main stream UDA methods discussed in this paper.}
\label{fig:udaclass} 
\end{figure}

The past few years have witnessed a proliferation of UDA methods, following the rapid growth of neural network research.~Popular approaches include domain alignment with statistic divergence and adversarial training, generative domain mapping, normalization statistics alignment, ensemble-based methods, and self-training, as summarized in Fig. \ref{fig:udaclass}. In addition, these approaches can be combined to further enhance performance on a variety of tasks. In this section, we discuss each category in more detail as well as their combinations and connections.
 
\subsection{Statistic Divergence Alignment} 
 
Learning domain invariant feature representations is the most widely used philosophy in many deep UDA methods, which hinges on minimizing domain discrepancy in a latent feature space. To achieve this goal, choosing a proper divergence measure is at the core of these methods. Widely used measures include MMD \citep{rozantsev2018beyond}, correlation alignment (CORAL) \citep{sun2016return}, contrastive domain discrepancy (CDD) \citep{kang2019contrastive}, Wasserstein distance \citep{liu2020importance}, graph matching loss \citep{yan2016short}, etc.

\begin{figure}[t]
\begin{center}
\includegraphics[width=1\linewidth]{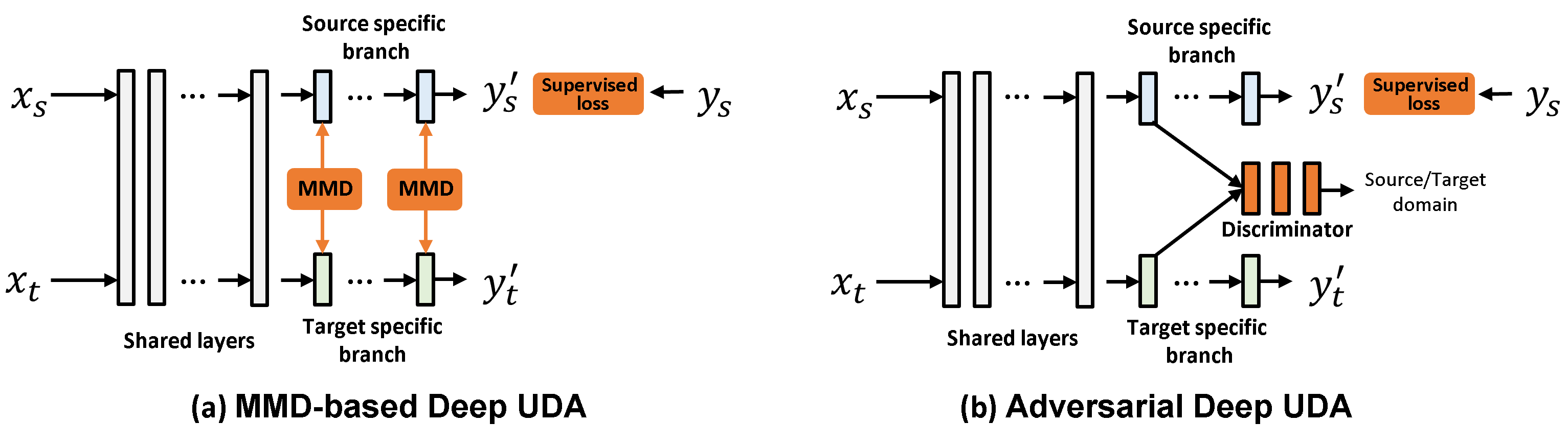}
\end{center} \vspace{-10pt}
\caption{Different architectures of (a) MMD-based UDA (e.g., DAN \citep{long2015learning}), and (b) adversarial training based deep UDA (e.g., domain-adversarial neural networks (DANN) \citep{ganin2016domain}).}
\label{fig:mm} 
\end{figure}

Following the hypothesis of a two-sample statistical test, MMD measures the distribution divergence with observed samples. Specifically, the mean of a smooth function w.r.t. the samples from two domains are compared, where a larger mean difference indicates a larger domain discrepancy. Conventionally, the unit ball in characteristic reproducing kernel Hilbert spaces (RKHS)---as a means of analyzing and comparing distributions---is used as the smooth function, which provides a zero population if and only if the two distributions are equal. In practice, the alignment component serves as another classifier akin to a task classifier. In what follows, MMD can be calculated and minimized between the outputs of the classifiers’ layers \citep{rozantsev2018beyond}, as shown in Fig. \ref{fig:mm}(a). Following vanilla MMD, multiple kernel MMD (MK-MMD) \citep{long2015learning} and joint MMD (JMMD) \citep{long2017deep} are further proposed to achieve a more robust MMD estimation. 

Similar to MMD, CORAL is proposed, based on a polynomial kernel \citep{sun2016return}. CORAL is defined as the difference of the second-order statistics, i.e., covariances, across the features of two domains. To measure the difference of the covariances, different distances have been explored, e.g., squared matrix Frobenius norm \citep{sun2016deep}, an Euclidean distance measure in mapped correlation alignment \citep{zhang2018unsupervised}, log-Euclidean distances \citep{wang2017deep}, and geodesic distances \citep{morerio2017minimal}. CORAL has also been generalized to possibly infinite-dimensional covariance matrices in RKHS \citep{zhang2018aligning}. The statistics beyond the first-order, e.g., MMD, and second-order, e.g., CORAL, are further investigated for more accurate CORAL estimation \citep{chen2020homm}.
 
To achieve class conditioned distribution alignment, CDD \citep{kang2019contrastive} is proposed to incorporate the class label into MMD. By minimizing CDD, cross-class divergence is enlarged, while within-class divergence is minimized. Considering that the label in a target domain is missing in UDA, contrastive adaptation networks (CAN) \citep{kang2019contrastive} is proposed to alternatively estimate the target domain label with clustering, while minimizing CDD. 
 
In addition, the Wasserstein distance \citep{liu2020wasserstein,han2020wasserstein,liu2019conservative,ge2021embedding,liu2019conservative}, a.k.a. optimal transportation distance or earth mover’s distance \citep{liu2020importance,liu2020severity}, could be another alternative to measure the distribution divergence. The joint distribution optimal transport (JDOT) is proposed to measure the Wasserstein distance between two feature distributions \citep{courty2017joint}. As a deep learning framework, DeepJDOT is further proposed to achieve an end-to-end UDA \citep{damodaran2018deepjdot}.  

Furthermore, graph matching has been used as a divergence measure, which aims at finding an optimal correspondence between two graphs \citep{yan2016short}. With a batch of samples, the feature extraction can be regarded as nodes in an undirected graph. The distance between two nodes represents their similarity. The domain divergence is defined as the matching cost between the graphs in source and target domain batches \citep{das2018graph,das2018unsupervised}. 
 
\subsection{Adversarial Learning}  

Instead of choosing a divergence measure, such as MMD, recent work focuses on adaptively learning a measure of divergence. With recent advances in generative adversarial networks (GAN), adversarial training is widely used to achieve domain invariant feature extraction. 

Following Theorem \textcolor{red}{1}, to efficiently minimize the upper bound, i.e., the right-hand side of Eq.~(\ref{e:1}), adversarial UDAs are used to minimize domain divergence at the feature level with guidance of a discriminator as an adaptively learned divergence measure. Specifically, as shown in Fig. \ref{fig:mm}(b), in \cite{ganin2016domain,tzeng2017adversarial}, a feature extractor $f(\cdot)$ is applied onto $x$ to extract a feature representation $f(x)\in\mathbb{R}^K$. We would expect that $d[p_s(f(x)),p_t(f(x))]$ could be a small value. Targeting this goal, in addition to training a classifier $Cls$ to correctly classify source data, $f(\cdot)$ is also optimized to encourage the source and target feature distributions to be similar to each other, following the supervision signal from a domain discriminator $Dis:\mathbb{R}^K\rightarrow(0,1)$. We note that the classifier $Cls:\mathbb{R}^K\times\mathcal{Y}\rightarrow(0,1)$ outputs the probability of an extracted feature $f(x)$ being a class $y$ among $c$ categories, i.e., $ C(f(x),y)=p(y|f(x);Cls)$. The objective of different modules can be {\begin{align}
   &~_{Cls}^{\rm{max}}~~^\mathbb{~~E}_{x\sim p_s} {\rm log} {{C}}(f(x),y) \label{e:2}\\ 
   &~_{Dis}^{\rm{max}}~~^\mathbb{~~E}_{x\sim p_s} {\rm log} (1-Dis(f(x))+^\mathbb{~~E}_{x\sim p_t} {\rm log}Dis(f(x)) \label{e:3}\\
   &~_{~f}^{\rm{max}}~^\mathbb{~~E}_{x\sim p_s} {\rm log} {{C}}(f(x),y)+\lambda^\mathbb{~~E}_{x\sim p_t} {\rm log} (1-Dis(f(x)), \label{e:4}
\end{align}}where $\lambda\in\mathbb{R}^+$ is used to balance between the two loss terms. Following the conventional adversarial UDA methods, the three $max$ strategy \citep{tran2019gotta,salimans2016improved} can be leveraged, and the three objectives above are used to update the corresponding three modules, respectively. In Eq.~(\ref{e:3}), if $f(x)$ is a source domain feature, then $Dis(f(x))$ is trained to produce 0 and vice versa. Note that {maximizing} $~^\mathbb{~~E}_{x\sim p_t} {\rm log}Dis(f(x))$ for $Dis$ in Eq.~(\ref{e:3}), while maximizing $~^\mathbb{~~E}_{x\sim p_t} {\rm log}(1-Dis(f(x)))$ for $f$ in Eq.~(\ref{e:4}) has made this formula a $minmax$ adversarial game. 

\begin{figure}[t]
\begin{center}
\includegraphics[width=1\linewidth]{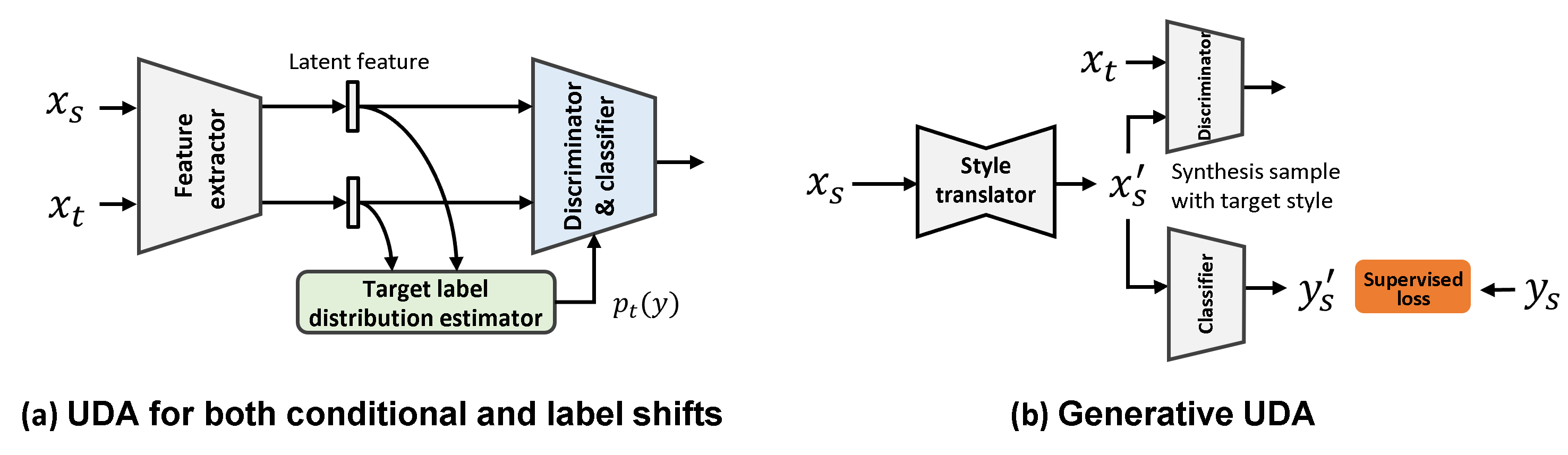}
\end{center}  \vspace{-15pt}
\caption{Different architectures for (a) adversarial UDA with Conditional and Label Shift (CLS) \citep{liu2021adversarial}, and (b) generative adversarial UDA \citep{bousmalis2017unsupervised}.}
\label{fig:adv} 
\end{figure}

Specifically, domain adversarial neural network (DANN) \citep{ganin2016domain} utilizes the gradient reversal layer as a domain discriminator $Dis$. In addition, adversarial discriminative domain adaptation (ADDA) \citep{tzeng2017adversarial} is proposed to initialize the target model with source domain training, followed by adversarial adaptation, which amounts to the target domain-specific classifier.~Other than minimizing cross-entropy based domain confusion losses, \cite{tzeng2015simultaneous} propose to enforce the prediction as a uniform distribution of binary bins. Assuming that the samples are the same, these two domain discriminative losses are essentially equivalent to each other \citep{goodfellow2016deep}. Similarly, \cite{motiian2017few} group the domains and classes as four pairs, by utilizing a four-class classifier for the domain discriminative network. The feature generator is further developed in \cite{volpi2018adversarial} to achieve source domain feature augmentation. 
  
Instead of modeling the domain divergence with the JS divergence as in conventional adversarial UDA \citep{salimans2016improved}, a discriminator for estimating the Wasserstein distance is further proposed \citep{shen2018wasserstein}. Following the recent Wasserstein GAN \citep{adler2018banach}, the Wasserstein distance can be used as a better distance measure, especially to cope with large discrepancies. This is because the JS-divergence cannot differentiate the distance between distributions if there is no overlap between two distributions. In \cite{saito2018maximum}, there are two discriminators to maximize the discrepancy of each class in the target domain, which renders the target domain features to have a wider class-wise boundary region to facilitate the classification.

\subsection{Normalization Statistics} 

\begin{figure}[t]
\begin{center}
\includegraphics[width=1\linewidth]{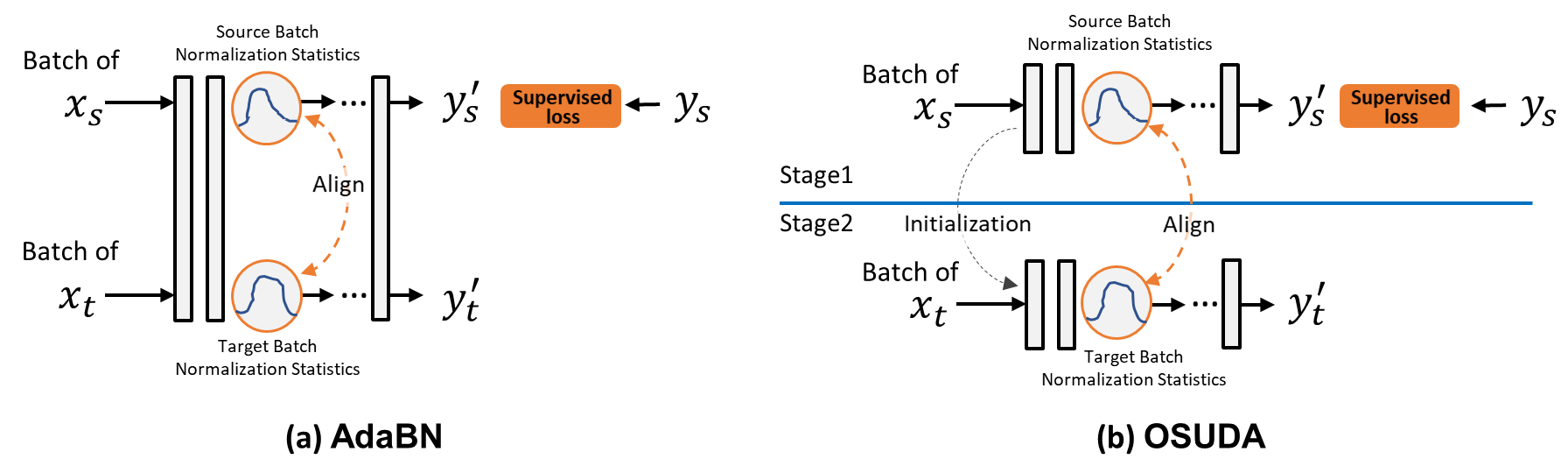}
\end{center} \vspace{-15pt}
\caption{Different architectures for BN-based methods, e.g., (a) AdaBN \citep{li2018adaptive} and OSUDA \citep{liu2021adapting,liu2022Off-the-Shelf}.}
\label{fig:bn} 
\end{figure}

In modern deep neural networks, batch normalization (BN) layers have played an important role in achieving faster training \citep{ioffe2015batch}, smoother optimization, and more stable convergence \citep{wu2018group}, due to its insensitivity to initialization \citep{santurkar2018does}. In each normalization layer, there are two low-order batch statistics, including mean and variance, and two learnable high-order batch statistics, including scaling and bias.  
 
Some early work assumes that the BN statistics of mean and variance inherit domain knowledge. As an early attempt of applying BN to domain adaptation, AdaBN \citep{li2018adaptive}, as illustrated in Fig. \ref{fig:bn}(a), is proposed to achieve UDA, by modulating BN statistics from a source domain to a target domain. In AdaBN, once training is completed, the parameters and weights learned during the training, except for BN layers, are fixed. As a result, BN layers can be simply added to a target domain, without having an interaction with the source domain \citep{li2018adaptive}. In addition, AutoDIAL \citep{carlucci2017autodial} is further proposed as a generalized AdaBN, which retrains the network weights simultaneously with additional domain alignment layers. 

Recent work \citep{chang2019domain,maria2017autodial,wang2019transferable,mancini2018boosting} demonstrates that the low-order batch statistics, including the mean and variance, are domain-specific, because of the divergence of feature representations across two domains. Note that simply forcing the mean and variance to be the same between source and target domains is likely to lose expressiveness of networks \citep{zhang2020generalizable}. Besides, once the low-order BN statistics discrepancy has been partially mitigated, the high-order BN statistics can be shareable between two domains \citep{maria2017autodial,wang2019transferable}. Note that all of the aforementioned approaches \citep{chang2019domain,maria2017autodial,zhang2020generalizable,wang2019transferable,mancini2018boosting} need joint training on source domain data. Recently, OSUDA \citep{liu2021adapting,liu2022Off-the-Shelf}, as shown in Fig.~\ref{fig:bn}(b), is proposed to reduce the domain discrepancy, by means of a momentum-based adaptive low-order batch statistics progression strategy and an explicit high-order BN statistics consistency loss for source-free UDA segmentation.

\subsection{Generative Domain Mapping} 

Rather than aligning features in a latent space, an alternative can be directly rendering the target domain data at the data level. The classifier or segmentation network can be trained on the generated target domain data from source domain data alongside their labels \citep{shrivastava2017learning}. In addition, the network can be trained simultaneously with GANs \citep{bousmalis2017unsupervised,hoffman2018cycada}, as shown in Fig. \ref{fig:adv}(b). 

Cycle reconstruction for image style translation plays an important role in unpaired translation tasks \citep{zhu2017unpaired,kim2017learning,yi2017dualgan}. However, it is challenging to efficiently constrain local structures, thus leading to significant distortions in the translated images and their segmentations \citep{yang2020unsupervised}. To address this, \cite{yang2020unsupervised} extract a modality-independent neighborhood descriptor (MIND) feature $M(x_t)$ and $M(G_{TC}(x_t))$ of $x_t$ and $G_{TC}(x_t)$ with a manually defined extractor $M$, and minimize their reconstruction loss $||M(x_t)-M(G_{TC}(x_t))||_1$. \cite{liu2022structure} propose to learn a general structure feature extractor $f$ in lieu of $M$. To achieve more fine-grained class-wise image mapping, conditional GANs have been widely used for generative domain mapping.

~

\subsection{Self-training}  

Unlike approaches that reduce domain discrepancy with a divergence measure, self-training is proposed as an alternative training scheme, by utilizing unlabeled target domain data to achieve domain adaptation \citep{zou2019confidence}. Self-training is based on a round-based alternative training scheme, which is originally developed for semi-supervised training and has recently been adapted for UDA. There are two steps involved in deep self-training based UDA: (1) creating a set of pseudo-labels in the target domain, and (2) retraining the network using the generated pseudo-labels with target domain data.

Recently, self-training-based approaches have surpassed adversarial training-based approaches in several deep UDA tasks \citep{wei2021theoretical,mei2020instance,shin2020two}. Whereas self-training was initially presented as part of semi-supervised learning \citep{triguero2015self}, recently proposed deep self-training methods combine feature embedding with alternative learning in a unified manner, thus yielding flexible domain adaptation \citep{zou2019confidence}. 

A crucial issue in self-training-based approaches, however, is that pseudo-labels in the target domain could be noisy; and thus it is likely that a large proportion could be unreliable. To mitigate this issue, selecting the prediction with high confidence is essential. To this end, in classification or segmentation tasks with softmax output unit, a possible solution would be to gauge the confidence as the maximum value of histogram \citep{zou2019confidence}. Additionally, to tackle the problem of the noisy and unreliable pseudo-labels, \cite{zou2019confidence} propose to construct a more conservative pseudo-label in order to smooth the one-hot hard label to a soft label vector. \cite{liu2021energy} further resort to an additional supervision signal of an energy-based model for regularization, which is independent of the pseudo-label. \cite{ke2020instance} propose to explore instance-wise self-training for UDA.
 
In addition to the discriminative tasks, such as classification and segmentation, \cite{liu2021generative} further extend self-training to a generative task, by controlling the confident pseudo-label of continuous pixel value with a Bayesian uncertainty mask. In learning-based tasks, two kinds of uncertainty exist, including an aleatoric uncertainty and an epistemic uncertainty \citep{der2009aleatory,kendall2017uncertainties,hu2019supervised}. Specifically, the aleatoric uncertainty is caused by the uncertainty from noisy training data observations, whereas the epistemic uncertainty is caused by models that are not sufficiently trained. In self-training, the pseudo-labels are typically noisy, thus leading to the aleatoric uncertainty. In addition, the epistemic uncertainty in self-training is caused by a limited number of iterations for model training and a limited number of target domain training samples. Therefore, taking both uncertainties into account is vital to build a robust model with a holistic uncertainty calibration. 

\subsection{Self-supervision}   

Another solution to UDA is to incorporate auxiliary self-supervision tasks into the network training. Self-supervised learning hinges on only unlabeled data to prescribe a pretext learning task, such as context prediction or image rotation, for which a target objective can be computed without supervision \citep{kolesnikov2019revisiting}. This group of work assumes that alignment can be achieved by carrying out source domain classification and reconstruction of target domain data \citep{ghifary2016deep} or both source and target domain data \citep{bousmalis2016domain}. In \cite{ghifary2016deep}, a deep reconstruction-classification network is optimized with a pair-wise squared reconstruction loss. In particular, the scale-invariant mean squared error reconstruction loss is introduced in \cite{bousmalis2016domain} to train its domain separation networks. 

In addition to the conventional reconstruction tasks \citep{liu2018data}, new self-supervision tasks have been proposed, e.g., image rotation and jigsaw predictions \citep{xu2019self}. \cite{kim2020cross} propose both in-domain and across-domain self-supervision to achieve UDA with fewer source domain labels. \cite{lian2019constructing} propose a self-motivated pyramid curriculum for segmentation.

\subsection{Low density target boundary} 

Several UDA approaches based on a popular clustering assumption \citep{chapelle2005semi} are proposed in the context of semi-supervised training, which indicates target domain samples from the same class are likely to be distributed closely as a cluster. The target domain class-wise decision boundaries should be located in the low-density regions \citep{he2020image2audio}. To this end, \cite{shu2018dirt} propose virtual adversarial domain adaptation. In addition, after training, a decision-boundary iterative refinement step with a teacher is further applied to refine the decision boundary in a target domain \citep{shu2018dirt}. \cite{kumar2018co} propose to combine variational adversarial training with a conditional entropy loss to achieve a low-density boundary and avoid overfitting in unlabeled data. Similarly, an entropy loss has been applied to AutoDIAL \citep{carlucci2017autodial}. Other than the feature level, generative methods at the image level have also been developed to make the decision boundary lie in a lower density region \citep{wei2018generative}.

\cite{saito2017adversarial} propose adversarial dropout regularization, which is seen as the difference between two dropout networks as a discriminator to generate target discriminative features. \cite{lee2019drop} extend the adversarial dropout for convolutional layers with a channel drop rather than an element drop. TDDA \citep{gholami2019taskdiscriminative} focuses on task-discriminative alignment for UDA.

\subsection{Other Methods} 

DEV \citep{you2019toward} is proposed to achieve UDA via model selection. The prototype with clustering is utilized in \cite{pan2019transferrable} for class-wise adaptation. \cite{liu2021subtype,liu2022subtype} further extend the class-wise prototype to fine-grained subtypes. \cite{wu2020dual} propose to apply dual mixup regularization to adversarial UDA. Domain randomization is proposed in \cite{rodriguez2019domain,kim2019diversify} to randomly generate source domain data with a different style to achieve a decent generalization ability in a target domain. To further utilize unlabeled data, a mean teacher has been used in \cite{cai2019exploring,deng2021unbiased}. The inter/intra object correlation is explored in a graph reasoning framework for domain adaptation in \cite{xu2020cross}. \cite{liu2022self} propose to utilize the self-semantic contour as an intermediate feature to facilitate domain adaptation.

\subsection{Combinations and Connections}  

Several aforementioned approaches can be combined with each other to exploit complementary optimization. Both feature-level adversarial alignment and image-level generative mapping can be combined sequentially, e.g., GraspGAN \citep{bousmalis2018using}, or jointly, e.g., CyCADA \citep{judy2018cycada}. Following AdaBN, several works have shown that the BN alignment can be added on top of other UDA methods \citep{li2018twin,bousmalis2018using,french2017self,kang2019contrastive}. The BN alignment and entropy minimization for low-density target boundary are combined for source data free UDA \citep{liu2021adapting,liu2022Off-the-Shelf}. Adversarial domain-invariant feature alignment has been applied on different levels, following an ensemble scheme \citep{kumar2018co}. Low-density target boundary and domain-invariant feature learning are jointly learned in \cite{lee2019sliced,saito2018maximum}. \cite{kang2018deep} combine generative image mapping with the alignment of model attention. PANDA \citep{hu2020panda} integrates  adversarial training with prototype-based normalization. 

\section{Applications} 
 
UDA has been successfully applied to a variety of application areas, including perception and understanding of images, video analysis, NLP, time-series data analysis, medical image analysis, and climate and geosciences. While some works are based on general principles of UDA, other works are targeted to tackle specific applications under consideration, by exploiting the characteristics of training and testing datasets. In this section, we do not intend to provide a comprehensive review, but rather opt to highlight examples of trends in UDA for various application areas, given the presence of a huge body of work and a number of excellent prior reviews.

\subsection{Image Analysis} 

Natural image analysis is the most explored area in UDA, due to the availability of large-scale visual databases. Depending on the label and corresponding output, popular tasks include image classification, e.g., object recognition and face recognition, object detection, semantic segmentation, image generation, image caption, etc. 

\subsubsection{Image Classification}

Classification or recognition of object categories has been a fundamental task in computer vision. As such, numerous attempts have been made to use deep learning and UDA for the classification. For instance, \cite{long2017deep} use AlexNet \citep{krizhevsky2012imagenet} backbone for the task, where the approach is compared against the source model, the DANN method \citep{ganin2015unsupervised}, and the variations of MMD, e.g., DDC \citep{tzeng2014deep}, DAN \citep{long2015learning}, JAN \citep{long2017deep} and RTN \citep{long2016unsupervised}. \cite{zellinger2017central} compare their CMD methods with other discrepancy-based methods, e.g., DDC \citep{tzeng2014deep}, deep CROAL \citep{sun2016deep}, DLID \citep{chopra2013dlid}, AdaBN \citep{li2016revisiting}) and adversarial DANN \citep{ganin2015unsupervised}. In addition to the object classification, UDA of face recognition is another hot research topic, in which the most important shifts include pose, illumination, expression, age, ethnicity, and imaging modality \citep{liu2021identity,liu2021mutualpr,liu2017adaptive}. Among these shifts, the expression, ethnicity, and imaging modality have discrete variations, while other attributes have continuous variations \citep{liu2019feature,liu2021mutualpami}. In \cite{kan2015bi}, a bi-shifting auto-encoder framework is proposed  for face identification with the domain shifts of view, ethnicity, and sensor. \cite{hong2017sspp} generate different face views for domain adaptation. \cite{sohn2017unsupervised} propose to achieve adversarial UDA for video face recognition. 

There are several widely adopted benchmarks for classification tasks. As for databases to test the domain shift in natural images, Office-31 dataset \citep{saenko2010adapting} is widely used, which contains data from three different sources, i.e., Amazon (A), DSLR (D), and Webcam (W). As for image synthesis to achieve real image domain adaptation, VisDA17 \citep{visda2017} is a preferred choice. DomainNet \citep{peng2019moment} is the largest domain adaptation dataset to date, which consists of $\sim$0.6M images with 345 sub-classes from 24 meta-classes.

In addition to classification with discrete labels, several tasks have ordinal class labels \citep{liu2019unimodal,liu2020unimodal}. In the case of medical diagnosis, it is likely that the labels are discrete and distributed successively.~As such, UDA for ordinal classification needs to induce a non-trivial ordinal distribution first, prior to projecting the data onto a latent space. In \cite{liu2021recursively,liu2022recursively}, a recursively conditional Gaussian distribution is adapted to ordered constraint modeling, which admits a tractable joint distribution prior.
 
\subsubsection{Image Detection}

In addition to recognizing objects, image detection has been further investigated, by localizing objects in a wide view of field with a bounding-box \citep{oza2021unsupervised}. Deep object detection has been an integral part of several tasks, e.g., surveillance, augmented/mixed reality (AR/MR), autonomous driving, and human-computer interface.

Adversarial feature alignment has been utilized for UDA object detection in \cite{chen2018domain,saito2019strong,sindagi2020prior,hsu2020every,vs2021mega}. In addition, adversarial generative mapping at the image level has been applied in \cite{zhang2019cycle,rodriguez2019domain,hsu2020progressive,chen2020harmonizing,yu2022sc}. In \cite{roychowdhury2019automatic,khodabandeh2019robust,kim2019self,zhao2020collaborative,li2021cross,gu2022ossid}, the pseudo-label based self-training is adopted for progressive adaptation.

For UDA in image detection, popular domain adaptation scenarios include adaptation of cross weather conditions, synthetic to real imagery, etc. For example, domain adaptation is performed from Cityscapes \citep{cordts2016cityscapes} to Foggy Cityscapes \citep{sakaridis2018semantic}, which is rendered from Cityscapes, by adding the fog noise. In addition, several works \citep{xu2019wasserstein} use SIM10k dataset as the source domain and the Cityscapes dataset as the target domain.

\subsubsection{Image Segmentation}

Image segmentation aims at pixel-wise classification \citep{tajbakhsh2020embracing,liu2022variational}. Rather than indicating the rough position of the object like a detection task, segmentation provides fine-grained delineation to support subsequent operations. Compared with the sample wise classification in UDA, it is difficult to apply the low-density target region and prototype based UDA methods. Since each pixel needs to be represented as a point in the feature space, it is difficult to scale up to large-scale data. Instead, adversarial training at both feature and image levels have been widely used \citep{yu2015multi,Tsai_adaptseg_2018,ke2020instance}. Self-training based methods have also been developed for semantic segmentation \citep{chen2019synergistic,liu2021domain}.

A typical task is to adapt the large-scale labeled game engine rendered data, i.e., GTA5 \citep{richter2016playing}, to the real-world data, i.e., Cityscapes \citep{cordts2016cityscapes}, for which there are a total of 19 shared labels for semantic segmentation. In a source domain, there are a total of 24,000 labeled game engine rendered images from Grand Theft Auto 5. As the standard evaluation protocol \citep{yang2020adversarial}, all of the samples in the GTA5 dataset are used as the source domain, while the training set of Cityscape with a total of 2,975 images is used as the target domain training set. The testing set of Cityscapes has a total of 500 images.

\subsubsection{Image Generation}

Generative models have been widely applied to diverse tasks, e.g., entertainment, image harmonization/stylization, and data completion and augmentation~\citep{yang2018image,yang2019towards,liu2021dual,liu2021unified,wang3916770advanced,xing2022brain}. 

The image style synthesis task itself can be regarded as a cross-domain translation task, when the input involves another style or domain. To address this, GAN-based methods have been widely used for cross-domain image generation tasks~\citep{he2021autoencoder}. Self-training has also been applied to the image synthesis task. For example, \cite{liu2021generative} propose to leverage self-training for the image synthesis task, which also considers both epistemic and aleatoric uncertainties \citep{der2009aleatory}. Specifically, \cite{liu2021generative} aim at cross modality synthesis using paired sets of images acquired from two different sites.

\subsection{Medical Image Analysis}  

Medical image analysis has been a major application ground for image analysis methods, due to its wide usage in real-life imaging problems. In addition, a variety of imaging modalities are used in a clinical setting, each of which poses unique challenges. An increasing amount of deep network-based methods have been proposed to achieve enhanced computational speed and better algorithmic performance over traditional medical image analysis methods. UDA has been successfully adopted in image segmentation, classification, and generation tasks in addition to a few other varying applications.

\cite{perone2019unsupervised} use a self-ensembling technique in semantic image segmentation, demonstrating that it can improve model generalization. Their method is evaluated using a small number of magnetic resonance imaging (MRI) datasets, serving as a proof-of-concept of the advantage of UDA in medical imaging, rather than showing an actual application in real medical problems. \cite{ouyang2019data} report UDA for multi-domain medical image segmentation via a VAE-based feature prior matching, which features data efficiency. It is applied to a multi-modality cardiac image dataset to achieve segmentation. \cite{zou2020unsupervised} propose UDA with the so-called Dual-Scheme Fusion Network, where both source-to-target and target-to-source connections are built to help bridge the gap between domain differences for improved performance. It is applied to the segmentation of both brain tumors and cardiac data, yielding decent results. \cite{he2020adversarial} propose to achieve cross-device retinal OCT segmentation. \cite{liu2022self} propose to facilitate cross-modality brain tumor segmentation with self-semantic contouring. Additionally, more methods have been proposed to improve the segmentation UDA networks from within their structures. To enable flexibility of two-way adaptations, \cite{ning2021new} propose a bidirectional UDA framework based on disentangled representation learning. It achieves decent performances in both the forward adaptation direction, from MRI to computed tomography (CT), and the backward direction, from CT to MRI. The popular evaluation setting is to use the MMWHS challenge dataset as the source domain and the MMAS dataset as the target domain, respectively \citep{zhuang2016multi}. 
 
It often poses a challenge to share medical data for collaboration due to sensitive patient information. To address the privacy concern of the large-scale and well-labeled medical data in the source domain, \cite{liu2021adapting,liu2022Unsupervised} propose to adapt a pre-trained ``off-the-shelf" segmentation model without source domain data at the adaptation stage. The test-time adaptable segmentation networks have been developed to achieve UDA in a source-free manner \citep{he2021autoencoder,karani2021test}. In addition, \cite{he2021autoencoder} show that their method can be generalized to image translation UDA tasks.

Besides, recent years have seen increased usage of UDA to solve segmentation problems using a variety of imaging modalities, such as CT, MRI, X-ray imaging (\cite{zhang2018task}) and optical coherence tomography imaging \citep{wang2021unsupervised,li2021unsupervised}. Additionally, UDA has been used in medical image classification \citep{ahn2020unsupervised,mahapatra2021gcn} and diagnosis \citep{zhang2020collaborative}. For instance, \cite{liu2021subtype,liu2022subtype} propose to explore the subtype of congenital heart disease \citep{wang2021automated}. The disease level has been investigated in \cite{liu2021recursively,liu2022recursively}, in which the Kaggle Diabetic Retinopathy (KDR)\footnote{\url{https://www.kaggle.com/c/diabetic-retinopathy-detection}} is used as the source domain, and the recent Indian Diabetic Retinopathy Image Dataset (IDRiD) dataset \citep{porwal2018indian} is used as the target domain. 

The evaluation database used in \cite{he2021autoencoder} for T1-weighted to T2-weighted MRI translation is brain MRI datasets from three IXI centers\footnote{\url{https://brain-development.org/ixi-dataset/}}. In addition, paired cine and tagged tongue MRI data in two clinical sites datasets are used in \cite{liu2021generative}.

\subsection{Video Analysis} 

\begin{figure}[t]
\begin{center}
\includegraphics[width=\linewidth]{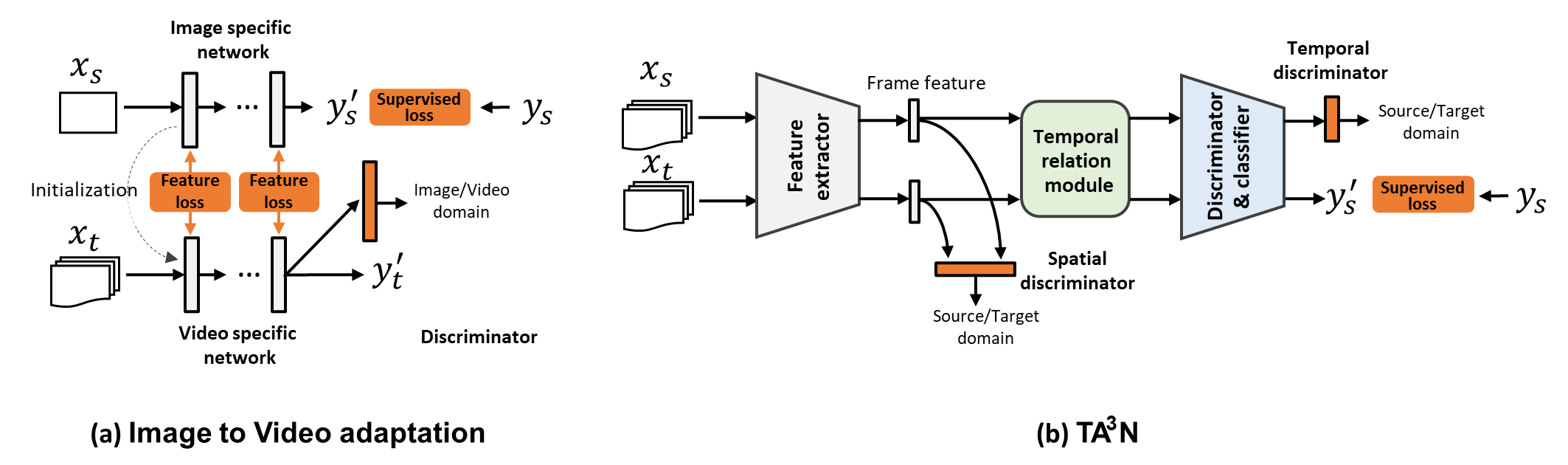}
\end{center} 
\caption{(a) Example of image-to-video adaptation \citep{sohn2017unsupervised} and (b) example of UDA for video analysis \citep{chen2019temporal}.}
\label{fig:video} 
\end{figure}

Video data contain rich spatial and temporal semantic information. However, it is challenging to collect and annotate a large volume of video data to learn useful spatiotemporal features. Annotation of all video frames is labor-intensive and time-consuming for different target applications and devices (\cite{li2019unsupervised, peng2019discriminative, saleh2019domain}). Accordingly, UDA has been applied to video analysis tasks, including action recognition (\cite{chen2019temporal,choi2020shuffle,pan2020adversarial,chen2022multi}), person re-identification (\cite{mekhazni2020unsupervised}), action segmentation (\cite{chen2020action, chen2020action_mixed}), video captioning (\cite{chen2021mind}), video quality assessment \citep{chen2021unsupervised}, and video artifact reduction \citep{ham2021training}.

Because there are few well-organized video datasets in early work, an image-to-video adaptation method is proposed to use a large-scale image dataset to train a model for video analysis. \cite{sohn2017unsupervised} improve accuracy in face recognition using a video through image-to-video domain adaptation as in Fig. \ref{fig:video}(a). They attempt to overcome the difference of visual quality between still images and video frames. \cite{liu2019deep} propose a deep image-to-video adaptation and fusion network (DIVAFN) to enhance accuracy in video action recognition, by transferring knowledge learned from images. In addition, UCF-HMDB$_{full}$ and Kinetics-Gameplay \citep{chen2019temporal} have been collected to promote video domain adaptation and benchmark the performance in the presence of large domain discrepancy. 

Most pre-trained networks for video analysis tend to perform poorly, when a pre-trained model encounters unseen temporal dynamics on the target side. There is prior work to resolve the problems in video action recognition, by overcoming domain discrepancies along the spatial and temporal directions.  \cite{chen2019temporal} propose a temporal attentive adversarial adaptation network (TA$^3$N) in Fig. \ref{fig:video}(b). They attempt to align two domains spatio-temporally, by encoding spatio-temporal features using an attention mechanism. \cite{choi2020shuffle}, \cite{pan2020adversarial}, and \cite{chen2022multi} improve the attention mechanism for better alignment. Video UDA on action recognition is extended to more realistic settings, using videos collected from surveillance cameras \citep{mou2021unsupervised} and drones \citep{choi2020unsupervised}. 

UDA is actively studied for video scene analysis and restoration.
\cite{chen2020generative} propose VideoGAN to focus on the translation of video-based data and transfer the data across different domains. \cite{guizilini2021geometric} present a video segmentation method using self-learning to bridge a domain gap between simulated and real videos. UDA is also applied to face recognition \citep{ekladious2020dual}, person re-identification \citep{mekhazni2020unsupervised}, and video captioning \citep{chen2021mind}. In addition to the analysis of high-level semantics, there is prior work for UDA in low-level video processing. \cite{chen2021unsupervised} and \cite{ham2021training} present various UDA methods for video quality assessment and video artifact reduction. They attempt to provide reliable performance of a network, when the visual quality of a video frame is different between source and target domains. 

The typical evaluation datasets for image-to-video adaptation include
UCF-Olympic, UCF-HMDBsmall, UCF-HMDBfull, and Kinetics-Gameplay \citep{chen2019temporal}. In addition, the Kinetics and NEC-DRONE datasets are utilized for evaluation of video UDA on action recognition \citep{choi2020unsupervised}. For the video quality assessment, UDA approaches are evaluated on DIV2K, BSD68, and Set12 datasets~\citep{ham2021training}.

\subsection{Natural Language Processing}  

Similar to the visual data processing, the necessity of developing UDA methods has emerged in NLP~\citep{sharir2020cost}, partly because it is costly and demanding to annotate the sheer volume of language data. 

Sentiment analysis is the most explored application to develop UDA methods in NLP \citep{ramponi2020neural}. In early attempts of UDA in NLP, \cite{ganin2016domain} propose a domain-adversarial neural network (DANN). UDA is carried out by adding a domain classifier that is connected to a feature extractor through a gradient reversal layer. It has motivated several studies \citep{li2018s, shen2018wasserstein, rocha2019comparative, ghosal2020kingdom}. \cite{shen2018wasserstein} utilize the adversarial training to minimize the estimated Wasserstein distance between source and target samples. \cite{rocha2019comparative} indicate that the adversarial training method can be more effective, when the source and target language datasets contain several content variations in addition to the language shift. Furthermore, UDA methods are applied to perform various NLP tasks, including dependency parsing \citep{sato2017adversarial, rotman2019deep}, POS (part-of-speech) tagging \citep{desai2019adaptive, lim2020semi}, relation extraction \citep{fu2017domain, rios2018generalizing, shi2018genre}, trigger identification \citep{naik2020towards}, language identification \citep{li2018s}, political data classification \citep{desai2019adaptive}, etc. 
 
Pre-training has become a key ingredient to deploy an NLP model due to the inherent complexity of the structure of language and the nature of NLP tasks \citep{sharir2020cost, ramponi2020neural}. In recent NLP studies, it is a standard training strategy to fine-tune a transformer-based model with a small amount of data for a target application. A large-scaled language dataset is used for pre-training in the source domain, and task-specific data become the target domain in the context of UDA. With the domain shift, adaptive pre-training has been proposed to compensate for the classical pre-training, such as BERT \citep{burstein2019proceedings}. AdaptBERT \citep{han2019unsupervised} performs domain-adaptive fine-tuning to adapt contextualized embedding by masked language modeling from the target domain. \cite{gururangan2020don} propose to use both domain-adaptive pre-training and task-specific pre-training methods.

Image captioning is an interdisciplinary area to connect computer vision and NLP. A typical solution to UDA for image captioning would be to leverage a convolutional encoder for extracting the necessary latent information of visual scenes, followed by adopting a text generator, e.g., recurrent neural networks. Similarly, \cite{chen2017show} propose to use adversarial training for the paired source domain data and unpaired target domain data. \cite{zhao2017dual} develop a dual learning scheme to fine-tune a source domain model trained on a limited dataset to the target domain. Because the output of an image captioning model is a sentence, it poses a challenge to model a conditional distribution. A possible solution would be to encode a sentence label with an additional recurrent neural network as in \cite{che2021deep}.

The sentiment classification UDA, across English, Chinese, and Arabic with the dataset in \cite{chen2018adversarial}, is used for evaluation \citep{rocha2019comparative}. The English OntoNotes 5.0 and the Universal Dependencies datasets are used for dependency parsing UDA evaluation \citep{rotman2019deep}. In addition, the English portion
of ACE2005 dataset is used for relation extraction UDA evaluation, which covers a total of 6 genres and 11 relation types.

\subsection{Time Series Data Analysis}  
\begin{figure}[t]
\begin{center}
\includegraphics[width=\linewidth]{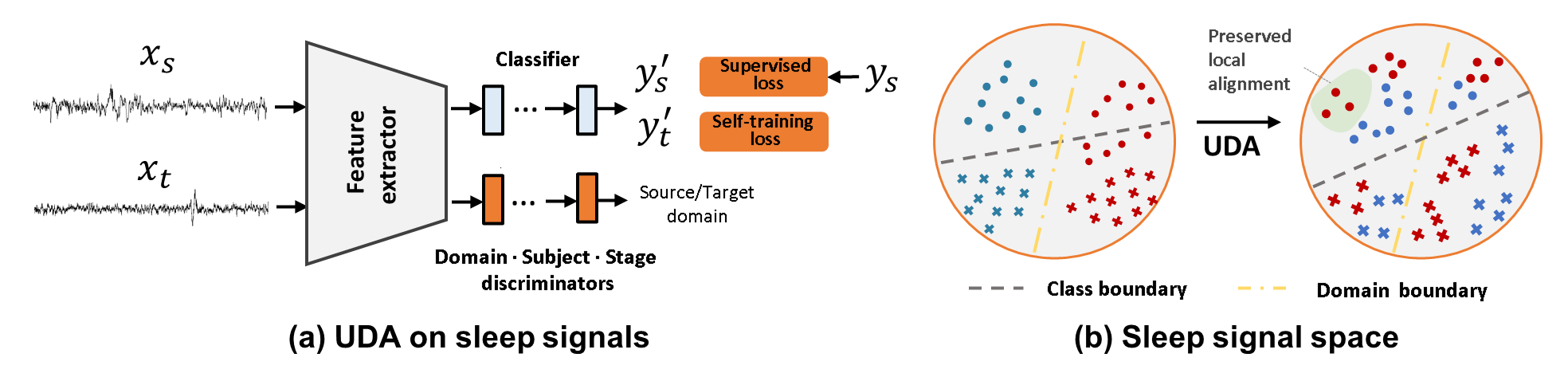}
\end{center} 
\caption{(a) Example of UDA for sleep classification \citep{yoo2021transferring} and (b) sleep signal space.}
\label{fig:sleep} 
\end{figure}
Various UDA strategies are exploited for tasks using time series data. Among others, with time series medical data, such as electroencephalogram (EEG), electrocardiogram (ECG), and multivariate healthcare data, UDA has been applied to perform sleep classification \citep{zhao2021unsupervised, yoo2021transferring, fan2022unsupervised}, arrhythmia classification \citep{niu2020deep, wang2021inter}, motor imagery \citep{raza2019bagging, tang2020conditional}, etc. Especially, these methods attempt to tackle the distribution discrepancy between different datasets and between subjects, because medical data vary depending on demographic features such as age, sex, and illness. For example, \cite{yoo2021transferring} apply both adversarial training and self-training with three different domain discriminators, including domain, subject, and stage discriminators, as shown in Fig.~\ref{fig:sleep}(a), to preserve local structures of sleep stages as shown in Fig.~\ref{fig:sleep}(b).

Existing work on emotion recognition \citep{li2018bi, yin2020speaker,he2021online, he2022adversarial}, speech recognition \citep{wang2018unsupervised, manohar2018teacher, khurana2021unsupervised, anoop2021unsupervised}, and imagined speech recognition \citep{jimenez2021standardization} also brings the concept of UDA. Moreover, the effectiveness of UDA is explored for applications that use industrial time series data, including human action recognition \citep{jiang2018towards,chang2020systematic, du2019unsupervised,sanabria2021unsupervised}, inertial tracking \citep{chen2019motiontransformer}, driving maneuver prediction \citep{tonutti2019robust}, anomaly detection \citep{michau2021unsupervised}, fault diagnosis \citep{lu2021new}, and lifetime prediction \citep{da2020remaining,ragab2020contrastive}.

Besides, time-series UDA approaches are developed to effectively capture the temporal dependencies of time series data that may be neglected, by visual data-based methods. For instance, based on DANN, recurrent domain adversarial neural network (R-DANN) and variational recurrent adversarial deep domain adaptation (VRADA) \citep{Purushotham2017VariationalRA} are proposed by exploiting the long short-term memory (LSTM) network \citep{hochreiter1997long} and variational RNN \citep{chung2015recurrent} as a feature extractor, respectively. More models, such as a sparse associative structure alignment (SASA) model \citep{cai2021time} and a convolutional deep domain adaptation model for time series data (CoDATS) \citep{wilson2020multi}, are developed to improve time series UDA performance.

For sleep signal UDA, the Montreal Archive of Sleep Studies (MASS) is used as the source domain, while the Sleep-EDF database and Sleep-EDF-st database are used as the target domain \citep{yoo2021transferring}. For emotion recognition, the DEAP dataset and DREAMER dataset are usually used as the benchmarks \citep{he2022adversarial}.

\subsubsection{Climate science and Geosciences}

In recent years, deep learning has been applied to numerous applications on the Earth science, e.g., climate science and geosciences \citep{camps2021deep}. Similar to the other application areas, the perception of remote sensing data can also have the problem of domain shift, across location and time. In \cite{huang2020deep}, UDA across active and passive satellite data is developed for cloud type detection. Notably, the active spaceborne Lidar sensor CALIOP onboard CALIPSO satellite has better representation capability and sensitivity to aerosol types and cloud phases, while the passive spectroradiometer sensor VIIRS onboard Suomi-NPP satellite has wide swaths and better spatial coverage. \cite{mengqiu2022sea} propose a UDA method to bridge the gap between the abundant labeled land fog data and the unlabeled sea fog data for sea fog detection. \cite{soto2020domain} exploit the cycleGAN-based UDA approach~\citep{zhu2017unpaired} for deforestation detection in the Amazon forest. 

In addition, UDA has been widely explored in many applications on geoscience research. \cite{nasim2022seismic} investigate a UDA approach to mitigate the domain gap between seismic images of the F3 block 3D dataset from offshore Netherlands and Penobscot 3D survey data from Canada, which utilizes the EarthAdaptNet to semantically segment the seismic images, when a few classes have data scarcity. The teacher-student network has been used in \cite{hu2020unsupervised} for the classification of the Sentinel-2 images across cities, e.g., Moscow and Munich. \cite{lucas2020unsupervised} conduct experiments on Satelite Image Time Series (SITS) classification using existing natural image-based UDA methods and find that those UDA methods are ineffective, due to the temporal nature of SITS. \cite{nyborg2022timematch} propose an explicit UDA method that learns the temporal shift of SITS for crop classification and introduce a dataset for cross-region adaptation from SITS in four different regions in Europe. \cite{ma2021multi} introduce a UDA approach for corn-yield prediction using time-series vegetation indices and weather observations.

\section{Promising Directions}

As stated above, advanced deep UDA methods have been widely applied to numerous tasks and applications.~In this section, we point to a number of underexplored areas that are of great theoretical and practical importance, which can be promising future research directions.

\subsection{Realistic Shift Assumption}   
 
Most of the current UDA methods have focused on the alignment of covariate shift. As analyzed in Sec.~1, however, there exist four kinds of possible shifts in real-world applications~\citep{kouw2019review}. While numerous works are proposed in the literature to address conditional or covariate shifts, label and concept shifts have not been investigated extensively. Notably, approaches for the adversarial feature alignment of the covariate shift and approaches without considering the conditional shift have been outperformed by several competing approaches, e.g., self-training \citep{zou2019confidence}, dropout \citep{saito2017adversarial}, and moment matching methods \citep{pan2019transferrable} in most of the benchmarks. As such, it is important to incorporate both conditional and covariate shifts, as it is ill-posed to take one of them into consideration \citep{zhang2013domain,kouw2018introduction}. In \cite{liu2021adversarial}, theoretical analysis and methodology under the conditional and label shift assumptions are discussed in adversarial learning-based UDA. 

It is, therefore, necessary to incorporate more realistic assumptions of the domain shifts, depending on real-world tasks at hand.

\subsection{Partial/Open-set Domain Adaptation}   

Partial UDA can be seen as a special category of label shifts, in which some classes have zero probability in a target domain.~Due to the mismatch of categories between source and target domains, conventional UDA approaches may result in negative transfer \citep{cao2018partiala,cao2018partialb,kim2020associative}. Similarly, open-set UDA and universal UDA are presented under the assumption that there are novel classes in a target domain; this approach thus could lead to novel class discovery or out-of-distribution detection~\citep{panareda2017open}.

\cite{lipton2018detecting} propose a test distribution estimator to detect the label shift. \cite{azizzadenesheli2019regularized} introduce a regularization approach to correct the label shift. \cite{chen2018re} cast the problem of the label shift as an optimal transportation-based UDA task, which is closely related to the class imbalance problem in the MMD framework. \cite{wu2019domain} propose an asymmetrically-relaxed alignment approach using the adversarial UDA. However, these approaches assume that there is no conditional shift.

As noted above, partial UDA can be regarded as a special case of the label shifts. Therefore, developing more general label shift UDA methods for both small label distribution shift and partial UDA can be more practical for real-world applications. In addition, novel class/subtype discovery could be incorporated into the open-set UDA.

\subsection{Source-free domain adaptation} 

Data privacy has been a critical concern over cross-center collaboration, especially in the medical domain. Conventional UDA requires the large scale and well-labeled source domain data to be shared, which may cause issues over source domain data leakage and intellectual property \citep{bateson2020source}. To address this, \cite{liu2021adapting,liu2022Off-the-Shelf} propose a source-free UDA approach with white-box domain adaptation to delineate anatomic structures in medical imaging data. Specifically, that work leverages an off-the-shelf pre-trained segmentation model to adapt to a target domain, by migrating its batch normalization statistics. In addition, recently, \cite{yin2020dreaming} propose a deep inversion technique to demonstrate that original training data can be recovered from knowledge used in the course of white-box domain adaptation~\citep{zhang2021unsupervised}. To address this, a recent work~\citep{liu2022Unsupervised} uses black-box UDA segmentation, for which no prior knowledge of network weights is needed for adaptation. \cite{liu2022frontier} further propose that a target domain network structure could be different from a trained source domain model to achieve UDA for segmentation.

Source-free domain adaptation is also closely related to test time adaptation, in that we encounter a single  or a few test samples that are different from source domain data \citep{royer2015classifier,fredericks2014towards,wang2020tent,liu2021generative}. To accommodate continuously changing environments, we expect that frameworks employed would be source-free, involve low-cost training in mobile devices, and avoid catastrophic forgetting \citep{hoffman2014continuous,wulfmeier2018incremental,wu2019ace,mancini2019adagraph}.

Therefore, UDA under a more strict data sharing setting can be a promising direction, which only shares the pre-trained white/black-box source domain model. We note that the model sharing is also related to the federated learning, which is another important transfer learning problem \citep{yang2019federated}.

\subsection{Continuous and test time adaptation} 

Existing work on UDA usually assumes that several stationary domains exist for which prior work attempts to achieve domain adaptation between discrete distributions. In real-world environments, however, the change in distributions could be continuous. For example, when one drives from Seattle to Boston, one will cross snow-capped mountains, deserts, plateaus, flatlands, hilly areas, etc. There is, however, no distinct boundaries between these environments, and thus the shift is smoothly evolving. Therefore, one needs to consider lifelong learning to progressively adapt a trained model to new environments \citep{liu2021Generalization}.

There is a need to build well organized and gradually changing UDA datasets. In addition, the mixup or interpolation technology \citep{liu2018data} would be useful to hallucinate the intermediate data between two largely different domains to facilitate UDA.

\subsection{Adaptation in Foundation Model Era} 

Foundation models \citep{bommasani2021opportunities} are recently surged as a hot topic to utilize super large labeled data, which can incorporate sufficiently variant data. In addition, they are robust to the covariate shift in many cases. Then, one can ask: if we have a sufficiently large training set with diverse data distributions, can they generalize well on all of the implementation scenarios? Though applying domain generalization may address the covariate and conditional label shifts, it is challenging to alleviate the label shift, without access to target domain data. In addition, the concept shift can also cause a problem, even though there are sufficient training data.  

UDA methods to deal with label shift can be an important direction in the era of foundation models. In addition, it is interesting to investigate the generality of different foundation models.

\subsection{Semi-supervised Domain Adaptation}  
 
While there have been great advances in UDA, due to diverse target domains, the performance of UDA is not satisfactory in many cases \citep{liu2022ACT}. In such circumstances, labeling a small set of target domain data could be a viable solution \citep{van2020survey}. Along this direction, semi-supervised domain adaptation (SSDA) is proposed, as it can leverage both labeled source and target data as well as unlabeled target data. Further, several SSDA classification methods have been proposed to use instance constraints \citep{donahue2013semi}, subspace learning \citep{yao2015semi}, entropy minimax \citep{saito2019semi}, adversarial attack \citep{kim2020attract}, etc. These methods are based on discriminative class boundaries for image classification, which, however, cannot be directly applied to segmentation. In \cite{liu2022ACT}, the asymmetric co-training is proposed to achieve semi-supervised domain adaptation for medical image segmentation.

The unified framework for both SSL and UDA is able to utilize both labeled and unlabeled target domain data. The alternative training based methods, e.g., self-training, have been applied to these two tasks, which can have a great potential for semi-supervised domain adaptation. Semi-supervised domain adaptation for object detection and image generation, however, are largely underexplored.

\subsection{Domain Generalization}  
 
Most of prior work on UDA assumes that there is a single source domain, while recent work has shown that network generalization can be further improved with multiple source domain sets \citep{montesuma2021wasserstein}.~By observing varying datasets, networks can learn domain invariant cues. Domain generalization further removes the requirement of unlabeled target domain data in multi-source UDA \citep{matsuura2019domain}. There are two main streams for domain generalization tasks. The first stream is to learn domain invariant features \citep{ghifary2016scatter}. For example, \cite{li2018domain,liu2021mutualpami} utilize adversarial training to mitigate the domain divergence. The second stream targets to fuse the domain-specific feature representations. For instance, \cite{mancini2018best} develop the domain-specific classifiers with multiple independent models. Then, the domain agnostic components are fused to form the domain-wise classification probability. \cite{ding2017deep} propose to match the low-rank structure of domain-specific features. \cite{liu2021Generalization} further propose to align the conditional distribution with a variational inference scheme.

Domain generalization is usually considered a multi-task learning problem \citep{liu2021mutualpami}, by exploring multiple source domains. How to achieve good test time adaptation for an unseen domain can be a challenging problem. For example, the label shift can be adaptively corrected in test time implementation as in \cite{liu2021Generalization}.

\subsection{Out-of-distribution Detection}  

OOD detection or deep OOD detection has recently been an active research topic \citep{che2021deep}. If the domain shift is too large for reliable adaptation, a more reasonable choice would be to reject significant outliers rather than to make adapted predictions with high uncertainty. While detecting the OOD samples in a low-dimensional space has been well-studied \citep{Pimentel2014A}, it is still challenging to detect OOD in high-dimensional complex data, e.g., images \citep{Liang2018Enhancing}. For example, \cite{Hendrycks2016A} identify that trained DNNs usually have higher maximum softmax output for in-distribution examples than anomalous ones. A possible improvement of this baseline would be to consider both the in-distribution and out-of-distribution training samples during training \citep{Hendrycks2019deep}. However, enumerating all possible OOD distributions before deployment is usually not possible. \cite{Liang2018Enhancing} propose that the difference between maximum probabilities in softmax distributions on ID/OOD samples can be made more significant, by means of adversarial perturbation pre-processing during training.

\cite{Devries2018Learning} augment the classifier with a confidence estimation branch, and adjust the objective using the predicted confidence score for training. \cite{ Lee2017Training} train a classifier simultaneously with a GAN, with an additional objective to encourage low confidence in generated samples. \cite{Hendrycks2019deep} propose to use real OOD samples instead of generated ones to train the detector. \cite{Vyas2018Out} label a part of training data as OOD samples to train a classifier, where that approach dynamically changes the partition of ID and OOD samples. These improvements based on \cite{Hendrycks2016A} typically need to retrain a classifier with modified structures or optimization objectives. Recently, \cite{Lee2018A} propose a new framework for anomaly detection. A number of methods \citep{Liang2018Enhancing,Vyas2018Out,Lee2018A} need OOD samples for tuning hyper-parameter selection, e.g., the threshold for verification. DVN \citep{che2021deep} aims to verify the predictions of a trained deep model, by estimating $p(x|y)$ rather than $p(x)$.

It remains an exciting open problem of how to train good density estimators on complex datasets, which is an important module for OOD detection. In addition, the connection and difference between OOD sample and adversarial attack samples also need further explorations.


\section{CONCLUSION}

In this paper, we have systematically reviewed deep learning-based UDA approaches. Deep learning has already surpassed its predecessors in a variety of fields, and future research in deep learning will strive toward the seamless deployment of trained models in a source domain into unseen and new target domains. Toward this goal, we provided a comprehensive summary of recent deep UDA approaches along with the merits and demerits of those approaches. Furthermore, several successful applications of deep UDA methods were reviewed. Finally, several challenges of the current deep UDA approaches were identified, which could serve as promising future directions.



\vskip 0.2in
\bibliography{sample}

\begin{thebibliography}{321}
\providecommand{\natexlab}[1]{#1}
\providecommand{\url}[1]{\texttt{#1}}
\expandafter\ifx\csname urlstyle\endcsname\relax
  \providecommand{\doi}[1]{doi: #1}\else
  \providecommand{\doi}{doi: \begingroup \urlstyle{rm}\Url}\fi

\bibitem[Adler and Lunz(2018)]{adler2018banach}
Jonas Adler and Sebastian Lunz.
\newblock Banach wasserstein gan.
\newblock \emph{Advances in Neural Information Processing Systems}, 31, 2018.

\bibitem[Ahn et~al.(2020)Ahn, Kumar, Fulham, Feng, and
  Kim]{ahn2020unsupervised}
Euijoon Ahn, Ashnil Kumar, Michael Fulham, Dagan Feng, and Jinman Kim.
\newblock Unsupervised domain adaptation to classify medical images using
  zero-bias convolutional auto-encoders and context-based feature augmentation.
\newblock \emph{IEEE transactions on medical imaging}, 39\penalty0
  (7):\penalty0 2385--2394, 2020.

\bibitem[Anoop et~al.(2021)Anoop, Prathosh, and
  Ramakrishnan]{anoop2021unsupervised}
CS~Anoop, AP~Prathosh, and AG~Ramakrishnan.
\newblock Unsupervised domain adaptation schemes for building asr in
  low-resource languages.
\newblock In \emph{2021 IEEE Automatic Speech Recognition and Understanding
  Workshop (ASRU)}, pages 342--349. IEEE, 2021.

\bibitem[Azizzadenesheli et~al.(2019)Azizzadenesheli, Liu, Yang, and
  Anandkumar]{azizzadenesheli2019regularized}
Kamyar Azizzadenesheli, Anqi Liu, Fanny Yang, and Animashree Anandkumar.
\newblock Regularized learning for domain adaptation under label shifts.
\newblock \emph{ICLR}, 2019.

\bibitem[Bateson et~al.(2020)Bateson, Kervadec, Dolz, Lombaert, and
  Ayed]{bateson2020source}
Mathilde Bateson, Hoel Kervadec, Jose Dolz, Herve Lombaert, and Ismail~Ben
  Ayed.
\newblock Source-relaxed domain adaptation for image segmentation.
\newblock In \emph{International Conference on Medical Image Computing and
  Computer-Assisted Intervention}, pages 490--499. Springer, 2020.

\bibitem[Beijbom(2012)]{beijbom2012domain}
Oscar Beijbom.
\newblock Domain adaptations for computer vision applications.
\newblock \emph{arXiv preprint arXiv:1211.4860}, 2012.

\bibitem[Bengio et~al.(2013)Bengio, Courville, and
  Vincent]{bengio2013representation}
Yoshua Bengio, Aaron Courville, and Pascal Vincent.
\newblock Representation learning: A review and new perspectives.
\newblock \emph{IEEE transactions on pattern analysis and machine
  intelligence}, 35\penalty0 (8):\penalty0 1798--1828, 2013.

\bibitem[Betlehem et~al.(2015)Betlehem, Zhang, Poletti, and
  Abhayapala]{betlehem2015personal}
Terence Betlehem, Wen Zhang, Mark~A Poletti, and Thushara~D Abhayapala.
\newblock Personal sound zones: Delivering interface-free audio to multiple
  listeners.
\newblock \emph{IEEE Signal Processing Magazine}, 32\penalty0 (2):\penalty0
  81--91, 2015.

\bibitem[Bommasani et~al.(2021)Bommasani, Hudson, Adeli, Altman, Arora, von
  Arx, Bernstein, Bohg, Bosselut, Brunskill,
  et~al.]{bommasani2021opportunities}
Rishi Bommasani, Drew~A Hudson, Ehsan Adeli, Russ Altman, Simran Arora, Sydney
  von Arx, Michael~S Bernstein, Jeannette Bohg, Antoine Bosselut, Emma
  Brunskill, et~al.
\newblock On the opportunities and risks of foundation models.
\newblock \emph{arXiv preprint arXiv:2108.07258}, 2021.

\bibitem[Bousmalis et~al.(2016)Bousmalis, Trigeorgis, Silberman, Krishnan, and
  Erhan]{bousmalis2016domain}
Konstantinos Bousmalis, George Trigeorgis, Nathan Silberman, Dilip Krishnan,
  and Dumitru Erhan.
\newblock Domain separation networks.
\newblock \emph{Advances in neural information processing systems}, 29, 2016.

\bibitem[Bousmalis et~al.(2017)Bousmalis, Silberman, Dohan, Erhan, and
  Krishnan]{bousmalis2017unsupervised}
Konstantinos Bousmalis, Nathan Silberman, David Dohan, Dumitru Erhan, and Dilip
  Krishnan.
\newblock Unsupervised pixel-level domain adaptation with generative
  adversarial networks.
\newblock In \emph{Proceedings of the IEEE conference on computer vision and
  pattern recognition}, pages 3722--3731, 2017.

\bibitem[Bousmalis et~al.(2018)Bousmalis, Irpan, Wohlhart, Bai, Kelcey,
  Kalakrishnan, Downs, Ibarz, Pastor, Konolige, et~al.]{bousmalis2018using}
Konstantinos Bousmalis, Alex Irpan, Paul Wohlhart, Yunfei Bai, Matthew Kelcey,
  Mrinal Kalakrishnan, Laura Downs, Julian Ibarz, Peter Pastor, Kurt Konolige,
  et~al.
\newblock Using simulation and domain adaptation to improve efficiency of deep
  robotic grasping.
\newblock In \emph{2018 IEEE international conference on robotics and
  automation (ICRA)}, pages 4243--4250. IEEE, 2018.

\bibitem[Bungum and Gamb{\"a}ck(2011)]{bungum2011survey}
Lars Bungum and Bj{\"o}rn Gamb{\"a}ck.
\newblock A survey of domain adaptation in machine translation: Towards a
  refinement of domain space.
\newblock In \emph{Proceedings of the India-Norway Workshop on Web Concepts and
  Technologies}, volume 112, 2011.

\bibitem[Burstein et~al.(2019)Burstein, Doran, and
  Solorio]{burstein2019proceedings}
Jill Burstein, Christy Doran, and Thamar Solorio.
\newblock Proceedings of the 2019 conference of the north american chapter of
  the association for computational linguistics: human language technologies,
  volume 1 (long and short papers).
\newblock In \emph{Proceedings of the 2019 Conference of the North American
  Chapter of the Association for Computational Linguistics: Human Language
  Technologies, Volume 1 (Long and Short Papers)}, 2019.

\bibitem[Cai et~al.(2019)Cai, Pan, Ngo, Tian, Duan, and Yao]{cai2019exploring}
Qi~Cai, Yingwei Pan, Chong-Wah Ngo, Xinmei Tian, Lingyu Duan, and Ting Yao.
\newblock Exploring object relation in mean teacher for cross-domain detection.
\newblock In \emph{Proceedings of the IEEE/CVF Conference on Computer Vision
  and Pattern Recognition}, pages 11457--11466, 2019.

\bibitem[Cai et~al.(2021)Cai, Chen, Li, Chen, Zhang, Ye, Li, Yang, and
  Zhang]{cai2021time}
Ruichu Cai, Jiawei Chen, Zijian Li, Wei Chen, Keli Zhang, Junjian Ye, Zhuozhang
  Li, Xiaoyan Yang, and Zhenjie Zhang.
\newblock Time series domain adaptation via sparse associative structure
  alignment.
\newblock In \emph{Proceedings of the AAAI Conference on Artificial
  Intelligence}, volume~35, pages 6859--6867, 2021.

\bibitem[Camps-Valls et~al.(2021)Camps-Valls, Tuia, Zhu, and
  Reichstein]{camps2021deep}
Gustau Camps-Valls, Devis Tuia, Xiao~Xiang Zhu, and Markus Reichstein.
\newblock \emph{Deep learning for the Earth Sciences: A comprehensive approach
  to remote sensing, climate science and geosciences}.
\newblock John Wiley \& Sons, 2021.

\bibitem[Cao et~al.(2018{\natexlab{a}})Cao, Long, Wang, and
  Jordan]{cao2018partialb}
Zhangjie Cao, Mingsheng Long, Jianmin Wang, and Michael~I Jordan.
\newblock Partial transfer learning with selective adversarial networks.
\newblock In \emph{CVPR}, pages 2724--2732, 2018{\natexlab{a}}.

\bibitem[Cao et~al.(2018{\natexlab{b}})Cao, Ma, Long, and
  Wang]{cao2018partiala}
Zhangjie Cao, Lijia Ma, Mingsheng Long, and Jianmin Wang.
\newblock Partial adversarial domain adaptation.
\newblock In \emph{Proceedings of the European Conference on Computer Vision
  (ECCV)}, pages 135--150, 2018{\natexlab{b}}.

\bibitem[Carlucci et~al.(2017)Carlucci, Porzi, Caputo, Ricci, and
  Bulo]{carlucci2017autodial}
Fabio~Maria Carlucci, Lorenzo Porzi, Barbara Caputo, Elisa Ricci, and
  Samuel~Rota Bulo.
\newblock Autodial: Automatic domain alignment layers.
\newblock In \emph{2017 IEEE international conference on computer vision
  (ICCV)}, pages 5077--5085. IEEE, 2017.

\bibitem[Chan and Ng(2005)]{chan2005word}
Yee~Seng Chan and Hwee~Tou Ng.
\newblock Word sense disambiguation with distribution estimation.
\newblock In \emph{IJCAI}, volume~5, pages 1010--5, 2005.

\bibitem[Chang et~al.(2019)Chang, You, Seo, Kwak, and Han]{chang2019domain}
Woong-Gi Chang, Tackgeun You, Seonguk Seo, Suha Kwak, and Bohyung Han.
\newblock Domain-specific batch normalization for unsupervised domain
  adaptation.
\newblock In \emph{Proceedings of the IEEE/CVF Conference on Computer Vision
  and Pattern Recognition}, pages 7354--7362, 2019.

\bibitem[Chang et~al.(2020)Chang, Mathur, Isopoussu, Song, and
  Kawsar]{chang2020systematic}
Youngjae Chang, Akhil Mathur, Anton Isopoussu, Junehwa Song, and Fahim Kawsar.
\newblock A systematic study of unsupervised domain adaptation for robust
  human-activity recognition.
\newblock \emph{Proceedings of the ACM on Interactive, Mobile, Wearable and
  Ubiquitous Technologies}, 4\penalty0 (1):\penalty0 1--30, 2020.

\bibitem[Chapelle and Zien(2005)]{chapelle2005semi}
Olivier Chapelle and Alexander Zien.
\newblock Semi-supervised classification by low density separation.
\newblock In \emph{International workshop on artificial intelligence and
  statistics}, pages 57--64. PMLR, 2005.

\bibitem[Che et~al.(2021)Che, Liu, Li, Ge, Zhang, Xiong, and
  Bengio]{che2021deep}
Tong Che, Xiaofeng Liu, Site Li, Yubin Ge, Ruixiang Zhang, Caiming Xiong, and
  Yoshua Bengio.
\newblock Deep verifier networks: Verification of deep discriminative models
  with deep generative models, 2021.

\bibitem[Chen et~al.(2019{\natexlab{a}})Chen, Miao, Lu, Xie, Blunsom, Markham,
  and Trigoni]{chen2019motiontransformer}
Changhao Chen, Yishu Miao, Chris~Xiaoxuan Lu, Linhai Xie, Phil Blunsom, Andrew
  Markham, and Niki Trigoni.
\newblock Motiontransformer: Transferring neural inertial tracking between
  domains.
\newblock In \emph{Proceedings of the AAAI Conference on Artificial
  Intelligence}, volume~33, pages 8009--8016, 2019{\natexlab{a}}.

\bibitem[Chen et~al.(2020{\natexlab{a}})Chen, Fu, Chen, Jin, Cheng, Jin, and
  Hua]{chen2020homm}
Chao Chen, Zhihang Fu, Zhihong Chen, Sheng Jin, Zhaowei Cheng, Xinyu Jin, and
  Xian-Sheng Hua.
\newblock Homm: Higher-order moment matching for unsupervised domain
  adaptation.
\newblock In \emph{Proceedings of the AAAI conference on artificial
  intelligence}, volume~34, pages 3422--3429, 2020{\natexlab{a}}.

\bibitem[Chen et~al.(2020{\natexlab{b}})Chen, Zheng, Ding, Huang, and
  Dou]{chen2020harmonizing}
Chaoqi Chen, Zebiao Zheng, Xinghao Ding, Yue Huang, and Qi~Dou.
\newblock Harmonizing transferability and discriminability for adapting object
  detectors.
\newblock In \emph{Proceedings of the IEEE/CVF Conference on Computer Vision
  and Pattern Recognition}, pages 8869--8878, 2020{\natexlab{b}}.

\bibitem[Chen et~al.(2019{\natexlab{b}})Chen, Dou, Chen, Qin, and
  Heng]{chen2019synergistic}
Cheng Chen, Qi~Dou, Hao Chen, Jing Qin, and Pheng-Ann Heng.
\newblock Synergistic image and feature adaptation: Towards cross-modality
  domain adaptation for medical image segmentation.
\newblock In \emph{Proceedings of the AAAI Conference on Artificial
  Intelligence}, volume~33, pages 865--872, 2019{\natexlab{b}}.

\bibitem[Chen et~al.(2020{\natexlab{c}})Chen, Li, Ma, and
  Zheng]{chen2020generative}
Jiawei Chen, Yuexiang Li, Kai Ma, and Yefeng Zheng.
\newblock Generative adversarial networks for video-to-video domain adaptation.
\newblock In \emph{Proceedings of the AAAI Conference on Artificial
  Intelligence}, volume~34, pages 3462--3469, 2020{\natexlab{c}}.

\bibitem[Chen et~al.(2019{\natexlab{c}})Chen, Kira, AlRegib, Yoo, Chen, and
  Zheng]{chen2019temporal}
Min-Hung Chen, Zsolt Kira, Ghassan AlRegib, Jaekwon Yoo, Ruxin Chen, and Jian
  Zheng.
\newblock Temporal attentive alignment for large-scale video domain adaptation.
\newblock In \emph{Proceedings of the IEEE/CVF International Conference on
  Computer Vision}, pages 6321--6330, 2019{\natexlab{c}}.

\bibitem[Chen et~al.(2020{\natexlab{d}})Chen, Li, Bao, and
  AlRegib]{chen2020action_mixed}
Min-Hung Chen, Baopu Li, Yingze Bao, and Ghassan AlRegib.
\newblock Action segmentation with mixed temporal domain adaptation.
\newblock In \emph{Proceedings of the IEEE/CVF Winter Conference on
  Applications of Computer Vision}, pages 605--614, 2020{\natexlab{d}}.

\bibitem[Chen et~al.(2020{\natexlab{e}})Chen, Li, Bao, AlRegib, and
  Kira]{chen2020action}
Min-Hung Chen, Baopu Li, Yingze Bao, Ghassan AlRegib, and Zsolt Kira.
\newblock Action segmentation with joint self-supervised temporal domain
  adaptation.
\newblock In \emph{Proceedings of the IEEE/CVF Conference on Computer Vision
  and Pattern Recognition}, pages 9454--9463, 2020{\natexlab{e}}.

\bibitem[Chen et~al.(2022)Chen, Gao, and Ma]{chen2022multi}
Peipeng Chen, Yuan Gao, and Andy~J Ma.
\newblock Multi-level attentive adversarial learning with temporal dilation for
  unsupervised video domain adaptation.
\newblock In \emph{Proceedings of the IEEE/CVF Winter Conference on
  Applications of Computer Vision}, pages 1259--1268, 2022.

\bibitem[Chen et~al.(2021{\natexlab{a}})Chen, Li, Wu, Dong, and
  Shi]{chen2021unsupervised}
Pengfei Chen, Leida Li, Jinjian Wu, Weisheng Dong, and Guangming Shi.
\newblock Unsupervised curriculum domain adaptation for no-reference video
  quality assessment.
\newblock In \emph{Proceedings of the IEEE/CVF International Conference on
  Computer Vision}, pages 5178--5187, 2021{\natexlab{a}}.

\bibitem[Chen et~al.(2018{\natexlab{a}})Chen, Liu, Wang, Wassell, and
  Chetty]{chen2018re}
Qingchao Chen, Yang Liu, Zhaowen Wang, Ian Wassell, and Kevin Chetty.
\newblock Re-weighted adversarial adaptation network for unsupervised domain
  adaptation.
\newblock In \emph{Proceedings of the IEEE Conference on Computer Vision and
  Pattern Recognition}, pages 7976--7985, 2018{\natexlab{a}}.

\bibitem[Chen et~al.(2021{\natexlab{b}})Chen, Liu, and Albanie]{chen2021mind}
Qingchao Chen, Yang Liu, and Samuel Albanie.
\newblock Mind-the-gap! unsupervised domain adaptation for text-video
  retrieval.
\newblock In \emph{Proceedings of the AAAI Conference on Artificial
  Intelligence}, volume~35, pages 1072--1080, 2021{\natexlab{b}}.

\bibitem[Chen et~al.(2017)Chen, Liao, Chuang, Hsu, Fu, and Sun]{chen2017show}
Tseng-Hung Chen, Yuan-Hong Liao, Ching-Yao Chuang, Wan-Ting Hsu, Jianlong Fu,
  and Min Sun.
\newblock Show, adapt and tell: Adversarial training of cross-domain image
  captioner.
\newblock In \emph{Proceedings of the IEEE international conference on computer
  vision}, pages 521--530, 2017.

\bibitem[Chen et~al.(2018{\natexlab{b}})Chen, Sun, Athiwaratkun, Cardie, and
  Weinberger]{chen2018adversarial}
Xilun Chen, Yu~Sun, Ben Athiwaratkun, Claire Cardie, and Kilian Weinberger.
\newblock Adversarial deep averaging networks for cross-lingual sentiment
  classification.
\newblock \emph{Transactions of the Association for Computational Linguistics},
  6:\penalty0 557--570, 2018{\natexlab{b}}.

\bibitem[Chen et~al.(2018{\natexlab{c}})Chen, Li, Sakaridis, Dai, and
  Van~Gool]{chen2018domain}
Yuhua Chen, Wen Li, Christos Sakaridis, Dengxin Dai, and Luc Van~Gool.
\newblock Domain adaptive faster r-cnn for object detection in the wild.
\newblock In \emph{Proceedings of the IEEE conference on computer vision and
  pattern recognition}, pages 3339--3348, 2018{\natexlab{c}}.

\bibitem[Choi et~al.(2020{\natexlab{a}})Choi, Sharma, Chandraker, and
  Huang]{choi2020unsupervised}
Jinwoo Choi, Gaurav Sharma, Manmohan Chandraker, and Jia-Bin Huang.
\newblock Unsupervised and semi-supervised domain adaptation for action
  recognition from drones.
\newblock In \emph{Proceedings of the IEEE/CVF Winter Conference on
  Applications of Computer Vision}, pages 1717--1726, 2020{\natexlab{a}}.

\bibitem[Choi et~al.(2020{\natexlab{b}})Choi, Sharma, Schulter, and
  Huang]{choi2020shuffle}
Jinwoo Choi, Gaurav Sharma, Samuel Schulter, and Jia-Bin Huang.
\newblock Shuffle and attend: Video domain adaptation.
\newblock In \emph{European Conference on Computer Vision}, pages 678--695.
  Springer, 2020{\natexlab{b}}.

\bibitem[Chopra et~al.(2013)Chopra, Balakrishnan, and Gopalan]{chopra2013dlid}
Sumit Chopra, Suhrid Balakrishnan, and Raghuraman Gopalan.
\newblock Dlid: Deep learning for domain adaptation by interpolating between
  domains.
\newblock In \emph{ICML workshop on challenges in representation learning},
  volume~2. Citeseer, 2013.

\bibitem[Chung et~al.(2015)Chung, Kastner, Dinh, Goel, Courville, and
  Bengio]{chung2015recurrent}
Junyoung Chung, Kyle Kastner, Laurent Dinh, Kratarth Goel, Aaron~C Courville,
  and Yoshua Bengio.
\newblock A recurrent latent variable model for sequential data.
\newblock \emph{Advances in neural information processing systems}, 28, 2015.

\bibitem[Cook et~al.(2013)Cook, Feuz, and Krishnan]{cook2013transfer}
Diane Cook, Kyle~D Feuz, and Narayanan~C Krishnan.
\newblock Transfer learning for activity recognition: A survey.
\newblock \emph{Knowledge and information systems}, 36\penalty0 (3):\penalty0
  537--556, 2013.

\bibitem[Cordts et~al.(2016)Cordts, Omran, Ramos, Rehfeld, Enzweiler, Benenson,
  Franke, Roth, and Schiele]{cordts2016cityscapes}
Marius Cordts, Mohamed Omran, Sebastian Ramos, Timo Rehfeld, Markus Enzweiler,
  Rodrigo Benenson, Uwe Franke, Stefan Roth, and Bernt Schiele.
\newblock The cityscapes dataset for semantic urban scene understanding.
\newblock In \emph{Proceedings of the IEEE conference on computer vision and
  pattern recognition}, pages 3213--3223, 2016.

\bibitem[Courty et~al.(2017)Courty, Flamary, Habrard, and
  Rakotomamonjy]{courty2017joint}
Nicolas Courty, Remi Flamary, Amaury Habrard, and Alain Rakotomamonjy.
\newblock Joint distribution optimal transportation for domain adaptation.
\newblock \emph{Advances in Neural Information Processing Systems}, 30, 2017.

\bibitem[Csurka(2017)]{csurka2017domain}
Gabriela Csurka.
\newblock Domain adaptation for visual applications: A comprehensive survey.
\newblock \emph{arXiv preprint arXiv:1702.05374}, 2017.

\bibitem[da~Costa et~al.(2020)da~Costa, Ak{\c{c}}ay, Zhang, and
  Kaymak]{da2020remaining}
Paulo Roberto de~Oliveira da~Costa, Alp Ak{\c{c}}ay, Yingqian Zhang, and Uzay
  Kaymak.
\newblock Remaining useful lifetime prediction via deep domain adaptation.
\newblock \emph{Reliability Engineering \& System Safety}, 195:\penalty0
  106682, 2020.

\bibitem[Damodaran et~al.(2018)Damodaran, Kellenberger, Flamary, Tuia, and
  Courty]{damodaran2018deepjdot}
Bharath~Bhushan Damodaran, Benjamin Kellenberger, Remi Flamary, Devis Tuia, and
  Nicolas Courty.
\newblock Deepjdot: Deep joint distribution optimal transport for unsupervised
  domain adaptation.
\newblock In \emph{Proceedings of the European Conference on Computer Vision
  (ECCV)}, pages 447--463, 2018.

\bibitem[Das and Lee(2018{\natexlab{a}})]{das2018graph}
Debasmit Das and CS~Lee.
\newblock Graph matching and pseudo-label guided deep unsupervised domain
  adaptation.
\newblock In \emph{International conference on artificial neural networks},
  pages 342--352. Springer, 2018{\natexlab{a}}.

\bibitem[Das and Lee(2018{\natexlab{b}})]{das2018unsupervised}
Debasmit Das and CS~George Lee.
\newblock Unsupervised domain adaptation using regularized hyper-graph
  matching.
\newblock In \emph{2018 25th IEEE International Conference on Image Processing
  (ICIP)}, pages 3758--3762. IEEE, 2018{\natexlab{b}}.

\bibitem[Deng et~al.(2021)Deng, Li, Chen, and Duan]{deng2021unbiased}
Jinhong Deng, Wen Li, Yuhua Chen, and Lixin Duan.
\newblock Unbiased mean teacher for cross-domain object detection.
\newblock In \emph{Proceedings of the IEEE/CVF Conference on Computer Vision
  and Pattern Recognition}, pages 4091--4101, 2021.

\bibitem[Der~Kiureghian and Ditlevsen(2009)]{der2009aleatory}
Armen Der~Kiureghian and Ove Ditlevsen.
\newblock Aleatory or epistemic? does it matter?
\newblock \emph{Structural safety}, 31\penalty0 (2):\penalty0 105--112, 2009.

\bibitem[Desai et~al.(2019)Desai, Sinno, Rosenfeld, and Li]{desai2019adaptive}
Shrey Desai, Barea Sinno, Alex Rosenfeld, and Junyi~Jessy Li.
\newblock Adaptive ensembling: Unsupervised domain adaptation for political
  document analysis.
\newblock In \emph{Proceedings of the 2019 Conference on Empirical Methods in
  Natural Language Processing and the 9th International Joint Conference on
  Natural Language Processing (EMNLP-IJCNLP)}, pages 4718--4730, 2019.

\bibitem[Devries and Taylor(2018)]{Devries2018Learning}
Terrance Devries and Graham~W. Taylor.
\newblock Learning confidence for out-of-distribution detection in neural
  networks.
\newblock 2018.

\bibitem[Ding and Fu(2017)]{ding2017deep}
Zhengming Ding and Yun Fu.
\newblock Deep domain generalization with structured low-rank constraint.
\newblock \emph{TIP}, 2017.

\bibitem[Donahue et~al.(2013)Donahue, Hoffman, Rodner, Saenko, and
  Darrell]{donahue2013semi}
Jeff Donahue, Judy Hoffman, Erik Rodner, Kate Saenko, and Trevor Darrell.
\newblock Semi-supervised domain adaptation with instance constraints.
\newblock In \emph{Proceedings of the IEEE conference on computer vision and
  pattern recognition}, pages 668--675, 2013.

\bibitem[Du et~al.(2019)Du, Jin, Song, and Dai]{du2019unsupervised}
Hao Du, Tian Jin, Yongping Song, and Yongpeng Dai.
\newblock Unsupervised adversarial domain adaptation for micro-doppler based
  human activity classification.
\newblock \emph{IEEE geoscience and remote sensing letters}, 17\penalty0
  (1):\penalty0 62--66, 2019.

\bibitem[Ekladious et~al.(2020)Ekladious, Lemoine, Granger, Kamali, and
  Moudache]{ekladious2020dual}
George Ekladious, Hugo Lemoine, Eric Granger, Kaveh Kamali, and Salim Moudache.
\newblock Dual-triplet metric learning for unsupervised domain adaptation in
  video face recognition.
\newblock In \emph{2020 International Joint Conference on Neural Networks
  (IJCNN)}, pages 1--9. IEEE, 2020.

\bibitem[Fan et~al.(2022)Fan, Zhu, Jiang, Meng, Chen, Fu, Yu, Dai, and
  Chen]{fan2022unsupervised}
Jiahao Fan, Hangyu Zhu, Xinyu Jiang, Long Meng, Chen Chen, Cong Fu, Huan Yu,
  Chenyun Dai, and Wei Chen.
\newblock Unsupervised domain adaptation by statistics alignment for deep sleep
  staging networks.
\newblock \emph{IEEE Transactions on Neural Systems and Rehabilitation
  Engineering}, 30:\penalty0 205--216, 2022.

\bibitem[Fredericks et~al.(2014)Fredericks, DeVries, and
  Cheng]{fredericks2014towards}
Erik~M Fredericks, Byron DeVries, and Betty~HC Cheng.
\newblock Towards run-time adaptation of test cases for self-adaptive systems
  in the face of uncertainty.
\newblock In \emph{Proceedings of the 9th International Symposium on Software
  Engineering for Adaptive and Self-Managing Systems}, pages 17--26, 2014.

\bibitem[French et~al.(2017)French, Mackiewicz, and Fisher]{french2017self}
Geoffrey French, Michal Mackiewicz, and Mark Fisher.
\newblock Self-ensembling for visual domain adaptation.
\newblock \emph{arXiv preprint arXiv:1706.05208}, 2017.

\bibitem[Fu et~al.(2017)Fu, Nguyen, Min, and Grishman]{fu2017domain}
Lisheng Fu, Thien~Huu Nguyen, Bonan Min, and Ralph Grishman.
\newblock Domain adaptation for relation extraction with domain adversarial
  neural network.
\newblock In \emph{Proceedings of the Eighth International Joint Conference on
  Natural Language Processing (Volume 2: Short Papers)}, pages 425--429, 2017.

\bibitem[Ganin and Lempitsky(2015)]{ganin2015unsupervised}
Yaroslav Ganin and Victor Lempitsky.
\newblock Unsupervised domain adaptation by backpropagation.
\newblock In \emph{ICML}, 2015.

\bibitem[Ganin et~al.(2016)Ganin, Ustinova, Ajakan, Germain, Larochelle,
  Laviolette, Marchand, and Lempitsky]{ganin2016domain}
Yaroslav Ganin, Evgeniya Ustinova, Hana Ajakan, Pascal Germain, Hugo
  Larochelle, Fran{\c{c}}ois Laviolette, Mario Marchand, and Victor Lempitsky.
\newblock Domain-adversarial training of neural networks.
\newblock \emph{The Journal of Machine Learning Research}, 17\penalty0
  (1):\penalty0 2096--2030, 2016.

\bibitem[Ge et~al.(2021{\natexlab{a}})Ge, Dinh, Liu, Su, Lu, Wang, and
  Diesner]{ge2021baco}
Yubin Ge, Ly~Dinh, Xiaofeng Liu, Jinsong Su, Ziyao Lu, Ante Wang, and Jana
  Diesner.
\newblock Baco: A background knowledge-and content-based framework for citing
  sentence generation.
\newblock In \emph{Proceedings of the 59th Annual Meeting of the Association
  for Computational Linguistics and the 11th International Joint Conference on
  Natural Language Processing (Volume 1: Long Papers)}, pages 1466--1478,
  2021{\natexlab{a}}.

\bibitem[Ge et~al.(2021{\natexlab{b}})Ge, Li, Li, Fan, Xie, You, and
  Liu]{ge2021embedding}
Yubin Ge, Site Li, Xuyang Li, Fangfang Fan, Wanqing Xie, Jane You, and Xiaofeng
  Liu.
\newblock Embedding semantic hierarchy in discrete optimal transport for risk
  minimization.
\newblock In \emph{ICASSP 2021-2021 IEEE International Conference on Acoustics,
  Speech and Signal Processing (ICASSP)}, pages 2835--2839. IEEE,
  2021{\natexlab{b}}.

\bibitem[Ghifary et~al.(2016)Ghifary, Kleijn, Zhang, Balduzzi, and
  Li]{ghifary2016deep}
Muhammad Ghifary, W~Bastiaan Kleijn, Mengjie Zhang, David Balduzzi, and Wen Li.
\newblock Deep reconstruction-classification networks for unsupervised domain
  adaptation.
\newblock In \emph{European conference on computer vision}, pages 597--613.
  Springer, 2016.

\bibitem[Ghifary et~al.(2017)Ghifary, Balduzzi, Kleijn, and
  Zhang]{ghifary2016scatter}
Muhammad Ghifary, David Balduzzi, W~Bastiaan Kleijn, and Mengjie Zhang.
\newblock Scatter component analysis: A unified framework for domain adaptation
  and domain generalization.
\newblock \emph{IEEE T-PAMI}, 2017.

\bibitem[Gholami et~al.(2019)Gholami, Sahu, Kim, and
  Pavlovic]{gholami2019taskdiscriminative}
Behnam Gholami, Pritish Sahu, Minyoung Kim, and Vladimir Pavlovic.
\newblock Task-discriminative domain alignment for unsupervised domain
  adaptation.
\newblock In \emph{ICCV}, 2019.

\bibitem[Ghosal et~al.(2020)Ghosal, Hazarika, Roy, Majumder, Mihalcea, and
  Poria]{ghosal2020kingdom}
Deepanway Ghosal, Devamanyu Hazarika, Abhinaba Roy, Navonil Majumder, Rada
  Mihalcea, and Soujanya Poria.
\newblock Kingdom: Knowledge-guided domain adaptation for sentiment analysis.
\newblock In \emph{Proceedings of the 58th Annual Meeting of the Association
  for Computational Linguistics}, pages 3198--3210, 2020.

\bibitem[Goodfellow et~al.(2016)Goodfellow, Bengio, and
  Courville]{goodfellow2016deep}
I.~Goodfellow, Y.~Bengio, and A.~Courville.
\newblock \emph{Deep Learning}.
\newblock MIT press, 2016.

\bibitem[Gu et~al.(2022)Gu, Okorn, and Held]{gu2022ossid}
Qiao Gu, Brian Okorn, and David Held.
\newblock Ossid: Online self-supervised instance detection by (and for) pose
  estimation.
\newblock \emph{IEEE Robotics and Automation Letters}, 2022.

\bibitem[Guan and Liu(2021)]{guan2021domain}
Hao Guan and Mingxia Liu.
\newblock Domain adaptation for medical image analysis: a survey.
\newblock \emph{IEEE Transactions on Biomedical Engineering}, 2021.

\bibitem[Guizilini et~al.(2021)Guizilini, Li, Ambruș, and
  Gaidon]{guizilini2021geometric}
Vitor Guizilini, Jie Li, Rareș Ambruș, and Adrien Gaidon.
\newblock Geometric unsupervised domain adaptation for semantic segmentation.
\newblock In \emph{Proceedings of the IEEE/CVF International Conference on
  Computer Vision}, pages 8537--8547, 2021.

\bibitem[Gururangan et~al.(2020)Gururangan, Marasovi{\'c}, Swayamdipta, Lo,
  Beltagy, Downey, and Smith]{gururangan2020don}
Suchin Gururangan, Ana Marasovi{\'c}, Swabha Swayamdipta, Kyle Lo, Iz~Beltagy,
  Doug Downey, and Noah~A Smith.
\newblock Don’t stop pretraining: Adapt language models to domains and tasks.
\newblock In \emph{Proceedings of the 58th Annual Meeting of the Association
  for Computational Linguistics}, pages 8342--8360, 2020.

\bibitem[Ham et~al.(2021)Ham, Yoo, and Kang]{ham2021training}
Yu-Jin Ham, Chaehwa Yoo, and Je-Won Kang.
\newblock Training compression artifacts reduction network with domain
  adaptation.
\newblock In \emph{Applications of Digital Image Processing XLIV}, volume
  11842, page 118420U. International Society for Optics and Photonics, 2021.

\bibitem[Han and Eisenstein(2019)]{han2019unsupervised}
Xiaochuang Han and Jacob Eisenstein.
\newblock Unsupervised domain adaptation of contextualized embeddings for
  sequence labeling.
\newblock In \emph{Proceedings of the 2019 Conference on Empirical Methods in
  Natural Language Processing and the 9th International Joint Conference on
  Natural Language Processing (EMNLP-IJCNLP)}, pages 4238--4248, 2019.

\bibitem[Han et~al.(2020)Han, Liu, Sheng, Ren, Han, You, Liu, and
  Luo]{han2020wasserstein}
Yuzhuo Han, Xiaofeng Liu, Zhenfei Sheng, Yutao Ren, Xu~Han, Jane You, Risheng
  Liu, and Zhongxuan Luo.
\newblock Wasserstein loss-based deep object detection.
\newblock In \emph{Proceedings of the IEEE/CVF Conference on Computer Vision
  and Pattern Recognition Workshops}, pages 998--999, 2020.

\bibitem[He et~al.(2020{\natexlab{a}})He, Liu, Fan, and
  You]{he2020classification}
Gewen He, Xiaofeng Liu, Fangfang Fan, and Jane You.
\newblock Classification-aware semi-supervised domain adaptation.
\newblock In \emph{Proceedings of the IEEE/CVF Conference on Computer Vision
  and Pattern Recognition Workshops}, pages 964--965, 2020{\natexlab{a}}.

\bibitem[He et~al.(2020{\natexlab{b}})He, Liu, Fan, and You]{he2020image2audio}
Gewen He, Xiaofeng Liu, Fangfang Fan, and Jane You.
\newblock Image2audio: Facilitating semi-supervised audio emotion recognition
  with facial expression image.
\newblock In \emph{Proceedings of the IEEE/CVF Conference on Computer Vision
  and Pattern Recognition Workshops}, pages 912--913, 2020{\natexlab{b}}.

\bibitem[He et~al.(2021{\natexlab{a}})He, Ye, Li, Pan, and Lu]{he2021online}
Wenwen He, Yalan Ye, Yunxia Li, Tongjie Pan, and Li~Lu.
\newblock Online cross-subject emotion recognition from ecg via unsupervised
  domain adaptation.
\newblock In \emph{2021 43rd Annual International Conference of the IEEE
  Engineering in Medicine \& Biology Society (EMBC)}, pages 1001--1005. IEEE,
  2021{\natexlab{a}}.

\bibitem[He et~al.(2020{\natexlab{c}})He, Carass, Liu, Saidha, Calabresi, and
  Prince]{he2020adversarial}
Yufan He, Aaron Carass, Yihao Liu, Shiv Saidha, Peter~A Calabresi, and Jerry~L
  Prince.
\newblock Adversarial domain adaptation for multi-device retinal oct
  segmentation.
\newblock In \emph{Medical Imaging 2020: Image Processing}, volume 11313, page
  1131309. International Society for Optics and Photonics, 2020{\natexlab{c}}.

\bibitem[He et~al.(2021{\natexlab{b}})He, Carass, Zuo, Dewey, and
  Prince]{he2021autoencoder}
Yufan He, Aaron Carass, Lianrui Zuo, Blake~E Dewey, and Jerry~L Prince.
\newblock Autoencoder based self-supervised test-time adaptation for medical
  image analysis.
\newblock \emph{Medical image analysis}, 72:\penalty0 102136,
  2021{\natexlab{b}}.

\bibitem[He et~al.(2022)He, Zhong, and Pan]{he2022adversarial}
Zhipeng He, Yongshi Zhong, and Jiahui Pan.
\newblock An adversarial discriminative temporal convolutional network for
  eeg-based cross-domain emotion recognition.
\newblock \emph{Computers in biology and medicine}, 141:\penalty0 105048, 2022.

\bibitem[Hendrycks and Gimpel(2017)]{Hendrycks2016A}
Dan Hendrycks and Kevin Gimpel.
\newblock A baseline for detecting misclassified and out-of-distribution
  examples in neural networks.
\newblock \emph{ICLR}, 2017.

\bibitem[Hendrycks et~al.(2019)Hendrycks, Mazeika, and
  Dietterich]{Hendrycks2019deep}
Dan Hendrycks, Mantas Mazeika, and Thomas Dietterich.
\newblock Deep anomaly detection with outlier exposure.
\newblock \emph{ICLR}, 2019.

\bibitem[Hochreiter and Schmidhuber(1997)]{hochreiter1997long}
Sepp Hochreiter and J{\"u}rgen Schmidhuber.
\newblock Long short-term memory.
\newblock \emph{Neural computation}, 1997.

\bibitem[Hoffman et~al.(2014)Hoffman, Darrell, and
  Saenko]{hoffman2014continuous}
Judy Hoffman, Trevor Darrell, and Kate Saenko.
\newblock Continuous manifold based adaptation for evolving visual domains.
\newblock In \emph{Proceedings of the IEEE Conference on Computer Vision and
  Pattern Recognition}, pages 867--874, 2014.

\bibitem[Hoffman et~al.(2018)Hoffman, Tzeng, Park, Zhu, Isola, Saenko, Efros,
  and Darrell]{hoffman2018cycada}
Judy Hoffman, Eric Tzeng, Taesung Park, Jun-Yan Zhu, Phillip Isola, Kate
  Saenko, Alexei~A Efros, and Trevor Darrell.
\newblock Cycada: Cycle-consistent adversarial domain adaptation.
\newblock In \emph{ICML}, 2018.

\bibitem[Hong et~al.(2017)Hong, Im, Ryu, and Yang]{hong2017sspp}
Sungeun Hong, Woobin Im, Jongbin Ryu, and Hyun~S Yang.
\newblock Sspp-dan: Deep domain adaptation network for face recognition with
  single sample per person.
\newblock In \emph{2017 IEEE International Conference on Image Processing
  (ICIP)}, pages 825--829. IEEE, 2017.

\bibitem[Hsu et~al.(2020{\natexlab{a}})Hsu, Tsai, Lin, and Yang]{hsu2020every}
Cheng-Chun Hsu, Yi-Hsuan Tsai, Yen-Yu Lin, and Ming-Hsuan Yang.
\newblock Every pixel matters: Center-aware feature alignment for domain
  adaptive object detector.
\newblock In \emph{European Conference on Computer Vision}, pages 733--748.
  Springer, 2020{\natexlab{a}}.

\bibitem[Hsu et~al.(2020{\natexlab{b}})Hsu, Yao, Tsai, Hung, Tseng, Singh, and
  Yang]{hsu2020progressive}
Han-Kai Hsu, Chun-Han Yao, Yi-Hsuan Tsai, Wei-Chih Hung, Hung-Yu Tseng, Maneesh
  Singh, and Ming-Hsuan Yang.
\newblock Progressive domain adaptation for object detection.
\newblock In \emph{Proceedings of the IEEE/CVF winter conference on
  applications of computer vision}, pages 749--757, 2020{\natexlab{b}}.

\bibitem[Hu et~al.(2020{\natexlab{a}})Hu, Liang, Hou, Yan, Chen, Yan, and
  Feng]{hu2020panda}
Dapeng Hu, Jian Liang, Qibin Hou, Hanshu Yan, Yunpeng Chen, Shuicheng Yan, and
  Jiashi Feng.
\newblock Panda: Prototypical unsupervised domain adaptation.
\newblock \emph{ECCV}, 2020{\natexlab{a}}.

\bibitem[Hu et~al.(2020{\natexlab{b}})Hu, Mou, and Zhu]{hu2020unsupervised}
Jingliang Hu, Lichao Mou, and Xiao~Xiang Zhu.
\newblock Unsupervised domain adaptation using a teacher-student network for
  cross-city classification of sentinel-2 images.
\newblock \emph{The International Archives of Photogrammetry, Remote Sensing
  and Spatial Information Sciences}, 43:\penalty0 1569--1574,
  2020{\natexlab{b}}.

\bibitem[Hu et~al.(2019)Hu, Worrall, Knegt, Veeling, Huisman, and
  Welling]{hu2019supervised}
Shi Hu, Daniel Worrall, Stefan Knegt, Bas Veeling, Henkjan Huisman, and Max
  Welling.
\newblock Supervised uncertainty quantification for segmentation with multiple
  annotations.
\newblock In \emph{MICCAI}, pages 137--145. Springer, 2019.

\bibitem[Huang et~al.(2020)Huang, Ali, Wang, Ning, Purushotham, Wang, and
  Zhang]{huang2020deep}
Xin Huang, Sahara Ali, Chenxi Wang, Zeyu Ning, Sanjay Purushotham, Jianwu Wang,
  and Zhibo Zhang.
\newblock Deep domain adaptation based cloud type detection using active and
  passive satellite data.
\newblock In \emph{2020 IEEE International Conference on Big Data (Big Data)},
  pages 1330--1337. IEEE, 2020.

\bibitem[Huo et~al.(2022)Huo, Xie, Hu, Zhou, Li, and Tian]{huo2022domain}
Xinyue Huo, Lingxi Xie, Hengtong Hu, Wengang Zhou, Houqiang Li, and Qi~Tian.
\newblock Domain-agnostic prior for transfer semantic segmentation.
\newblock \emph{arXiv preprint arXiv:2204.02684}, 2022.

\bibitem[Ioffe and Szegedy(2015)]{ioffe2015batch}
Sergey Ioffe and Christian Szegedy.
\newblock Batch normalization: Accelerating deep network training by reducing
  internal covariate shift.
\newblock In \emph{International conference on machine learning}, pages
  448--456, 2015.

\bibitem[Jiang et~al.(2018)Jiang, Miao, Ma, Yao, Wang, Yuan, Xue, Song, Ma,
  Koutsonikolas, et~al.]{jiang2018towards}
Wenjun Jiang, Chenglin Miao, Fenglong Ma, Shuochao Yao, Yaqing Wang, Ye~Yuan,
  Hongfei Xue, Chen Song, Xin Ma, Dimitrios Koutsonikolas, et~al.
\newblock Towards environment independent device free human activity
  recognition.
\newblock In \emph{Proceedings of the 24th Annual International Conference on
  Mobile Computing and Networking}, pages 289--304, 2018.

\bibitem[Jim{e}nez-Guarneros and G{o}mez-Gil(2021)]{jimenez2021standardization}
Magdiel Jim{e}nez-Guarneros and Pilar G{o}mez-Gil.
\newblock Standardization-refinement domain adaptation method for cross-subject
  eeg-based classification in imagined speech recognition.
\newblock \emph{Pattern Recognition Letters}, 141:\penalty0 54--60, 2021.

\bibitem[Judy et~al.(2018)Judy, Eric, Taesung, Jun-Yan, Phillip, Kate, Alexei,
  and Trevor]{judy2018cycada}
Hoffman Judy, Tzeng Eric, Park Taesung, Zhu Jun-Yan, Isola Phillip, Saenko
  Kate, Efros Alexei, and Darrell Trevor.
\newblock Cycada: Cycle-consistent adversarial domain adaptation.
\newblock In \emph{ICML}, pages 1994--2003, 2018.

\bibitem[Kan et~al.(2015)Kan, Shan, and Chen]{kan2015bi}
Meina Kan, Shiguang Shan, and Xilin Chen.
\newblock Bi-shifting auto-encoder for unsupervised domain adaptation.
\newblock In \emph{Proceedings of the IEEE international conference on computer
  vision}, pages 3846--3854, 2015.

\bibitem[Kang et~al.(2018)Kang, Zheng, Yan, and Yang]{kang2018deep}
Guoliang Kang, Liang Zheng, Yan Yan, and Yi~Yang.
\newblock Deep adversarial attention alignment for unsupervised domain
  adaptation: the benefit of target expectation maximization.
\newblock In \emph{Proceedings of the European conference on computer vision
  (ECCV)}, pages 401--416, 2018.

\bibitem[Kang et~al.(2019)Kang, Jiang, Yang, and
  Hauptmann]{kang2019contrastive}
Guoliang Kang, Lu~Jiang, Yi~Yang, and Alexander~G Hauptmann.
\newblock Contrastive adaptation network for unsupervised domain adaptation.
\newblock In \emph{Proceedings of the IEEE/CVF Conference on Computer Vision
  and Pattern Recognition}, pages 4893--4902, 2019.

\bibitem[Karani et~al.(2021)Karani, Erdil, Chaitanya, and
  Konukoglu]{karani2021test}
Neerav Karani, Ertunc Erdil, Krishna Chaitanya, and Ender Konukoglu.
\newblock Test-time adaptable neural networks for robust medical image
  segmentation.
\newblock \emph{Medical Image Analysis}, 68:\penalty0 101907, 2021.

\bibitem[Kendall and Gal(2017)]{kendall2017uncertainties}
Alex Kendall and Yarin Gal.
\newblock What uncertainties do we need in bayesian deep learning for computer
  vision?
\newblock \emph{arXiv:1703.04977}, 2017.

\bibitem[Khodabandeh et~al.(2019)Khodabandeh, Vahdat, Ranjbar, and
  Macready]{khodabandeh2019robust}
Mehran Khodabandeh, Arash Vahdat, Mani Ranjbar, and William~G Macready.
\newblock A robust learning approach to domain adaptive object detection.
\newblock In \emph{Proceedings of the IEEE/CVF International Conference on
  Computer Vision}, pages 480--490, 2019.

\bibitem[Khurana et~al.(2021)Khurana, Moritz, Hori, and
  Le~Roux]{khurana2021unsupervised}
Sameer Khurana, Niko Moritz, Takaaki Hori, and Jonathan Le~Roux.
\newblock Unsupervised domain adaptation for speech recognition via uncertainty
  driven self-training.
\newblock In \emph{ICASSP 2021-2021 IEEE International Conference on Acoustics,
  Speech and Signal Processing (ICASSP)}, pages 6553--6557. IEEE, 2021.

\bibitem[Kim et~al.(2020{\natexlab{a}})Kim, Saito, Oh, Plummer, Sclaroff, and
  Saenko]{kim2020cross}
Donghyun Kim, Kuniaki Saito, Tae-Hyun Oh, Bryan~A Plummer, Stan Sclaroff, and
  Kate Saenko.
\newblock Cross-domain self-supervised learning for domain adaptation with few
  source labels.
\newblock \emph{arXiv preprint arXiv:2003.08264}, 2020{\natexlab{a}}.

\bibitem[Kim et~al.(2019{\natexlab{a}})Kim, Choi, Kim, and Kim]{kim2019self}
Seunghyeon Kim, Jaehoon Choi, Taekyung Kim, and Changick Kim.
\newblock Self-training and adversarial background regularization for
  unsupervised domain adaptive one-stage object detection.
\newblock In \emph{Proceedings of the IEEE/CVF International Conference on
  Computer Vision}, pages 6092--6101, 2019{\natexlab{a}}.

\bibitem[Kim et~al.(2017)Kim, Cha, Kim, Lee, and Kim]{kim2017learning}
Taeksoo Kim, Moonsu Cha, Hyunsoo Kim, Jung~Kwon Lee, and Jiwon Kim.
\newblock Learning to discover cross-domain relations with generative
  adversarial networks.
\newblock In \emph{International Conference on Machine Learning}, pages
  1857--1865. PMLR, 2017.

\bibitem[Kim and Kim(2020)]{kim2020attract}
Taekyung Kim and Changick Kim.
\newblock Attract, perturb, and explore: Learning a feature alignment network
  for semi-supervised domain adaptation.
\newblock In \emph{European Conference on Computer Vision}, pages 591--607.
  Springer, 2020.

\bibitem[Kim et~al.(2019{\natexlab{b}})Kim, Jeong, Kim, Choi, and
  Kim]{kim2019diversify}
Taekyung Kim, Minki Jeong, Seunghyeon Kim, Seokeon Choi, and Changick Kim.
\newblock Diversify and match: A domain adaptive representation learning
  paradigm for object detection.
\newblock In \emph{Proceedings of the IEEE/CVF Conference on Computer Vision
  and Pattern Recognition}, pages 12456--12465, 2019{\natexlab{b}}.

\bibitem[Kim et~al.(2020{\natexlab{b}})Kim, Hong, Yang, Kang, Jeon, and
  Kim]{kim2020associative}
Youngeun Kim, Sungeun Hong, Seunghan Yang, Sungil Kang, Yunho Jeon, and Jiwon
  Kim.
\newblock Associative partial domain adaptation.
\newblock \emph{arXiv preprint arXiv:2008.03111}, 2020{\natexlab{b}}.

\bibitem[Kolesnikov et~al.(2019)Kolesnikov, Zhai, and
  Beyer]{kolesnikov2019revisiting}
Alexander Kolesnikov, Xiaohua Zhai, and Lucas Beyer.
\newblock Revisiting self-supervised visual representation learning.
\newblock In \emph{Proceedings of the IEEE/CVF conference on computer vision
  and pattern recognition}, pages 1920--1929, 2019.

\bibitem[Kouw(2018)]{kouw2018introduction}
Wouter~M Kouw.
\newblock An introduction to domain adaptation and transfer learning.
\newblock \emph{arXiv preprint arXiv:1812.11806}, 2018.

\bibitem[Kouw and Loog(2019)]{kouw2019review}
Wouter~M Kouw and Marco Loog.
\newblock A review of domain adaptation without target labels.
\newblock \emph{IEEE transactions on pattern analysis and machine
  intelligence}, 43\penalty0 (3):\penalty0 766--785, 2019.

\bibitem[Krizhevsky et~al.(2012)Krizhevsky, Sutskever, and
  Hinton]{krizhevsky2012imagenet}
Alex Krizhevsky, Ilya Sutskever, and Geoffrey~E Hinton.
\newblock Imagenet classification with deep convolutional neural networks.
\newblock \emph{Advances in neural information processing systems}, 25, 2012.

\bibitem[Kumar et~al.(2018)Kumar, Sattigeri, Wadhawan, Karlinsky, Feris,
  Freeman, and Wornell]{kumar2018co}
Abhishek Kumar, Prasanna Sattigeri, Kahini Wadhawan, Leonid Karlinsky, Rogerio
  Feris, Bill Freeman, and Gregory Wornell.
\newblock Co-regularized alignment for unsupervised domain adaptation.
\newblock \emph{Advances in Neural Information Processing Systems}, 31, 2018.

\bibitem[Lazaric(2012)]{lazaric2012transfer}
Alessandro Lazaric.
\newblock Transfer in reinforcement learning: a framework and a survey.
\newblock In \emph{Reinforcement Learning}, pages 143--173. Springer, 2012.

\bibitem[LeCun et~al.(2015)LeCun, Bengio, and Hinton]{lecun2015deep}
Yann LeCun, Yoshua Bengio, and Geoffrey Hinton.
\newblock Deep learning.
\newblock \emph{nature}, 521\penalty0 (7553):\penalty0 436--444, 2015.

\bibitem[Lee et~al.(2019{\natexlab{a}})Lee, Batra, Baig, and
  Ulbricht]{lee2019sliced}
Chen-Yu Lee, Tanmay Batra, Mohammad~Haris Baig, and Daniel Ulbricht.
\newblock Sliced wasserstein discrepancy for unsupervised domain adaptation.
\newblock In \emph{Proceedings of the IEEE/CVF Conference on Computer Vision
  and Pattern Recognition}, pages 10285--10295, 2019{\natexlab{a}}.

\bibitem[Lee et~al.(2018{\natexlab{a}})Lee, Lee, Lee, and
  Shin]{Lee2017Training}
Kimin Lee, Honglak Lee, Kibok Lee, and Jinwoo Shin.
\newblock Training confidence-calibrated classifiers for detecting
  out-of-distribution samples.
\newblock \emph{ICLR}, 2018{\natexlab{a}}.

\bibitem[Lee et~al.(2018{\natexlab{b}})Lee, Lee, Lee, and Shin]{Lee2018A}
Kimin Lee, Kibok Lee, Honglak Lee, and Jinwoo Shin.
\newblock A simple unified framework for detecting out-of-distribution samples
  and adversarial attacks.
\newblock \emph{NIPS}, 2018{\natexlab{b}}.

\bibitem[Lee et~al.(2019{\natexlab{b}})Lee, Kim, Kim, and Jeong]{lee2019drop}
Seungmin Lee, Dongwan Kim, Namil Kim, and Seong-Gyun Jeong.
\newblock Drop to adapt: Learning discriminative features for unsupervised
  domain adaptation.
\newblock In \emph{Proceedings of the IEEE/CVF International Conference on
  Computer Vision}, pages 91--100, 2019{\natexlab{b}}.

\bibitem[Li et~al.(2018{\natexlab{a}})Li, Jialin~Pan, Wang, and
  Kot]{li2018domain}
Haoliang Li, Sinno Jialin~Pan, Shiqi Wang, and Alex~C Kot.
\newblock Domain generalization with adversarial feature learning.
\newblock In \emph{CVPR}, 2018{\natexlab{a}}.

\bibitem[Li(2018)]{li2018twin}
Jerry Li.
\newblock Twin-gan--unpaired cross-domain image translation with weight-sharing
  gans.
\newblock \emph{arXiv preprint arXiv:1809.00946}, 2018.

\bibitem[Li et~al.(2019{\natexlab{a}})Li, Wei, Zhang, Wang, and
  Zhang]{li2019unsupervised}
Wei Li, Wei Wei, Lei Zhang, Cong Wang, and Yanning Zhang.
\newblock Unsupervised deep domain adaptation for hyperspectral image
  classification.
\newblock In \emph{IGARSS 2019-2019 IEEE International Geoscience and Remote
  Sensing Symposium}, pages 1--4. IEEE, 2019{\natexlab{a}}.

\bibitem[Li et~al.(2021{\natexlab{a}})Li, Niu, Gao, Liu, and
  Dong]{li2021unsupervised}
Xiaohui Li, Sijie Niu, Xizhan Gao, Tingting Liu, and Jiwen Dong.
\newblock Unsupervised domain adaptation with self-selected active learning for
  cross-domain oct image segmentation.
\newblock In \emph{International Conference on Neural Information Processing},
  pages 585--596. Springer, 2021{\natexlab{a}}.

\bibitem[Li et~al.(2018{\natexlab{b}})Li, Zheng, Zong, Cui, Zhang, and
  Zhou]{li2018bi}
Yang Li, Wenming Zheng, Yuan Zong, Zhen Cui, Tong Zhang, and Xiaoyan Zhou.
\newblock A bi-hemisphere domain adversarial neural network model for eeg
  emotion recognition.
\newblock \emph{IEEE Transactions on Affective Computing}, 12\penalty0
  (2):\penalty0 494--504, 2018{\natexlab{b}}.

\bibitem[Li et~al.(2016)Li, Wang, Shi, Liu, and Hou]{li2016revisiting}
Yanghao Li, Naiyan Wang, Jianping Shi, Jiaying Liu, and Xiaodi Hou.
\newblock Revisiting batch normalization for practical domain adaptation.
\newblock \emph{arXiv preprint arXiv:1603.04779}, 2016.

\bibitem[Li et~al.(2018{\natexlab{c}})Li, Wang, Shi, Hou, and
  Liu]{li2018adaptive}
Yanghao Li, Naiyan Wang, Jianping Shi, Xiaodi Hou, and Jiaying Liu.
\newblock Adaptive batch normalization for practical domain adaptation.
\newblock \emph{Pattern Recognition}, 80:\penalty0 109--117,
  2018{\natexlab{c}}.

\bibitem[Li et~al.(2019{\natexlab{b}})Li, Baldwin, and Cohn]{li2018s}
Yitong Li, Timothy Baldwin, and Trevor Cohn.
\newblock What’s in a domain? learning domain-robust text representations
  using adversarial training.
\newblock In \emph{Proceedings of NAACL-HLT}, pages 474--479,
  2019{\natexlab{b}}.

\bibitem[Li et~al.(2021{\natexlab{b}})Li, Dai, Ma, Liu, Chen, Wu, He, Kitani,
  and Vadja]{li2021cross}
Yu-Jhe Li, Xiaoliang Dai, Chih-Yao Ma, Yen-Cheng Liu, Kan Chen, Bichen Wu,
  Zijian He, Kris Kitani, and Peter Vadja.
\newblock Cross-domain object detection via adaptive self-training.
\newblock \emph{arXiv preprint arXiv:2111.13216}, 2021{\natexlab{b}}.

\bibitem[Lian et~al.(2019)Lian, Lv, Duan, and Gong]{lian2019constructing}
Qing Lian, Fengmao Lv, Lixin Duan, and Boqing Gong.
\newblock Constructing self-motivated pyramid curriculums for cross-domain
  semantic segmentation: A non-adversarial approach.
\newblock \emph{ICCV}, 2019.

\bibitem[Liang et~al.(2018)Liang, Li, and Srikant]{Liang2018Enhancing}
Shiyu Liang, Yixuan Li, and R.~Srikant.
\newblock Enhancing the reliability of out-of-distribution image detection in
  neural networks.
\newblock \emph{ICLR}, 2018.

\bibitem[Lim et~al.(2020)Lim, Lee, Carbonell, and Poibeau]{lim2020semi}
KyungTae Lim, Jay~Yoon Lee, Jaime Carbonell, and Thierry Poibeau.
\newblock Semi-supervised learning on meta structure: Multi-task tagging and
  parsing in low-resource scenarios.
\newblock In \emph{Proceedings of the AAAI Conference on Artificial
  Intelligence}, volume~34, pages 8344--8351, 2020.

\bibitem[Lipton et~al.(2018)Lipton, Wang, and Smola]{lipton2018detecting}
Zachary Lipton, Yu-Xiang Wang, and Alexander Smola.
\newblock Detecting and correcting for label shift with black box predictors.
\newblock In \emph{ICML}, 2018.

\bibitem[Liu et~al.(2017)Liu, Vijaya~Kumar, You, and Jia]{liu2017adaptive}
Xiaofeng Liu, BVK Vijaya~Kumar, Jane You, and Ping Jia.
\newblock Adaptive deep metric learning for identity-aware facial expression
  recognition.
\newblock In \emph{CVPR Workshops}, pages 20--29, 2017.

\bibitem[Liu et~al.(2018)Liu, Zou, Kong, Diao, Yan, Wang, Li, Jia, and
  You]{liu2018data}
Xiaofeng Liu, Yang Zou, Lingsheng Kong, Zhihui Diao, Junliang Yan, Jun Wang,
  Site Li, Ping Jia, and Jane You.
\newblock Data augmentation via latent space interpolation for image
  classification.
\newblock In \emph{24th International Conference on Pattern Recognition
  (ICPR)}, pages 728--733, 2018.

\bibitem[Liu et~al.(2019{\natexlab{a}})Liu, Han, Qiao, Ge, Li, and
  Lu]{liu2019unimodal}
Xiaofeng Liu, Xu~Han, Yukai Qiao, Yi~Ge, Site Li, and Jun Lu.
\newblock Unimodal-uniform constrained wasserstein training for medical
  diagnosis.
\newblock In \emph{Proceedings of the IEEE/CVF International Conference on
  Computer Vision Workshops}, pages 0--0, 2019{\natexlab{a}}.

\bibitem[Liu et~al.(2019{\natexlab{b}})Liu, Li, Kong, Xie, Jia, You, and
  Kumar]{liu2019feature}
Xiaofeng Liu, Site Li, Lingsheng Kong, Wanqing Xie, Ping Jia, Jane You, and BVK
  Kumar.
\newblock Feature-level frankenstein: Eliminating variations for discriminative
  recognition.
\newblock In \emph{CVPR}, 2019{\natexlab{b}}.

\bibitem[Liu et~al.(2019{\natexlab{c}})Liu, Zou, Che, Ding, Jia, You, and
  Kumar]{liu2019conservative}
Xiaofeng Liu, Yang Zou, Tong Che, Peng Ding, Ping Jia, Jane You, and BVK Kumar.
\newblock Conservative wasserstein training for pose estimation.
\newblock In \emph{Proceedings of the IEEE/CVF International Conference on
  Computer Vision}, pages 8262--8272, 2019{\natexlab{c}}.

\bibitem[Liu et~al.(2020{\natexlab{a}})Liu, Fan, Kong, Diao, Xie, Lu, and
  You]{liu2020unimodal}
Xiaofeng Liu, Fangfang Fan, Lingsheng Kong, Zhihui Diao, Wanqing Xie, Jun Lu,
  and Jane You.
\newblock Unimodal regularized neuron stick-breaking for ordinal
  classification.
\newblock \emph{Neurocomputing}, 2020{\natexlab{a}}.

\bibitem[Liu et~al.(2020{\natexlab{b}})Liu, Han, Bai, Ge, Wang, Han, Li, You,
  and Lu]{liu2020importance}
Xiaofeng Liu, Yuzhuo Han, Song Bai, Yi~Ge, Tianxing Wang, Xu~Han, Site Li, Jane
  You, and Jun Lu.
\newblock Importance-aware semantic segmentation in self-driving with discrete
  wasserstein training.
\newblock In \emph{AAAI}, pages 11629--11636, 2020{\natexlab{b}}.

\bibitem[Liu et~al.(2020{\natexlab{c}})Liu, Ji, You, Fakhri, and
  Woo]{liu2020severity}
Xiaofeng Liu, Wenxuan Ji, Jane You, Georges~El Fakhri, and Jonghye Woo.
\newblock Severity-aware semantic segmentation with reinforced wasserstein
  training.
\newblock In \emph{Proceedings of the IEEE/CVF Conference on Computer Vision
  and Pattern Recognition}, pages 12566--12575, 2020{\natexlab{c}}.

\bibitem[Liu et~al.(2020{\natexlab{d}})Liu, Lu, Liu, Bai, Li, and
  You]{liu2020wasserstein}
Xiaofeng Liu, Yunhong Lu, Xiongchang Liu, Song Bai, Site Li, and Jane You.
\newblock Wasserstein loss with alternative reinforcement learning for
  severity-aware semantic segmentation.
\newblock \emph{IEEE Transactions on Intelligent Transportation Systems},
  2020{\natexlab{d}}.

\bibitem[Liu et~al.(2021{\natexlab{a}})Liu, Chao, You, Kuo, and
  Vijayakumar]{liu2021mutualpami}
Xiaofeng Liu, Yang Chao, Jane~J You, C-C~Jay Kuo, and Bhagavatula Vijayakumar.
\newblock Mutual information regularized feature-level frankenstein for
  discriminative recognition.
\newblock \emph{IEEE T-PAMI}, 2021{\natexlab{a}}.

\bibitem[Liu et~al.(2021{\natexlab{b}})Liu, Guo, Li, Xing, You, Kuo, El~Fakhri,
  and Woo]{liu2021adversarial}
Xiaofeng Liu, Zhenhua Guo, Site Li, Fangxu Xing, Jane You, C-C~Jay Kuo, Georges
  El~Fakhri, and Jonghye Woo.
\newblock Adversarial unsupervised domain adaptation with conditional and label
  shift: Infer, align and iterate.
\newblock In \emph{Proceedings of the IEEE/CVF International Conference on
  Computer Vision}, pages 10367--10376, 2021{\natexlab{b}}.

\bibitem[Liu et~al.(2021{\natexlab{c}})Liu, Hu, Jin, Han, Xing, Ouyang, Lu,
  El~Fakhri, and Woo]{liu2021Generalization}
Xiaofeng Liu, Bo~Hu, Linghao Jin, Xu~Han, Fangxu Xing, Jinsong Ouyang, Jun Lu,
  Georges El~Fakhri, and Jonghye Woo.
\newblock Domain generalization under conditional and label shifts via
  variational bayesian inference.
\newblock In \emph{IJCAI}, 2021{\natexlab{c}}.

\bibitem[Liu et~al.(2021{\natexlab{d}})Liu, Hu, Jin, Han, Xing, Ouyang, Lu,
  Fakhri, and Woo]{liu2021domain}
Xiaofeng Liu, Bo~Hu, Linghao Jin, Xu~Han, Fangxu Xing, Jinsong Ouyang, Jun Lu,
  Georges~EL Fakhri, and Jonghye Woo.
\newblock Domain generalization under conditional and label shifts via
  variational {B}ayesian inference.
\newblock \emph{IJCAI}, 2021{\natexlab{d}}.

\bibitem[Liu et~al.(2021{\natexlab{e}})Liu, Hu, Liu, Lu, You, and
  Kong]{liu2021energy}
Xiaofeng Liu, Bo~Hu, Xiongchang Liu, Jun Lu, Jane You, and Lingsheng Kong.
\newblock Energy-constrained self-training for unsupervised domain adaptation.
\newblock In \emph{2020 25th International Conference on Pattern Recognition
  (ICPR)}, pages 7515--7520. IEEE, 2021{\natexlab{e}}.

\bibitem[Liu et~al.(2021{\natexlab{f}})Liu, Jin, Han, Lu, You, and
  Kong]{liu2021identity}
Xiaofeng Liu, Linghao Jin, Xu~Han, Jun Lu, Jane You, and Lingsheng Kong.
\newblock Identity-aware facial expression recognition in compressed video.
\newblock In \emph{2020 25th International Conference on Pattern Recognition
  (ICPR)}, pages 7508--7514. IEEE, 2021{\natexlab{f}}.

\bibitem[Liu et~al.(2021{\natexlab{g}})Liu, Jin, Han, and You]{liu2021mutualpr}
Xiaofeng Liu, Linghao Jin, Xu~Han, and Jane You.
\newblock Mutual information regularized identity-aware facial expression
  recognition in compressed video.
\newblock \emph{Pattern Recognition}, 2021{\natexlab{g}}.

\bibitem[Liu et~al.(2021{\natexlab{h}})Liu, Li, Ge, Ye, You, and
  Lu]{liu2021recursively}
Xiaofeng Liu, Site Li, Yubin Ge, Pengyi Ye, Jane You, and Jun Lu.
\newblock Recursively conditional gaussian for ordinal unsupervised domain
  adaptation.
\newblock In \emph{Proceedings of the IEEE/CVF International Conference on
  Computer Vision}, pages 764--773, 2021{\natexlab{h}}.

\bibitem[Liu et~al.(2021{\natexlab{i}})Liu, Liu, Hu, Ji, Xing, Lu, You, Kuo,
  Fakhri, and Woo]{liu2021subtype}
Xiaofeng Liu, Xiongchang Liu, Bo~Hu, Wenxuan Ji, Fangxu Xing, Jun Lu, Jane You,
  C-C~Jay Kuo, Georges~El Fakhri, and Jonghye Woo.
\newblock Subtype-aware unsupervised domain adaptation for medical diagnosis.
\newblock \emph{AAAI}, 2021{\natexlab{i}}.

\bibitem[Liu et~al.(2021{\natexlab{j}})Liu, Xing, El~Fakhri, and
  Woo]{liu2021unified}
Xiaofeng Liu, Fangxu Xing, Georges El~Fakhri, and Jonghye Woo.
\newblock A unified conditional disentanglement framework for multimodal brain
  mr image translation.
\newblock In \emph{2021 IEEE 18th International Symposium on Biomedical Imaging
  (ISBI)}, pages 10--14. IEEE, 2021{\natexlab{j}}.

\bibitem[Liu et~al.(2021{\natexlab{k}})Liu, Xing, Prince, Carass, Stone,
  El~Fakhri, and Woo]{liu2021dual}
Xiaofeng Liu, Fangxu Xing, Jerry~L Prince, Aaron Carass, Maureen Stone, Georges
  El~Fakhri, and Jonghye Woo.
\newblock Dual-cycle constrained bijective vae-gan for tagged-to-cine magnetic
  resonance image synthesis.
\newblock In \emph{2021 IEEE 18th International Symposium on Biomedical Imaging
  (ISBI)}, pages 1448--1452. IEEE, 2021{\natexlab{k}}.

\bibitem[Liu et~al.(2021{\natexlab{l}})Liu, Xing, Stone, Zhuo, Reese, Prince,
  El~Fakhri, and Woo]{liu2021generative}
Xiaofeng Liu, Fangxu Xing, Maureen Stone, Jiachen Zhuo, Timothy Reese, Jerry~L
  Prince, Georges El~Fakhri, and Jonghye Woo.
\newblock Generative self-training for cross-domain unsupervised tagged-to-cine
  mri synthesis.
\newblock In \emph{International Conference on Medical Image Computing and
  Computer-Assisted Intervention}, pages 138--148. Springer,
  2021{\natexlab{l}}.

\bibitem[Liu et~al.(2021{\natexlab{m}})Liu, Xing, Yang, El~Fakhri, and
  Woo]{liu2021adapting}
Xiaofeng Liu, Fangxu Xing, Chao Yang, Georges El~Fakhri, and Jonghye Woo.
\newblock Adapting off-the-shelf source segmenter for target medical image
  segmentation.
\newblock In \emph{International Conference on Medical Image Computing and
  Computer-Assisted Intervention}, pages 549--559. Springer,
  2021{\natexlab{m}}.

\bibitem[Liu et~al.(2022{\natexlab{a}})Liu, Li, Ge, Ye, You, and
  Lu]{liu2022recursively}
Xiaofeng Liu, Site Li, Yubin Ge, Pengyi Ye, Jane You, and Jun Lu.
\newblock Ordinal unsupervised domain adaptation with recursively conditional
  gaussian imposed variational disentanglement.
\newblock In \emph{IEEE Transactions on Pattern Analysis and Machine
  Intelligence}, 2022{\natexlab{a}}.

\bibitem[Liu et~al.(2022{\natexlab{b}})Liu, Xing, El~Fakhri, and
  Woo]{liu2022Off-the-Shelf}
Xiaofeng Liu, Fangxu Xing, Georges El~Fakhri, and Jonghye Woo.
\newblock Memory consistent unsupervised off-the-shelf model adaptation for
  source-relaxed medical image segmentation.
\newblock In \emph{Medical Image Analysis}, 2022{\natexlab{b}}.

\bibitem[Liu et~al.(2022{\natexlab{c}})Liu, Xing, Fakhri, and Woo]{liu2022self}
Xiaofeng Liu, Fangxu Xing, Georges~El Fakhri, and Jonghye Woo.
\newblock Self-semantic contour adaptation for cross modality brain tumor
  segmentation.
\newblock \emph{IEEE International Symposium on Biomedical Imaging (ISBI)},
  2022{\natexlab{c}}.

\bibitem[Liu et~al.(2022{\natexlab{d}})Liu, Xing, Marin, Fakhri, and
  Woo]{liu2022variational}
Xiaofeng Liu, Fangxu Xing, Thibault Marin, Georges~El Fakhri, and Jonghye Woo.
\newblock Variational inference for quantifying inter-observer variability in
  segmentation of anatomical structures.
\newblock \emph{arXiv preprint arXiv:2201.07106}, 2022{\natexlab{d}}.

\bibitem[Liu et~al.(2022{\natexlab{e}})Liu, Xing, Prince, Stone, Fakhri, and
  Woo]{liu2022structure}
Xiaofeng Liu, Fangxu Xing, Jerry~L Prince, Maureen Stone, Georges~El Fakhri,
  and Jonghye Woo.
\newblock Structure-aware unsupervised tagged-to-cine mri synthesis with self
  disentanglement.
\newblock \emph{arXiv preprint arXiv:2202.12474}, 2022{\natexlab{e}}.

\bibitem[Liu et~al.(2022{\natexlab{f}})Liu, Xing, Shusharina, Lim, Kuo, Fakhri,
  and Woo]{liu2022ACT}
Xiaofeng Liu, Fangxu Xing, Nadya Shusharina, Ruth Lim, C-C~Jay Kuo, Georges~El
  Fakhri, and Jonghye Woo.
\newblock Act: Semi-supervised domain-adaptive medical image segmentation with
  asymmetric co-training.
\newblock In \emph{MICCAI}, 2022{\natexlab{f}}.

\bibitem[Liu et~al.(2022{\natexlab{g}})Liu, Xing, You, Lu, Kuo, Fakhri, and
  Woo]{liu2022subtype}
Xiaofeng Liu, Fangxu Xing, Jane You, Jun Lu, C-C~Jay Kuo, Georges~El Fakhri,
  and Jonghye Woo.
\newblock Subtype-aware dynamic unsupervised domain adaptation.
\newblock \emph{IEEE Transactions on Neural Networks and Learning Systems},
  2022{\natexlab{g}}.

\bibitem[Liu et~al.(2022{\natexlab{h}})Liu, Yoo, Xing, Kuo, El~Fakhri, and
  Woo]{liu2022Unsupervised}
Xiaofeng Liu, Chaehwa Yoo, Fangxu Xing, C.-C.~Jay Kuo, Georges El~Fakhri, and
  Jonghye Woo.
\newblock Unsupervised domain adaptation for segmentation with black-box source
  model.
\newblock \emph{SPIE Medical Imaging 2022: Image Processing},
  2022{\natexlab{h}}.

\bibitem[Liu et~al.(2022{\natexlab{i}})Liu, Yoo, Xing, Kuo, El~Fakhri, and
  Woo]{liu2022frontier}
Xiaofeng Liu, Chaehwa Yoo, Fangxu Xing, C.-C.~Jay Kuo, Georges El~Fakhri, and
  Jonghye Woo.
\newblock Unsupervised black-box model domain adaptation for brain tumor
  segmentation.
\newblock \emph{Frontiers in Neuroscience}, 2022{\natexlab{i}}.

\bibitem[Liu et~al.(2019{\natexlab{d}})Liu, Lu, Li, Yang, and Yao]{liu2019deep}
Yang Liu, Zhaoyang Lu, Jing Li, Tao Yang, and Chao Yao.
\newblock Deep image-to-video adaptation and fusion networks for action
  recognition.
\newblock \emph{IEEE Transactions on Image Processing}, 29:\penalty0
  3168--3182, 2019{\natexlab{d}}.

\bibitem[Long et~al.(2015)Long, Cao, Wang, and Jordan]{long2015learning}
Mingsheng Long, Yue Cao, Jianmin Wang, and Michael~I Jordan.
\newblock Learning transferable features with deep adaptation networks.
\newblock \emph{ICML}, 2015.

\bibitem[Long et~al.(2016)Long, Zhu, Wang, and Jordan]{long2016unsupervised}
Mingsheng Long, Han Zhu, Jianmin Wang, and Michael~I Jordan.
\newblock Unsupervised domain adaptation with residual transfer networks.
\newblock In \emph{Advances in Neural Information Processing Systems}, pages
  136--144, 2016.

\bibitem[Long et~al.(2017)Long, Zhu, Wang, and Jordan]{long2017deep}
Mingsheng Long, Han Zhu, Jianmin Wang, and Michael~I Jordan.
\newblock Deep transfer learning with joint adaptation networks.
\newblock In \emph{Proceedings of the 34th International Conference on Machine
  Learning-Volume 70}, pages 2208--2217. JMLR. org, 2017.

\bibitem[Long et~al.(2018)Long, Cao, Wang, and Jordan]{long2018conditional}
Mingsheng Long, Zhangjie Cao, Jianmin Wang, and Michael~I Jordan.
\newblock Conditional adversarial domain adaptation.
\newblock In \emph{Advances in Neural Information Processing Systems}, pages
  1647--1657, 2018.

\bibitem[Lu et~al.(2021)Lu, Xiao, Sun, Han, and Wang]{lu2021new}
Nannan Lu, Hanhan Xiao, Yanjing Sun, Min Han, and Yanfen Wang.
\newblock A new method for intelligent fault diagnosis of machines based on
  unsupervised domain adaptation.
\newblock \emph{Neurocomputing}, 427:\penalty0 96--109, 2021.

\bibitem[Lucas et~al.(2020)Lucas, Pelletier, Schmidt, Webb, and
  Petitjean]{lucas2020unsupervised}
Benjamin Lucas, Charlotte Pelletier, Daniel Schmidt, Geoffrey~I Webb, and
  Fran{\c{c}}ois Petitjean.
\newblock Unsupervised domain adaptation techniques for classification of
  satellite image time series.
\newblock In \emph{IGARSS 2020-2020 IEEE International Geoscience and Remote
  Sensing Symposium}, pages 1074--1077. IEEE, 2020.

\bibitem[Ma and Zhang(2021)]{ma2021multi}
Yuchi Ma and Zhou Zhang.
\newblock Multi-source unsupervised domain adaptation on corn yield prediction.
\newblock In \emph{AI for Agriculture and Food Systems}, 2021.

\bibitem[Mahapatra et~al.(2021)Mahapatra, Tennakoon, et~al.]{mahapatra2021gcn}
Dwarikanath Mahapatra, Ruwan Tennakoon, et~al.
\newblock Gcn based unsupervised domain adaptation with feature disentanglement
  for medical image classification.
\newblock 2021.

\bibitem[Mancini et~al.(2018{\natexlab{a}})Mancini, Bulo, Caputo, and
  Ricci]{mancini2018best}
Massimiliano Mancini, Samuel~Rota Bulo, Barbara Caputo, and Elisa Ricci.
\newblock Best sources forward: domain generalization through source-specific
  nets.
\newblock In \emph{ICIP}, 2018{\natexlab{a}}.

\bibitem[Mancini et~al.(2018{\natexlab{b}})Mancini, Porzi, Bulo, Caputo, and
  Ricci]{mancini2018boosting}
Massimiliano Mancini, Lorenzo Porzi, Samuel~Rota Bulo, Barbara Caputo, and
  Elisa Ricci.
\newblock Boosting domain adaptation by discovering latent domains.
\newblock In \emph{Proceedings of the IEEE Conference on Computer Vision and
  Pattern Recognition}, pages 3771--3780, 2018{\natexlab{b}}.

\bibitem[Mancini et~al.(2019)Mancini, Bulo, Caputo, and
  Ricci]{mancini2019adagraph}
Massimiliano Mancini, Samuel~Rota Bulo, Barbara Caputo, and Elisa Ricci.
\newblock Adagraph: Unifying predictive and continuous domain adaptation
  through graphs.
\newblock In \emph{Proceedings of the IEEE/CVF Conference on Computer Vision
  and Pattern Recognition}, pages 6568--6577, 2019.

\bibitem[Manohar et~al.(2018)Manohar, Ghahremani, Povey, and
  Khudanpur]{manohar2018teacher}
Vimal Manohar, Pegah Ghahremani, Daniel Povey, and Sanjeev Khudanpur.
\newblock A teacher-student learning approach for unsupervised domain
  adaptation of sequence-trained asr models.
\newblock In \emph{2018 IEEE Spoken Language Technology Workshop (SLT)}, pages
  250--257. IEEE, 2018.

\bibitem[Maria~Carlucci et~al.(2017)Maria~Carlucci, Porzi, Caputo, Ricci, and
  Rota~Bulo]{maria2017autodial}
Fabio Maria~Carlucci, Lorenzo Porzi, Barbara Caputo, Elisa Ricci, and Samuel
  Rota~Bulo.
\newblock Autodial: Automatic domain alignment layers.
\newblock In \emph{Proceedings of the IEEE International Conference on Computer
  Vision}, pages 5067--5075, 2017.

\bibitem[Matsuura and Harada(2020)]{matsuura2019domain}
Toshihiko Matsuura and Tatsuya Harada.
\newblock Domain generalization using a mixture of multiple latent domains.
\newblock \emph{AAAI}, 2020.

\bibitem[Mei et~al.(2020{\natexlab{a}})Mei, Zhu, Zou, and
  Zhang]{ke2020instance}
Ke~Mei, Chuang Zhu, Jiaqi Zou, and Shanghang Zhang.
\newblock Instance adaptive self-training for unsupervised domain adaptation.
\newblock \emph{ECCV}, 2020{\natexlab{a}}.

\bibitem[Mei et~al.(2020{\natexlab{b}})Mei, Zhu, Zou, and
  Zhang]{mei2020instance}
Ke~Mei, Chuang Zhu, Jiaqi Zou, and Shanghang Zhang.
\newblock Instance adaptive self-training for unsupervised domain adaptation.
\newblock \emph{ECCV}, 2020{\natexlab{b}}.

\bibitem[Mekhazni et~al.(2020)Mekhazni, Bhuiyan, Ekladious, and
  Granger]{mekhazni2020unsupervised}
Djebril Mekhazni, Amran Bhuiyan, George Ekladious, and Eric Granger.
\newblock Unsupervised domain adaptation in the dissimilarity space for person
  re-identification.
\newblock In \emph{European Conference on Computer Vision}, pages 159--174.
  Springer, 2020.

\bibitem[Mengqiu et~al.(2022)Mengqiu, Ming, Jun, Zhang, Yubo, and
  Zhanyu]{mengqiu2022sea}
XU~Mengqiu, WU~Ming, GUO Jun, Chuang Zhang, WANG Yubo, and MA~Zhanyu.
\newblock Sea fog detection based on unsupervised domain adaptation.
\newblock \emph{Chinese Journal of Aeronautics}, 35\penalty0 (4):\penalty0
  415--425, 2022.

\bibitem[Michau and Fink(2021)]{michau2021unsupervised}
Gabriel Michau and Olga Fink.
\newblock Unsupervised transfer learning for anomaly detection: Application to
  complementary operating condition transfer.
\newblock \emph{Knowledge-Based Systems}, 216:\penalty0 106816, 2021.

\bibitem[Montesuma and Mboula(2021)]{montesuma2021wasserstein}
Eduardo~Fernandes Montesuma and Fred Maurice~Ngole Mboula.
\newblock Wasserstein barycenter for multi-source domain adaptation.
\newblock In \emph{Proceedings of the IEEE/CVF Conference on Computer Vision
  and Pattern Recognition}, pages 16785--16793, 2021.

\bibitem[Morerio et~al.(2017)Morerio, Cavazza, and Murino]{morerio2017minimal}
Pietro Morerio, Jacopo Cavazza, and Vittorio Murino.
\newblock Minimal-entropy correlation alignment for unsupervised deep domain
  adaptation.
\newblock \emph{arXiv preprint arXiv:1711.10288}, 2017.

\bibitem[Motiian et~al.(2017)Motiian, Jones, Iranmanesh, and
  Doretto]{motiian2017few}
Saeid Motiian, Quinn Jones, Seyed Iranmanesh, and Gianfranco Doretto.
\newblock Few-shot adversarial domain adaptation.
\newblock \emph{Advances in neural information processing systems}, 30, 2017.

\bibitem[Mou et~al.(2021)Mou, Wei, Wang, Luo, He, Zhang, Xu, Luo, and
  Gao]{mou2021unsupervised}
Quanzheng Mou, Longsheng Wei, Conghao Wang, Dapeng Luo, Songze He, Jing Zhang,
  Huimin Xu, Chen Luo, and Changxin Gao.
\newblock Unsupervised domain-adaptive scene-specific pedestrian detection for
  static video surveillance.
\newblock \emph{Pattern Recognition}, 118:\penalty0 108038, 2021.

\bibitem[Naik and Rose(2020)]{naik2020towards}
Aakanksha Naik and Carolyn Rose.
\newblock Towards open domain event trigger identification using adversarial
  domain adaptation.
\newblock In \emph{Proceedings of the 58th Annual Meeting of the Association
  for Computational Linguistics}, pages 7618--7624, 2020.

\bibitem[Nasim et~al.(2022)Nasim, Maiti, Srivastava, Singh, and
  Mei]{nasim2022seismic}
M~Quamer Nasim, Tannistha Maiti, Ayush Srivastava, Tarry Singh, and Jie Mei.
\newblock Seismic facies analysis: a deep domain adaptation approach.
\newblock \emph{IEEE Transactions on Geoscience and Remote Sensing},
  60:\penalty0 1--16, 2022.

\bibitem[Ning et~al.(2021)Ning, Bian, Wei, Yu, Yuan, Wang, Guo, Ma, and
  Zheng]{ning2021new}
Munan Ning, Cheng Bian, Dong Wei, Shuang Yu, Chenglang Yuan, Yaohua Wang, Yang
  Guo, Kai Ma, and Yefeng Zheng.
\newblock A new bidirectional unsupervised domain adaptation segmentation
  framework.
\newblock In \emph{International Conference on Information Processing in
  Medical Imaging}, pages 492--503. Springer, 2021.

\bibitem[Niu et~al.(2020)Niu, Chen, Liu, Zhou, and Shu]{niu2020deep}
Lisha Niu, Chao Chen, Hui Liu, Shuwang Zhou, and Minglei Shu.
\newblock A deep-learning approach to ecg classification based on adversarial
  domain adaptation.
\newblock In \emph{Healthcare}, volume~8, page 437. Multidisciplinary Digital
  Publishing Institute, 2020.

\bibitem[Nyborg et~al.(2022)Nyborg, Pelletier, Lef{\`e}vre, and
  Assent]{nyborg2022timematch}
Joachim Nyborg, Charlotte Pelletier, S{\'e}bastien Lef{\`e}vre, and Ira Assent.
\newblock Timematch: Unsupervised cross-region adaptation by temporal shift
  estimation.
\newblock \emph{ISPRS Journal of Photogrammetry and Remote Sensing},
  188:\penalty0 301--313, 2022.

\bibitem[Ouyang et~al.(2019)Ouyang, Kamnitsas, Biffi, Duan, and
  Rueckert]{ouyang2019data}
Cheng Ouyang, Konstantinos Kamnitsas, Carlo Biffi, Jinming Duan, and Daniel
  Rueckert.
\newblock Data efficient unsupervised domain adaptation for cross-modality
  image segmentation.
\newblock In \emph{International Conference on Medical Image Computing and
  Computer-Assisted Intervention}, pages 669--677. Springer, 2019.

\bibitem[Oza et~al.(2021)Oza, Sindagi, VS, and Patel]{oza2021unsupervised}
Poojan Oza, Vishwanath~A Sindagi, Vibashan VS, and Vishal~M Patel.
\newblock Unsupervised domain adaptation of object detectors: A survey.
\newblock \emph{arXiv preprint arXiv:2105.13502}, 2021.

\bibitem[Pan et~al.(2020)Pan, Cao, Adeli, and Niebles]{pan2020adversarial}
Boxiao Pan, Zhangjie Cao, Ehsan Adeli, and Juan~Carlos Niebles.
\newblock Adversarial cross-domain action recognition with co-attention.
\newblock In \emph{Proceedings of the AAAI Conference on Artificial
  Intelligence}, volume~34, pages 11815--11822, 2020.

\bibitem[Pan and Yang(2009)]{pan2009survey}
Sinno~Jialin Pan and Qiang Yang.
\newblock A survey on transfer learning.
\newblock \emph{IEEE Transactions on knowledge and data engineering},
  22\penalty0 (10):\penalty0 1345--1359, 2009.

\bibitem[Pan et~al.(2019)Pan, Yao, Li, Wang, Ngo, and
  Mei]{pan2019transferrable}
Yingwei Pan, Ting Yao, Yehao Li, Yu~Wang, Chong-Wah Ngo, and Tao Mei.
\newblock Transferrable prototypical networks for unsupervised domain
  adaptation.
\newblock In \emph{Proceedings of the IEEE Conference on Computer Vision and
  Pattern Recognition}, pages 2239--2247, 2019.

\bibitem[Panareda~Busto and Gall(2017)]{panareda2017open}
Pau Panareda~Busto and Juergen Gall.
\newblock Open set domain adaptation.
\newblock In \emph{Proceedings of the IEEE International Conference on Computer
  Vision}, pages 754--763, 2017.

\bibitem[Peng et~al.(2019{\natexlab{a}})Peng, Sun, Ma, and
  Du]{peng2019discriminative}
Jiangtao Peng, Weiwei Sun, Li~Ma, and Qian Du.
\newblock Discriminative transfer joint matching for domain adaptation in
  hyperspectral image classification.
\newblock \emph{IEEE Geoscience and Remote Sensing Letters}, 16\penalty0
  (6):\penalty0 972--976, 2019{\natexlab{a}}.

\bibitem[Peng et~al.(2017)Peng, Usman, Kaushik, Hoffman, Wang, and
  Saenko]{visda2017}
Xingchao Peng, Ben Usman, Neela Kaushik, Judy Hoffman, Dequan Wang, and Kate
  Saenko.
\newblock Visda: The visual domain adaptation challenge, 2017.

\bibitem[Peng et~al.(2019{\natexlab{b}})Peng, Bai, Xia, Huang, Saenko, and
  Wang]{peng2019moment}
Xingchao Peng, Qinxun Bai, Xide Xia, Zijun Huang, Kate Saenko, and Bo~Wang.
\newblock Moment matching for multi-source domain adaptation.
\newblock \emph{ICCV}, 2019{\natexlab{b}}.

\bibitem[Perone et~al.(2019)Perone, Ballester, Barros, and
  Cohen-Adad]{perone2019unsupervised}
Christian~S Perone, Pedro Ballester, Rodrigo~C Barros, and Julien Cohen-Adad.
\newblock Unsupervised domain adaptation for medical imaging segmentation with
  self-ensembling.
\newblock \emph{NeuroImage}, 194:\penalty0 1--11, 2019.

\bibitem[Pimentel et~al.(2014)Pimentel, Clifton, Lei, and
  Tarassenko]{Pimentel2014A}
Marco A.~F. Pimentel, David~A. Clifton, Clifton Lei, and Lionel Tarassenko.
\newblock A review of novelty detection.
\newblock \emph{Signal Processing}, 99\penalty0 (6):\penalty0 215--249, 2014.

\bibitem[Porwal et~al.(2018)Porwal, Pachade, Kamble, Kokare, Deshmukh,
  Sahasrabuddhe, and Meriaudeau]{porwal2018indian}
Prasanna Porwal, Samiksha Pachade, Ravi Kamble, Manesh Kokare, Girish Deshmukh,
  Vivek Sahasrabuddhe, and Fabrice Meriaudeau.
\newblock Indian diabetic retinopathy image dataset (idrid): a database for
  diabetic retinopathy screening research.
\newblock \emph{Data}, 3\penalty0 (3):\penalty0 25, 2018.

\bibitem[Purushotham et~al.(2017)Purushotham, Carvalho, Nilanon, and
  Liu]{Purushotham2017VariationalRA}
S.~Purushotham, Wilka Carvalho, Tanachat Nilanon, and Yan Liu.
\newblock Variational recurrent adversarial deep domain adaptation.
\newblock In \emph{ICLR}, 2017.

\bibitem[Ragab et~al.(2020)Ragab, Chen, Wu, Foo, Kwoh, Yan, and
  Li]{ragab2020contrastive}
Mohamed Ragab, Zhenghua Chen, Min Wu, Chuan~Sheng Foo, Chee~Keong Kwoh, Ruqiang
  Yan, and Xiaoli Li.
\newblock Contrastive adversarial domain adaptation for machine remaining
  useful life prediction.
\newblock \emph{IEEE Transactions on Industrial Informatics}, 17\penalty0
  (8):\penalty0 5239--5249, 2020.

\bibitem[Ramponi and Plank(2020)]{ramponi2020neural}
Alan Ramponi and Barbara Plank.
\newblock Neural unsupervised domain adaptation in nlp---a survey.
\newblock \emph{arXiv preprint arXiv:2006.00632}, 2020.

\bibitem[Raza and Samothrakis(2019)]{raza2019bagging}
Haider Raza and Spyridon Samothrakis.
\newblock Bagging adversarial neural networks for domain adaptation in
  non-stationary eeg.
\newblock In \emph{2019 International Joint Conference on Neural Networks
  (IJCNN)}, pages 1--7. IEEE, 2019.

\bibitem[Richter et~al.(2016)Richter, Vineet, Roth, and
  Koltun]{richter2016playing}
Stephan~R Richter, Vibhav Vineet, Stefan Roth, and Vladlen Koltun.
\newblock Playing for data: Ground truth from computer games.
\newblock In \emph{ECCV}, 2016.

\bibitem[Rios et~al.(2018)Rios, Kavuluru, and Lu]{rios2018generalizing}
Anthony Rios, Ramakanth Kavuluru, and Zhiyong Lu.
\newblock Generalizing biomedical relation classification with neural
  adversarial domain adaptation.
\newblock \emph{Bioinformatics}, 34\penalty0 (17):\penalty0 2973--2981, 2018.

\bibitem[Rocha and Cardoso(2019)]{rocha2019comparative}
Gil Rocha and Henrique~Lopes Cardoso.
\newblock A comparative analysis of unsupervised language adaptation methods.
\newblock In \emph{Proceedings of the 2nd Workshop on Deep Learning Approaches
  for Low-Resource NLP (DeepLo 2019)}, pages 11--21, 2019.

\bibitem[Rodriguez and Mikolajczyk(2019)]{rodriguez2019domain}
Adrian~Lopez Rodriguez and Krystian Mikolajczyk.
\newblock Domain adaptation for object detection via style consistency.
\newblock \emph{arXiv preprint arXiv:1911.10033}, 2019.

\bibitem[Rotman and Reichart(2019)]{rotman2019deep}
Guy Rotman and Roi Reichart.
\newblock Deep contextualized self-training for low resource dependency
  parsing.
\newblock \emph{Transactions of the Association for Computational Linguistics},
  7:\penalty0 695--713, 2019.

\bibitem[RoyChowdhury et~al.(2019)RoyChowdhury, Chakrabarty, Singh, Jin, Jiang,
  Cao, and Learned-Miller]{roychowdhury2019automatic}
Aruni RoyChowdhury, Prithvijit Chakrabarty, Ashish Singh, SouYoung Jin, Huaizu
  Jiang, Liangliang Cao, and Erik Learned-Miller.
\newblock Automatic adaptation of object detectors to new domains using
  self-training.
\newblock In \emph{Proceedings of the IEEE/CVF Conference on Computer Vision
  and Pattern Recognition}, pages 780--790, 2019.

\bibitem[Royer and Lampert(2015)]{royer2015classifier}
Amelie Royer and Christoph~H Lampert.
\newblock Classifier adaptation at prediction time.
\newblock In \emph{Proceedings of the IEEE Conference on Computer Vision and
  Pattern Recognition}, pages 1401--1409, 2015.

\bibitem[Rozantsev et~al.(2018)Rozantsev, Salzmann, and
  Fua]{rozantsev2018beyond}
Artem Rozantsev, Mathieu Salzmann, and Pascal Fua.
\newblock Beyond sharing weights for deep domain adaptation.
\newblock \emph{IEEE transactions on pattern analysis and machine
  intelligence}, 41\penalty0 (4):\penalty0 801--814, 2018.

\bibitem[Saenko et~al.(2010)Saenko, Kulis, Fritz, and
  Darrell]{saenko2010adapting}
Kate Saenko, Brian Kulis, Mario Fritz, and Trevor Darrell.
\newblock Adapting visual category models to new domains.
\newblock In \emph{European conference on computer vision}, pages 213--226.
  Springer, 2010.

\bibitem[Saito et~al.(2017)Saito, Ushiku, Harada, and
  Saenko]{saito2017adversarial}
Kuniaki Saito, Yoshitaka Ushiku, Tatsuya Harada, and Kate Saenko.
\newblock Adversarial dropout regularization.
\newblock \emph{arXiv preprint arXiv:1711.01575}, 2017.

\bibitem[Saito et~al.(2018)Saito, Watanabe, Ushiku, and
  Harada]{saito2018maximum}
Kuniaki Saito, Kohei Watanabe, Yoshitaka Ushiku, and Tatsuya Harada.
\newblock Maximum classifier discrepancy for unsupervised domain adaptation.
\newblock In \emph{Proceedings of the IEEE conference on computer vision and
  pattern recognition}, pages 3723--3732, 2018.

\bibitem[Saito et~al.(2019{\natexlab{a}})Saito, Kim, Sclaroff, Darrell, and
  Saenko]{saito2019semi}
Kuniaki Saito, Donghyun Kim, Stan Sclaroff, Trevor Darrell, and Kate Saenko.
\newblock Semi-supervised domain adaptation via minimax entropy.
\newblock In \emph{Proceedings of the IEEE/CVF International Conference on
  Computer Vision}, pages 8050--8058, 2019{\natexlab{a}}.

\bibitem[Saito et~al.(2019{\natexlab{b}})Saito, Ushiku, Harada, and
  Saenko]{saito2019strong}
Kuniaki Saito, Yoshitaka Ushiku, Tatsuya Harada, and Kate Saenko.
\newblock Strong-weak distribution alignment for adaptive object detection.
\newblock In \emph{Proceedings of the IEEE/CVF Conference on Computer Vision
  and Pattern Recognition}, pages 6956--6965, 2019{\natexlab{b}}.

\bibitem[Sakaridis et~al.(2018)Sakaridis, Dai, and
  Van~Gool]{sakaridis2018semantic}
Christos Sakaridis, Dengxin Dai, and Luc Van~Gool.
\newblock Semantic foggy scene understanding with synthetic data.
\newblock \emph{International Journal of Computer Vision}, 126\penalty0
  (9):\penalty0 973--992, 2018.

\bibitem[Saleh et~al.(2019)Saleh, Abobakr, Attia, Iskander, Nahavandi, Hossny,
  and Nahvandi]{saleh2019domain}
Khaled Saleh, Ahmed Abobakr, Mohammed Attia, Julie Iskander, Darius Nahavandi,
  Mohammed Hossny, and Saeid Nahvandi.
\newblock Domain adaptation for vehicle detection from bird's eye view lidar
  point cloud data.
\newblock In \emph{Proceedings of the IEEE/CVF International Conference on
  Computer Vision Workshops}, pages 0--0, 2019.

\bibitem[Salimans et~al.(2016)Salimans, Goodfellow, Zaremba, Cheung, Radford,
  and Chen]{salimans2016improved}
Tim Salimans, Ian Goodfellow, Wojciech Zaremba, Vicki Cheung, Alec Radford, and
  Xi~Chen.
\newblock Improved techniques for training gans.
\newblock In \emph{Advances in neural information processing systems}, pages
  2234--2242, 2016.

\bibitem[Sanabria et~al.(2021)Sanabria, Zambonelli, and
  Ye]{sanabria2021unsupervised}
Andrea~Rosales Sanabria, Franco Zambonelli, and Juan Ye.
\newblock Unsupervised domain adaptation in activity recognition: A gan-based
  approach.
\newblock \emph{IEEE Access}, 9:\penalty0 19421--19438, 2021.

\bibitem[Sankaranarayanan et~al.(2018)Sankaranarayanan, Balaji, Castillo, and
  Chellappa]{sankaranarayanan2018generate}
Swami Sankaranarayanan, Yogesh Balaji, Carlos~D Castillo, and Rama Chellappa.
\newblock Generate to adapt: Aligning domains using generative adversarial
  networks.
\newblock In \emph{CVPR}, 2018.

\bibitem[Santurkar et~al.(2018)Santurkar, Tsipras, Ilyas, and
  Madry]{santurkar2018does}
Shibani Santurkar, Dimitris Tsipras, Andrew Ilyas, and Aleksander Madry.
\newblock How does batch normalization help optimization?
\newblock \emph{Advances in neural information processing systems}, 31, 2018.

\bibitem[Sato et~al.(2017)Sato, Manabe, Noji, and
  Matsumoto]{sato2017adversarial}
Motoki Sato, Hitoshi Manabe, Hiroshi Noji, and Yuji Matsumoto.
\newblock Adversarial training for cross-domain universal dependency parsing.
\newblock In \emph{Proceedings of the CoNLL 2017 Shared Task: Multilingual
  Parsing from Raw Text to Universal Dependencies}, pages 71--79, 2017.

\bibitem[Shao et~al.(2014)Shao, Zhu, and Li]{shao2014transfer}
Ling Shao, Fan Zhu, and Xuelong Li.
\newblock Transfer learning for visual categorization: A survey.
\newblock \emph{IEEE transactions on neural networks and learning systems},
  26\penalty0 (5):\penalty0 1019--1034, 2014.

\bibitem[Sharir et~al.(2020)Sharir, Peleg, and Shoham]{sharir2020cost}
Or~Sharir, Barak Peleg, and Yoav Shoham.
\newblock The cost of training nlp models: A concise overview.
\newblock \emph{arXiv preprint arXiv:2004.08900}, 2020.

\bibitem[Shen et~al.(2018)Shen, Qu, Zhang, and Yu]{shen2018wasserstein}
Jian Shen, Yanru Qu, Weinan Zhang, and Yong Yu.
\newblock Wasserstein distance guided representation learning for domain
  adaptation.
\newblock In \emph{Thirty-second AAAI conference on artificial intelligence},
  2018.

\bibitem[Shi et~al.(2018)Shi, Feng, Huang, Zhang, Ji, Liao, and
  Huang]{shi2018genre}
Ge~Shi, Chong Feng, Lifu Huang, Boliang Zhang, Heng Ji, Lejian Liao, and He-Yan
  Huang.
\newblock Genre separation network with adversarial training for cross-genre
  relation extraction.
\newblock In \emph{Proceedings of the 2018 Conference on Empirical Methods in
  Natural Language Processing}, pages 1018--1023, 2018.

\bibitem[Shin et~al.(2020)Shin, Woo, Pan, and Kweon]{shin2020two}
Inkyu Shin, Sanghyun Woo, Fei Pan, and In~So Kweon.
\newblock Two-phase pseudo label densification for self-training based domain
  adaptation.
\newblock In \emph{European Conference on Computer Vision}, pages 532--548.
  Springer, 2020.

\bibitem[Shrivastava et~al.(2017)Shrivastava, Pfister, Tuzel, Susskind, Wang,
  and Webb]{shrivastava2017learning}
Ashish Shrivastava, Tomas Pfister, Oncel Tuzel, Joshua Susskind, Wenda Wang,
  and Russell Webb.
\newblock Learning from simulated and unsupervised images through adversarial
  training.
\newblock In \emph{Proceedings of the IEEE conference on computer vision and
  pattern recognition}, pages 2107--2116, 2017.

\bibitem[Shu et~al.(2018)Shu, Bui, Narui, and Ermon]{shu2018dirt}
Rui Shu, Hung~H Bui, Hirokazu Narui, and Stefano Ermon.
\newblock A dirt-t approach to unsupervised domain adaptation.
\newblock In \emph{ICLR}, 2018.

\bibitem[Sindagi et~al.(2020)Sindagi, Oza, Yasarla, and
  Patel]{sindagi2020prior}
Vishwanath~A Sindagi, Poojan Oza, Rajeev Yasarla, and Vishal~M Patel.
\newblock Prior-based domain adaptive object detection for hazy and rainy
  conditions.
\newblock In \emph{European Conference on Computer Vision}, pages 763--780.
  Springer, 2020.

\bibitem[Sohn et~al.(2017)Sohn, Liu, Zhong, Yu, Yang, and
  Chandraker]{sohn2017unsupervised}
Kihyuk Sohn, Sifei Liu, Guangyu Zhong, Xiang Yu, Ming-Hsuan Yang, and Manmohan
  Chandraker.
\newblock Unsupervised domain adaptation for face recognition in unlabeled
  videos.
\newblock In \emph{Proceedings of the IEEE International Conference on Computer
  Vision}, pages 3210--3218, 2017.

\bibitem[Soto et~al.(2020)Soto, Costa, Feitosa, Happ, Ortega, Noa, Almeida, and
  Heipke]{soto2020domain}
PJ~Soto, GAOP Costa, RQ~Feitosa, PN~Happ, MX~Ortega, J~Noa, CA~Almeida, and
  Christian Heipke.
\newblock Domain adaptation with cyclegan for change detection in the amazon
  forest.
\newblock \emph{ISPRS Archives; 43, B3}, 43\penalty0 (B3):\penalty0 1635--1643,
  2020.

\bibitem[Sun and Saenko(2016)]{sun2016deep}
Baochen Sun and Kate Saenko.
\newblock Deep coral: Correlation alignment for deep domain adaptation.
\newblock In \emph{ECCV}, 2016.

\bibitem[Sun et~al.(2016)Sun, Feng, and Saenko]{sun2016return}
Baochen Sun, Jiashi Feng, and Kate Saenko.
\newblock Return of frustratingly easy domain adaptation.
\newblock In \emph{Proceedings of the AAAI Conference on Artificial
  Intelligence}, volume~30, 2016.

\bibitem[Sun et~al.(2015)Sun, Shi, and Wu]{sun2015survey}
Shiliang Sun, Honglei Shi, and Yuanbin Wu.
\newblock A survey of multi-source domain adaptation.
\newblock \emph{Information Fusion}, 24:\penalty0 84--92, 2015.

\bibitem[Tajbakhsh et~al.(2020)Tajbakhsh, Jeyaseelan, Li, Chiang, Wu, and
  Ding]{tajbakhsh2020embracing}
Nima Tajbakhsh, Laura Jeyaseelan, Qian Li, Jeffrey~N Chiang, Zhihao Wu, and
  Xiaowei Ding.
\newblock Embracing imperfect datasets: A review of deep learning solutions for
  medical image segmentation.
\newblock \emph{Medical Image Analysis}, 63:\penalty0 101693, 2020.

\bibitem[Tan et~al.(2018)Tan, Sun, Kong, Zhang, Yang, and Liu]{tan2018survey}
Chuanqi Tan, Fuchun Sun, Tao Kong, Wenchang Zhang, Chao Yang, and Chunfang Liu.
\newblock A survey on deep transfer learning.
\newblock In \emph{International conference on artificial neural networks},
  pages 270--279. Springer, 2018.

\bibitem[Tang and Zhang(2020)]{tang2020conditional}
Xingliang Tang and Xianrui Zhang.
\newblock Conditional adversarial domain adaptation neural network for motor
  imagery eeg decoding.
\newblock \emph{Entropy}, 22\penalty0 (1):\penalty0 96, 2020.

\bibitem[Tonutti et~al.(2019)Tonutti, Ruffaldi, Cattaneo, and
  Avizzano]{tonutti2019robust}
Michele Tonutti, Emanuele Ruffaldi, Alessandro Cattaneo, and Carlo~Alberto
  Avizzano.
\newblock Robust and subject-independent driving manoeuvre anticipation through
  domain-adversarial recurrent neural networks.
\newblock \emph{Robotics and Autonomous Systems}, 115:\penalty0 162--173, 2019.

\bibitem[Tran et~al.(2019)Tran, Sohn, Yu, Liu, and Chandraker]{tran2019gotta}
Luan Tran, Kihyuk Sohn, Xiang Yu, Xiaoming Liu, and Manmohan Chandraker.
\newblock Gotta adapt'em all: Joint pixel and feature-level domain adaptation
  for recognition in the wild.
\newblock In \emph{Proceedings of the IEEE Conference on Computer Vision and
  Pattern Recognition}, pages 2672--2681, 2019.

\bibitem[Triguero et~al.(2015)Triguero, Garc{\i}a, and
  Herrera]{triguero2015self}
Isaac Triguero, Salvador Garc{\i}a, and Francisco Herrera.
\newblock Self-labeled techniques for semi-supervised learning: taxonomy,
  software and empirical study.
\newblock \emph{Knowledge and Information Systems}, 42\penalty0 (2):\penalty0
  245--284, 2015.

\bibitem[Tsai et~al.(2018)Tsai, Hung, Schulter, Sohn, Yang, and
  Chandraker]{Tsai_adaptseg_2018}
Yi-Hsuan Tsai, Wei-Chih Hung, Samuel Schulter, Kihyuk Sohn, Ming-Hsuan Yang,
  and Manmohan Chandraker.
\newblock Learning to adapt structured output space for semantic segmentation.
\newblock In \emph{CVPR}, 2018.

\bibitem[Tzeng et~al.(2014)Tzeng, Hoffman, Zhang, Saenko, and
  Darrell]{tzeng2014deep}
Eric Tzeng, Judy Hoffman, Ning Zhang, Kate Saenko, and Trevor Darrell.
\newblock Deep domain confusion: Maximizing for domain invariance.
\newblock \emph{arXiv preprint arXiv:1412.3474}, 2014.

\bibitem[Tzeng et~al.(2015)Tzeng, Hoffman, Darrell, and
  Saenko]{tzeng2015simultaneous}
Eric Tzeng, Judy Hoffman, Trevor Darrell, and Kate Saenko.
\newblock Simultaneous deep transfer across domains and tasks.
\newblock In \emph{Proceedings of the IEEE international conference on computer
  vision}, pages 4068--4076, 2015.

\bibitem[Tzeng et~al.(2017)Tzeng, Hoffman, Saenko, and
  Darrell]{tzeng2017adversarial}
Eric Tzeng, Judy Hoffman, Kate Saenko, and Trevor Darrell.
\newblock Adversarial discriminative domain adaptation.
\newblock In \emph{CVPR}, 2017.

\bibitem[Van~Engelen and Hoos(2020)]{van2020survey}
Jesper~E Van~Engelen and Holger~H Hoos.
\newblock A survey on semi-supervised learning.
\newblock \emph{Machine Learning}, 109\penalty0 (2):\penalty0 373--440, 2020.

\bibitem[Volpi et~al.(2018)Volpi, Morerio, Savarese, and
  Murino]{volpi2018adversarial}
Riccardo Volpi, Pietro Morerio, Silvio Savarese, and Vittorio Murino.
\newblock Adversarial feature augmentation for unsupervised domain adaptation.
\newblock In \emph{Proceedings of the IEEE conference on computer vision and
  pattern recognition}, pages 5495--5504, 2018.

\bibitem[VS et~al.(2021)VS, Gupta, Oza, Sindagi, and Patel]{vs2021mega}
Vibashan VS, Vikram Gupta, Poojan Oza, Vishwanath~A Sindagi, and Vishal~M
  Patel.
\newblock Mega-cda: Memory guided attention for category-aware unsupervised
  domain adaptive object detection.
\newblock In \emph{Proceedings of the IEEE/CVF Conference on Computer Vision
  and Pattern Recognition}, pages 4516--4526, 2021.

\bibitem[Vyas et~al.(2018)Vyas, Jammalamadaka, Zhu, Das, and
  Willke]{Vyas2018Out}
Apoorv Vyas, Nataraj Jammalamadaka, Xia Zhu, Dipankar Das, and Theodore~L.
  Willke.
\newblock Out-of-distribution detection using an ensemble of self supervised
  leave-out classifiers.
\newblock \emph{ECCV}, 2018.

\bibitem[Wang et~al.()Wang, Wang, Liu, Xu, Wang, Zheng, Dong, Wang, Zhang, and
  Xie]{wang3916770advanced}
Chen Wang, Jing Wang, Xiaofeng Liu, Manzhu Xu, Fangyun Wang, Lin Zheng, Huahua
  Dong, Binbin Wang, Xin Zhang, and Wanqing Xie.
\newblock Advanced congenital heart disease diagnosis based on automatic
  generation of echocardiogram.
\newblock \emph{Available at SSRN 3916770}.

\bibitem[Wang et~al.(2018)Wang, Macnaught, Papanastasiou, MacGillivray, and
  Newby]{wang2018unsupervised}
Chengjia Wang, Gillian Macnaught, Giorgos Papanastasiou, Tom MacGillivray, and
  David Newby.
\newblock Unsupervised learning for cross-domain medical image synthesis using
  deformation invariant cycle consistency networks.
\newblock In \emph{International Workshop on Simulation and Synthesis in
  Medical Imaging}, pages 52--60. Springer, 2018.

\bibitem[Wang et~al.(2020)Wang, Shelhamer, Liu, Olshausen, and
  Darrell]{wang2020tent}
Dequan Wang, Evan Shelhamer, Shaoteng Liu, Bruno Olshausen, and Trevor Darrell.
\newblock Tent: Fully test-time adaptation by entropy minimization.
\newblock \emph{arXiv preprint arXiv:2006.10726}, 2020.

\bibitem[Wang et~al.(2021{\natexlab{a}})Wang, Chen, Ding, Li, Yang, and
  Zhang]{wang2021inter}
Guijin Wang, Ming Chen, Zijian Ding, Jiawei Li, Huazhong Yang, and Ping Zhang.
\newblock Inter-patient ecg arrhythmia heartbeat classification based on
  unsupervised domain adaptation.
\newblock \emph{Neurocomputing}, 454:\penalty0 339--349, 2021{\natexlab{a}}.

\bibitem[Wang et~al.(2021{\natexlab{b}})Wang, He, Fang, Chen, Li, and
  Shi]{wang2021unsupervised}
Jing Wang, Yi~He, Wangyi Fang, Yiwei Chen, Wanyue Li, and Guohua Shi.
\newblock Unsupervised domain adaptation model for lesion detection in retinal
  oct images.
\newblock \emph{Physics in Medicine \& Biology}, 66\penalty0 (21):\penalty0
  215006, 2021{\natexlab{b}}.

\bibitem[Wang et~al.(2021{\natexlab{c}})Wang, Liu, Wang, Zheng, Gao, Zhang,
  Zhang, Xie, and Wang]{wang2021automated}
Jing Wang, Xiaofeng Liu, Fangyun Wang, Lin Zheng, Fengqiao Gao, Hanwen Zhang,
  Xin Zhang, Wanqing Xie, and Binbin Wang.
\newblock Automated interpretation of congenital heart disease from multi-view
  echocardiograms.
\newblock \emph{Medical Image Analysis}, 69:\penalty0 101942,
  2021{\natexlab{c}}.

\bibitem[Wang and Deng(2018)]{wang2018deep}
Mei Wang and Weihong Deng.
\newblock Deep visual domain adaptation: A survey.
\newblock \emph{Neurocomputing}, 312:\penalty0 135--153, 2018.

\bibitem[Wang et~al.(2019)Wang, Jin, Long, Wang, and
  Jordan]{wang2019transferable}
Ximei Wang, Ying Jin, Mingsheng Long, Jianmin Wang, and Michael Jordan.
\newblock Transferable normalization: Towards improving transferability of deep
  neural networks.
\newblock \emph{arXiv preprint arXiv:2019}, 2019.

\bibitem[Wang et~al.(2017)Wang, Li, Dai, and Van~Gool]{wang2017deep}
Yifei Wang, Wen Li, Dengxin Dai, and Luc Van~Gool.
\newblock Deep domain adaptation by geodesic distance minimization.
\newblock In \emph{Proceedings of the IEEE International Conference on Computer
  Vision Workshops}, pages 2651--2657, 2017.

\bibitem[Wei et~al.(2021)Wei, Shen, Chen, and Ma]{wei2021theoretical}
Colin Wei, Kendrick Shen, Yining Chen, and Tengyu Ma.
\newblock Theoretical analysis of self-training with deep networks on unlabeled
  data.
\newblock \emph{arXiv preprint arXiv:2010.03622}, 2021.

\bibitem[Wei and Hsu(2018)]{wei2018generative}
Kai-Ya Wei and Chiou-Ting Hsu.
\newblock Generative adversarial guided learning for domain adaptation.
\newblock In \emph{BMVC}, page 100, 2018.

\bibitem[Wilson and Cook(2020)]{wilson2020survey}
Garrett Wilson and Diane~J Cook.
\newblock A survey of unsupervised deep domain adaptation.
\newblock \emph{ACM Transactions on Intelligent Systems and Technology (TIST)},
  11\penalty0 (5):\penalty0 1--46, 2020.

\bibitem[Wilson et~al.(2020)Wilson, Doppa, and Cook]{wilson2020multi}
Garrett Wilson, Janardhan~Rao Doppa, and Diane~J Cook.
\newblock Multi-source deep domain adaptation with weak supervision for
  time-series sensor data.
\newblock In \emph{Proceedings of the 26th ACM SIGKDD International Conference
  on Knowledge Discovery \& Data Mining}, pages 1768--1778, 2020.

\bibitem[Wu et~al.(2019{\natexlab{a}})Wu, Winston, Kaushik, and
  Lipton]{wu2019domain}
Yifan Wu, Ezra Winston, Divyansh Kaushik, and Zachary Lipton.
\newblock Domain adaptation with asymmetrically-relaxed distribution alignment.
\newblock In \emph{International Conference on Machine Learning}, pages
  6872--6881, 2019{\natexlab{a}}.

\bibitem[Wu et~al.(2020)Wu, Inkpen, and El-Roby]{wu2020dual}
Yuan Wu, Diana Inkpen, and Ahmed El-Roby.
\newblock Dual mixup regularized learning for adversarial domain adaptation.
\newblock \emph{ECCV}, 2020.

\bibitem[Wu and He(2018)]{wu2018group}
Yuxin Wu and Kaiming He.
\newblock Group normalization.
\newblock In \emph{Proceedings of the European conference on computer vision
  (ECCV)}, pages 3--19, 2018.

\bibitem[Wu et~al.(2019{\natexlab{b}})Wu, Wang, Gonzalez, Goldstein, and
  Davis]{wu2019ace}
Zuxuan Wu, Xin Wang, Joseph~E Gonzalez, Tom Goldstein, and Larry~S Davis.
\newblock Ace: Adapting to changing environments for semantic segmentation.
\newblock In \emph{Proceedings of the IEEE/CVF International Conference on
  Computer Vision}, pages 2121--2130, 2019{\natexlab{b}}.

\bibitem[Wulfmeier et~al.(2018)Wulfmeier, Bewley, and
  Posner]{wulfmeier2018incremental}
Markus Wulfmeier, Alex Bewley, and Ingmar Posner.
\newblock Incremental adversarial domain adaptation for continually changing
  environments.
\newblock In \emph{2018 IEEE International conference on robotics and
  automation (ICRA)}, pages 4489--4495. IEEE, 2018.

\bibitem[Xing et~al.(2022)Xing, Liu, Kuo, Fakhri, and Woo]{xing2022brain}
Fangxu Xing, Xiaofeng Liu, Jay Kuo, Georges Fakhri, and Jonghye Woo.
\newblock Brain mr atlas construction using symmetric deep neural inpainting.
\newblock \emph{IEEE Journal of Biomedical and Health Informatics}, 2022.

\bibitem[Xu et~al.(2019{\natexlab{a}})Xu, Xiao, and Lopez]{xu2019self}
Jiaolong Xu, Liang Xiao, and Antonio~M Lopez.
\newblock Self-supervised domain adaptation for computer vision tasks.
\newblock \emph{IEEE Access}, 7:\penalty0 156694--156706, 2019{\natexlab{a}}.

\bibitem[Xu et~al.(2020)Xu, Wang, Ni, Tian, and Zhang]{xu2020cross}
Minghao Xu, Hang Wang, Bingbing Ni, Qi~Tian, and Wenjun Zhang.
\newblock Cross-domain detection via graph-induced prototype alignment.
\newblock In \emph{Proceedings of the IEEE/CVF Conference on Computer Vision
  and Pattern Recognition}, pages 12355--12364, 2020.

\bibitem[Xu et~al.(2019{\natexlab{b}})Xu, Gurram, Whipps, and
  Chellappa]{xu2019wasserstein}
Pengcheng Xu, Prudhvi Gurram, Gene Whipps, and Rama Chellappa.
\newblock Wasserstein distance based domain adaptation for object detection.
\newblock \emph{arXiv preprint arXiv:1909.08675}, 2019{\natexlab{b}}.

\bibitem[Xu and Yan(2022)]{xu2022cycle}
Yayun Xu and Hua Yan.
\newblock Cycle-reconstructive subspace learning with class discriminability
  for unsupervised domain adaptation.
\newblock \emph{Pattern Recognition}, page 108700, 2022.

\bibitem[Yan et~al.(2016)Yan, Yin, Lin, Deng, Zha, and Yang]{yan2016short}
Junchi Yan, Xu-Cheng Yin, Weiyao Lin, Cheng Deng, Hongyuan Zha, and Xiaokang
  Yang.
\newblock A short survey of recent advances in graph matching.
\newblock In \emph{Proceedings of the 2016 ACM on International Conference on
  Multimedia Retrieval}, pages 167--174, 2016.

\bibitem[Yang et~al.(2018)Yang, Song, Liu, Tang, and Kuo]{yang2018image}
Chao Yang, Yuhang Song, Xiaofeng Liu, Qingming Tang, and C-C~Jay Kuo.
\newblock Image inpainting using block-wise procedural training with annealed
  adversarial counterpart.
\newblock \emph{arXiv preprint arXiv:1803.08943}, 2018.

\bibitem[Yang et~al.(2019{\natexlab{a}})Yang, Liu, Tang, and
  Kuo]{yang2019towards}
Chao Yang, Xiaofeng Liu, Qingming Tang, and C-C~Jay Kuo.
\newblock Towards disentangled representations for human retargeting by
  multi-view learning.
\newblock \emph{arXiv preprint arXiv:1912.06265}, 2019{\natexlab{a}}.

\bibitem[Yang et~al.(2020{\natexlab{a}})Yang, Sun, Carass, Zhao, Lee, Prince,
  and Xu]{yang2020unsupervised}
Heran Yang, Jian Sun, Aaron Carass, Can Zhao, Junghoon Lee, Jerry~L Prince, and
  Zongben Xu.
\newblock Unsupervised mr-to-ct synthesis using structure-constrained cyclegan.
\newblock \emph{IEEE transactions on medical imaging}, 39\penalty0
  (12):\penalty0 4249--4261, 2020{\natexlab{a}}.

\bibitem[Yang et~al.(2020{\natexlab{b}})Yang, Xu, Li, Qi, Shen, Li, and
  Lin]{yang2020adversarial}
Jihan Yang, Ruijia Xu, Ruiyu Li, Xiaojuan Qi, Xiaoyong Shen, Guanbin Li, and
  Liang Lin.
\newblock An adversarial perturbation oriented domain adaptation approach for
  semantic segmentation.
\newblock 2020{\natexlab{b}}.

\bibitem[Yang et~al.(2019{\natexlab{b}})Yang, Liu, Cheng, Kang, Chen, and
  Yu]{yang2019federated}
Qiang Yang, Yang Liu, Yong Cheng, Yan Kang, Tianjian Chen, and Han Yu.
\newblock Federated learning.
\newblock \emph{Synthesis Lectures on Artificial Intelligence and Machine
  Learning}, 13\penalty0 (3):\penalty0 1--207, 2019{\natexlab{b}}.

\bibitem[Yao et~al.(2015)Yao, Pan, Ngo, Li, and Mei]{yao2015semi}
Ting Yao, Yingwei Pan, Chong-Wah Ngo, Houqiang Li, and Tao Mei.
\newblock Semi-supervised domain adaptation with subspace learning for visual
  recognition.
\newblock In \emph{Proceedings of the IEEE conference on Computer Vision and
  Pattern Recognition}, pages 2142--2150, 2015.

\bibitem[Yi et~al.(2017)Yi, Zhang, Tan, and Gong]{yi2017dualgan}
Zili Yi, Hao Zhang, Ping Tan, and Minglun Gong.
\newblock Dualgan: Unsupervised dual learning for image-to-image translation.
\newblock In \emph{Proceedings of the IEEE international conference on computer
  vision}, pages 2849--2857, 2017.

\bibitem[Yin et~al.(2020{\natexlab{a}})Yin, Molchanov, Alvarez, Li, Mallya,
  Hoiem, Jha, and Kautz]{yin2020dreaming}
Hongxu Yin, Pavlo Molchanov, Jose~M Alvarez, Zhizhong Li, Arun Mallya, Derek
  Hoiem, Niraj~K Jha, and Jan Kautz.
\newblock Dreaming to distill: Data-free knowledge transfer via deepinversion.
\newblock In \emph{Proceedings of the IEEE/CVF Conference on Computer Vision
  and Pattern Recognition}, pages 8715--8724, 2020{\natexlab{a}}.

\bibitem[Yin et~al.(2020{\natexlab{b}})Yin, Huang, Wu, and
  Soleymani]{yin2020speaker}
Yufeng Yin, Baiyu Huang, Yizhen Wu, and Mohammad Soleymani.
\newblock Speaker-invariant adversarial domain adaptation for emotion
  recognition.
\newblock In \emph{Proceedings of the 2020 International Conference on
  Multimodal Interaction}, pages 481--490, 2020{\natexlab{b}}.

\bibitem[Yoo et~al.(2021)Yoo, Lee, and Kang]{yoo2021transferring}
Chaehwa Yoo, Hyang~Woon Lee, and Jewon Kang.
\newblock Transferring structured knowledge in unsupervised domain adaptation
  of a sleep staging network.
\newblock \emph{IEEE Journal of Biomedical and Health Informatics}, 2021.

\bibitem[You et~al.(2019)You, Wang, Long, and Jordan]{you2019toward}
Kaichao You, Ximei Wang, Mingsheng Long, and Michael Jordan.
\newblock Towards accurate model selection in deep unsupervised domain
  adaptation.
\newblock In \emph{Proceedings of the twenty-first international conference on
  Machine learning}. ACM, 2019.

\bibitem[Yu and Koltun(2015)]{yu2015multi}
Fisher Yu and Vladlen Koltun.
\newblock Multi-scale context aggregation by dilated convolutions.
\newblock \emph{arXiv preprint arXiv:1511.07122}, 2015.

\bibitem[Yu et~al.(2022)Yu, Wang, Chen, Karianakis, Shen, Yu, Lymberopoulos,
  Lu, Shi, and Chen]{yu2022sc}
Fuxun Yu, Di~Wang, Yinpeng Chen, Nikolaos Karianakis, Tong Shen, Pei Yu,
  Dimitrios Lymberopoulos, Sidi Lu, Weisong Shi, and Xiang Chen.
\newblock Sc-uda: Style and content gaps aware unsupervised domain adaptation
  for object detection.
\newblock In \emph{Proceedings of the IEEE/CVF Winter Conference on
  Applications of Computer Vision}, pages 382--391, 2022.

\bibitem[Zellinger et~al.(2017)Zellinger, Grubinger, Lughofer, Natschl{\"a}ger,
  and Saminger-Platz]{zellinger2017central}
Werner Zellinger, Thomas Grubinger, Edwin Lughofer, Thomas Natschl{\"a}ger, and
  Susanne Saminger-Platz.
\newblock Central moment discrepancy (cmd) for domain-invariant representation
  learning.
\newblock \emph{arXiv preprint arXiv:1702.08811}, 2017.

\bibitem[Zhang et~al.(2022)Zhang, Yang, Xu, Cao, Zhen, and
  Shao]{zhang2022latent}
Anran Zhang, Yandan Yang, Jun Xu, Xianbin Cao, Xiantong Zhen, and Ling Shao.
\newblock Latent domain generation for unsupervised domain adaptation object
  counting.
\newblock \emph{IEEE Transactions on Multimedia}, 2022.

\bibitem[Zhang et~al.(2019{\natexlab{a}})Zhang, Li, Xiong, Lin, Ye, and
  Yang]{zhang2019cycle}
Dan Zhang, Jingjing Li, Lin Xiong, Lan Lin, Mao Ye, and Shangming Yang.
\newblock Cycle-consistent domain adaptive faster rcnn.
\newblock \emph{IEEE Access}, 7:\penalty0 123903--123911, 2019{\natexlab{a}}.

\bibitem[Zhang et~al.(2021)Zhang, Zhang, Jia, and Zhang]{zhang2021unsupervised}
Haojian Zhang, Yabin Zhang, Kui Jia, and Lei Zhang.
\newblock Unsupervised domain adaptation of black-box source models.
\newblock \emph{arXiv preprint arXiv:2101.02839}, 2021.

\bibitem[Zhang et~al.(2020{\natexlab{a}})Zhang, Qi, Shi, and
  Gao]{zhang2020generalizable}
Jian Zhang, Lei Qi, Yinghuan Shi, and Yang Gao.
\newblock Generalizable semantic segmentation via model-agnostic learning and
  target-specific normalization.
\newblock \emph{arXiv preprint arXiv:2003.12296}, 2020{\natexlab{a}}.

\bibitem[Zhang et~al.(2019{\natexlab{b}})Zhang, Li, Ogunbona, and
  Xu]{zhang2019recent}
Jing Zhang, Wanqing Li, Philip Ogunbona, and Dong Xu.
\newblock Recent advances in transfer learning for cross-dataset visual
  recognition: A problem-oriented perspective.
\newblock \emph{ACM Computing Surveys (CSUR)}, 52\penalty0 (1):\penalty0 1--38,
  2019{\natexlab{b}}.

\bibitem[Zhang et~al.(2013)Zhang, Sch{\"o}lkopf, Muandet, and
  Zhikun]{zhang2013domain}
Kun Zhang, Bernhard Sch{\"o}lkopf, Krikamol Muandet, and Zhikun.
\newblock Domain adaptation under target and conditional shift.
\newblock In \emph{ICML}, 2013.

\bibitem[Zhang et~al.(2020{\natexlab{b}})Zhang, Wei, Wu, Zhao, Niu, Huang, and
  Tan]{zhang2020collaborative}
Yifan Zhang, Ying Wei, Qingyao Wu, Peilin Zhao, Shuaicheng Niu, Junzhou Huang,
  and Mingkui Tan.
\newblock Collaborative unsupervised domain adaptation for medical image
  diagnosis.
\newblock \emph{IEEE Transactions on Image Processing}, 29:\penalty0
  7834--7844, 2020{\natexlab{b}}.

\bibitem[Zhang(2021)]{zhang2021survey}
Youshan Zhang.
\newblock A survey of unsupervised domain adaptation for visual recognition.
\newblock \emph{arXiv preprint arXiv:2112.06745}, 2021.

\bibitem[Zhang et~al.(2018{\natexlab{a}})Zhang, Miao, Mansi, and
  Liao]{zhang2018task}
Yue Zhang, Shun Miao, Tommaso Mansi, and Rui Liao.
\newblock Task driven generative modeling for unsupervised domain adaptation:
  Application to x-ray image segmentation.
\newblock In \emph{International Conference on Medical Image Computing and
  Computer-Assisted Intervention}, pages 599--607. Springer,
  2018{\natexlab{a}}.

\bibitem[Zhang et~al.(2018{\natexlab{b}})Zhang, Wang, Cai, and
  Song]{zhang2018unsupervised}
Yun Zhang, Nianbin Wang, Shaobin Cai, and Lei Song.
\newblock Unsupervised domain adaptation by mapped correlation alignment.
\newblock \emph{IEEE Access}, 6:\penalty0 44698--44706, 2018{\natexlab{b}}.

\bibitem[Zhang et~al.(2018{\natexlab{c}})Zhang, Wang, Huang, and
  Nehorai]{zhang2018aligning}
Zhen Zhang, Mianzhi Wang, Yan Huang, and Arye Nehorai.
\newblock Aligning infinite-dimensional covariance matrices in reproducing
  kernel hilbert spaces for domain adaptation.
\newblock In \emph{Proceedings of the IEEE Conference on Computer Vision and
  Pattern Recognition}, pages 3437--3445, 2018{\natexlab{c}}.

\bibitem[Zhao et~al.(2020{\natexlab{a}})Zhao, Li, Xu, and
  Lin]{zhao2020collaborative}
Ganlong Zhao, Guanbin Li, Ruijia Xu, and Liang Lin.
\newblock Collaborative training between region proposal localization and
  classification for domain adaptive object detection.
\newblock In \emph{European Conference on Computer Vision}, pages 86--102.
  Springer, 2020{\natexlab{a}}.

\bibitem[Zhao et~al.(2021)Zhao, Xia, and Zhang]{zhao2021unsupervised}
Ranqi Zhao, Yi~Xia, and Yongliang Zhang.
\newblock Unsupervised sleep staging system based on domain adaptation.
\newblock \emph{Biomedical Signal Processing and Control}, 69:\penalty0 102937,
  2021.

\bibitem[Zhao et~al.(2018)Zhao, Wu, Gonzalez, Seshia, and
  Keutzer]{zhao2018unsupervised}
Sicheng Zhao, Bichen Wu, Joseph Gonzalez, Sanjit~A Seshia, and Kurt Keutzer.
\newblock Unsupervised domain adaptation: From simulation engine to the
  realworld.
\newblock \emph{arXiv preprint arXiv:1803.09180}, 2018.

\bibitem[Zhao et~al.(2020{\natexlab{b}})Zhao, Li, Xu, and
  Keutzer]{zhao2020multi}
Sicheng Zhao, Bo~Li, Pengfei Xu, and Kurt Keutzer.
\newblock Multi-source domain adaptation in the deep learning era: A systematic
  survey.
\newblock \emph{arXiv preprint arXiv:2002.12169}, 2020{\natexlab{b}}.

\bibitem[Zhao et~al.(2017)Zhao, Xu, Yang, Ye, Zhao, Feng, and
  Qiao]{zhao2017dual}
Wei Zhao, Wei Xu, Min Yang, Jianbo Ye, Zhou Zhao, Yabing Feng, and Yu~Qiao.
\newblock Dual learning for cross-domain image captioning.
\newblock In \emph{Proceedings of the 2017 ACM on Conference on Information and
  Knowledge Management}, pages 29--38, 2017.

\bibitem[Zhu et~al.(2017)Zhu, Park, Isola, and Efros]{zhu2017unpaired}
Jun-Yan Zhu, Taesung Park, Phillip Isola, and Alexei~A Efros.
\newblock Unpaired image-to-image translation using cycle-consistent
  adversarial networks.
\newblock In \emph{Proceedings of the IEEE international conference on computer
  vision}, pages 2223--2232, 2017.

\bibitem[Zhuang and Shen(2016)]{zhuang2016multi}
Xiahai Zhuang and Juan Shen.
\newblock Multi-scale patch and multi-modality atlases for whole heart
  segmentation of mri.
\newblock \emph{Medical image analysis}, 31:\penalty0 77--87, 2016.

\bibitem[Zou et~al.(2020)Zou, Zhu, and Yan]{zou2020unsupervised}
Danbing Zou, Qikui Zhu, and Pingkun Yan.
\newblock Unsupervised domain adaptation with dual-scheme fusion network for
  medical image segmentation.
\newblock In \emph{IJCAI}, pages 3291--3298, 2020.

\bibitem[Zou et~al.(2019)Zou, Yu, Liu, and Kumar]{zou2019confidence}
Yang Zou, Zhiding Yu, Xiaofeng Liu, and BVK Kumar.
\newblock Confidence regularized self-training.
\newblock \emph{ICCV}, 2019.

\end{thebibliography}

\end{document}